\documentclass[10pt]{article}
\usepackage[utf8]{inputenc}
\usepackage[a4paper,margin=1.4in]{geometry}
\usepackage{pdfsync,amssymb,amsmath,ifpdf,color,placeins}
\usepackage{amssymb}
\usepackage{amsmath}
\usepackage{multirow}
\usepackage{graphicx}
\usepackage{booktabs}
\usepackage{caption} 
\usepackage{placeins}
\usepackage[dvipsnames]{xcolor}

\definecolor{refcolor}{rgb}{0,0,0.5}
\ifpdf
\PassOptionsToPackage{hyphens}{url}
\usepackage[%
  pdftitle={Extracting Sentence Embeddings from Pretrained Transformer Models},%
  pdfauthor={Lukas Stankevicius and Mantas Lukosevicius},%
  pdfkeywords={BERT; embeddings; large language models; natural language processing; text embeddings; sentence vector representation; semantic similarity; transformer models; prompt engineering; unsupervised learning},%
  pdfstartview=FitH,%
  bookmarks=true,%
  bookmarksopen=true,%
  breaklinks=true,%
  colorlinks=true,%
  linkcolor=refcolor,%
  anchorcolor=blue,%
  citecolor=refcolor,%
  filecolor=blue,%
  menucolor=blue,%
  urlcolor=refcolor,%
  pdfpagelabels]{hyperref} 
\else 
\usepackage[%
  breaklinks=true,%
  colorlinks=true,%
  linkcolor=blue,%
  anchorcolor=blue,%
  citecolor=blue,%
  filecolor=blue,%
  menucolor=blue,%
  urlcolor=blue]{hyperref} 
\fi 

\providecommand{\orcid}[1]{\href{https://orcid.org/#1}{\includegraphics[scale=0.5]{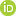}} }

\title{Extracting Sentence Embeddings from Pretrained Transformer Models}

\author{Lukas Stankevi\v{c}ius$^{\dagger}$\orcid{0000-0003-0012-5471} \and Mantas Luko\v{s}evi\v{c}ius$^{\ddagger}$\orcid{0000-0001-7963-285X}\vspace{5pt}
}

\date{{\small Faculty of Informatics, Kaunas University of Technology,\\
LT-51368 Kaunas, Lithuania\\ 
$\dagger$ lukas.stankevicius@ktu.lt, \quad $\ddagger$ mantas.lukosevicius@ktu.lt}\\[2ex]%
\today}

\begin{document}

\maketitle

\begin{abstract}
Pre-trained transformer models shine in many natural language processing tasks and therefore are expected to bear the representation of the input sentence or text meaning. These sentence-level embeddings are also important in retrieval-augmented generation. But do commonly used plain averaging or prompt templates sufficiently capture and represent the underlying meaning? 
After providing a comprehensive review of existing sentence embedding extraction and refinement methods, we thoroughly test different combinations and our original extensions of the most promising ones on pretrained models. Namely, given 110 M parameters, BERT's hidden representations from multiple layers, and many tokens, we try diverse ways to extract optimal sentence embeddings. We test various token aggregation and representation post-processing techniques. We also test multiple ways of using a general Wikitext dataset to complement BERT's sentence embeddings. All methods are tested on eight Semantic Textual Similarity (STS), six short text clustering, and twelve classification tasks. We also evaluate our representation-shaping techniques on other static models, including random token representations. 
Proposed representation extraction methods improve the performance on STS and clustering tasks for all models considered. Very high improvements for static token-based models, especially random embeddings for STS tasks, almost reach the performance of BERT-derived representations. 
Our work shows that the representation-shaping techniques significantly improve sentence embeddings extracted from BERT-based and simple baseline models.
\end{abstract}


\section{Introduction}

Early work on learnable word-level representations \cite{NIPS2013_9aa42b31} showed that semantic meaning can be embedded in numerical vector representations. Arithmetic operations such as \texttt{king $-$ man + woman $\approx$ queen} were valid. But can similar or higher abilities also be achieved for whole sentences, not just individual words? They would enable better clustering, classification, and other tasks depending on the whole meaning of the word sequence.

At the core of recent advancements in artificial intelligence is the transformer architecture \cite{NIPS2017_3f5ee243}. Compared to previous RNN-based models, it is fully parallelizable, allowing faster training and greater generalization acquisition from the data. On their 10th birthday, a child is expected to have already encountered and understood the meaning of more than 100 million words \cite{wordgap}, and now large language models surpass such scales by multiple orders of magnitude. In addition to improved throughput, transformer models are based on a self-attention mechanism \cite{BahdanauEtal15} that allows the model to attend to relevant parts of the input sequence. That is similar to how humans understand and experience the meaning of words: using context. The success of the transformer architecture in Natural Language Processing (NLP) tasks, starting with the BERT model \cite{devlin-etal-2019-bert}, was also repeated in other fields such as vision \cite{dosovitskiy2021an}, speech \cite{conformer}, and reinforcement learning \cite{chen2021decision}. Transformer models enabled the solving of more sophisticated tasks such as sentiment analysis or question answering, as well as became the state-of-the-art in almost every NLP task. Therefore, it should possess the representation of the whole text sequence.

Transformer models generally have at least 12 layers and their parameters are counted in hundreds of millions or much more. Condensing the sentence representation using these weights into a fixed 768-length vector (a common length for token vectors in base transformer models) is a challenge. It turns out that simply extracting features from a transformer model's last layer activations yields even worse results than much simpler models \cite{reimers-gurevych-2019-sentence}. There were multiple works that tried to optimize the architecture parameters (see \cite{narang-etal-2021-transformer}, which evaluated multiple proposed modifications), and multiple investigative works probing the properties of different parts of the model (see a critical review \cite{10.1162/coli_a_00422} of such approaches). Only the following methods gave tangible results: (1) the plain embedding averaging of all tokens comprising the sequence; (2) engineering a prompt template to condensate sentence representation into a single token; and (3) using a specially dedicated fine-tuned model designed to produce such vectors.

Fine-tuning is definitely the most efficient option. Resulting models, such as InferSent~\cite{conneau-etal-2017-supervised} or Sentence-BERT \cite{reimers-gurevych-2019-sentence}, showed state-of-the-art performance in sentence-level tasks at the time. However, the success of fine-tuning depends on several factors. It requires high-quantity and high-quality target domain data, as well as computational resources, which may not always be available. Sometimes, even the existing target domain data cannot be used, as they are very expensive to label. Other difficulties emerge if the data contain sensitive or private information and present a risk of it surfacing during inference. Having such constraints, one has to resort to the first two options of using raw encoded features to produce a vector for a text sequence.

Using feature aggregation instead of fine-tuning also allows us to better explain the inner workings of the state-of-the-art black-box models. Different parts of a transformer model may be responsible for different levels of representation, which favor different tasks. It is also important which levels in the representation hierarchy are easier to shape or process. A better understanding of the inner workings could help address hallucinations or other problems that current large consumer-grade language models face.

One of the two pre-training tasks of the famous BERT model \cite{devlin-etal-2019-bert} was next sentence prediction. It was optimized through a special \texttt{[CLS]} token, which sought to capture the whole-sequence-level representation. But in later works, such as  \cite{reimers-gurevych-2019-sentence}, it was revealed that such a representation is very poor, not better than the classic ones, and the authors opted for simple averaging of the last layer tokens instead. The authors of \cite{whitening_su} proposed using averaging of tokens of the first and last layers, and the authors of \cite{huang-etal-2021-whiteningbert-easy} also included hidden token representations of the second layer. But is that really the best way to obtain a numerical representation of the text's meaning?

Pre-trained models like BERT capture a lot of useful representations, yet it is not that trivial to extract them. The authors of \cite{jiang-etal-2022-promptbert} showed that further improvements can be achieved by removing the most frequent, sub-word, uppercase, and punctuation tokens before averaging. Furthermore, even larger gains can be achieved by using a prompt template, ``\texttt{This sentence: "[X]" means [MASK]}'', where the target sentence is placed instead of \texttt{[X]} and the representation of the token \texttt{[MASK]} is used as the final representation of the whole sentence. Such a method is a good example of representation extraction without any specific fine-tuning.

Inspired by the above findings, we hypothesize that there may be more ways to distill relevant sentence-level embeddings without directly fine-tuning the pre-trained BERT model. More concretely, in our analysis, we try to find a function that would shape and extract representations of \texttt{bert base-uncased} along its layers, target tokens, and additional corpus, so that the best performance in multiple tasks would be achieved. We evaluate our approach on short text clustering, semantic textual similarity, and classification tasks.

Considering the difficulties of acquiring representations from the state-of-the-art transformer models and following the success of recent works demonstrating it, our approach offers these main contributions:
\begin{itemize}
    \item We provide an extensive and organized review of related work on producing sentence-level embeddings from transformer models.
    \item We experimentally test how multiple combinations of various of the most promising token aggregation and sentence representation post-processing techniques impact the performance of three classes of different tasks and properties of representations on several models.
    \item We propose two competitive and simple static token models as baselines: random embeddings and averaged representations (``Avg'').
    \item We propose an improvement for BERT: the BERT + Avg combined model. We experimentally test many weights and layers of how the representations of BERT and Avg can be most effectively mixed.
\end{itemize}

The rest of this paper is organized as follows. We provide a review of related work in the literature on composing word vectors and representation reshaping to obtain sentence- or text-level embeddings in Section~\ref{related_work}. In Section~\ref{methods}, we outline the experimental setting and give a detailed background on our chosen approach, models, and datasets. In Section~\ref{results}, we present the results. Finally, we summarize the findings of this work in Section~\ref{conclusion}.

\section{Related Work}\label{related_work}

Taking the BERT model as an example, we are interested in how representation can be aggregated over tokens, layers, and possibly modified, and which models produce the best representations. Finally, we look at the evaluation options.

\subsection{Composing Word Vectors}\label{composing}

Thousands of works are being carried out on word-level vectors. Now, you can easily download popular publicly accessible Word2Vec \cite{https://doi.org/10.48550/arxiv.1301.3781} and GloVe \cite{pennington2014glove} embeddings. They are lightweight and usually perfectly fit various word-level tasks. On the other hand, modeling sequences of words is considered much more challenging. Count-based approaches lose information on word order and are sparse. Learning higher-order n-gram vector space models, not just words, but phrases or sentences, leads to sparsity, as frequencies vanish for target n-grams and contexts. The latter, in particular, is the main driving force for distributed representations. We will cover special methods dedicated to sentence level in Section~\ref{sentnece_level}. Nevertheless, the easiest solution is to reuse individual word vectors of the sequence.

The Principle of Semantic Compositionality (usually called Frege's principle) states that ``the meaning of an expression is a function of, and only of, the meanings of its parts together with the method by which those parts are combined.'' \cite{pelletier1994principle}. Many scholars use this as a guide to how a sentence/paragraph/document vector should be formed. According to the principle, for much easier acquisition of word vectors, only the method of combination needs to be found. Therefore, in this subsection, we review the most popular candidate combination methods that should give us a sequence vector from its multiple word vectors.

\subsubsection{Formal Semantics}
Historically, the first methods were formal and based on logic. Here, the meaning of a sentence lies in the conditions under which it is true. A semantic parser, such as Boxer \cite{bos-2008-wide}, is used to produce semantic representations of the given raw text. One can easily imagine the parse tree as a result of this analysis. As the structure is converted to first-order logic, resulting in a formula, it can then be checked with a theorem prover or interpreted with respect to a model, which is an abstract representation of a situation or setting \cite{https://doi.org/10.1111/j.1755-2567.1970.tb00434.x, https://doi.org/10.1111/j.1749-818X.2011.00284.x,  https://doi.org/10.1002/lnco.362}. Formal representations ensure that both semantic and syntactic information is preserved.

Unfortunately, formal methods have many practical shortcomings. A logic-based system must explicitly maintain the lexical knowledge necessary for the inference. Therefore, expensive human labor is involved in the construction and maintenance of these knowledge resources. More importantly, it must be domain-specific. One cannot simply use all the knowledge bases of the entire community, as this would hinder complex inferences of theorem provers and model building. There is an area of research on reducing processing time, given the large amount of knowledge resources \cite{Yoshikawa_Mineshima_Noji_Bekki_2019}. Due to this property, logic-based systems have been criticized for their lack of robustness and scalability; implemented systems tend to be small-scale and domain-specific \cite{Clark2007CombiningSA}. Although being white-box, theoretically clear, and promising, in practice, formal semantics methods are often surpassed by unsupervised distributional approaches capable of utilizing huge amounts of data. As shown in \cite{bjerva-etal-2014-meaning}, symbolic representations can at least provide additional features for neural approaches. 

\subsubsection{Tensor Products}

In \cite{SMOLENSKY1990159, Clark2007CombiningSA}, it was proposed to combine two representations, each in vector form, using tensor products, i.e., every element of the first vector is multiplied by every element of the second vector, retaining the products of all the pair combinations. This way, two rank-1 tensors result in a rank-2 tensor. Combining more vectors results in higher-rank tensors. It allows the tensors to represent the relations and role-filler bindings in a distributed fashion. However, it raises problems due to the dimensionality growing exponentially in size as more constituents are composed \cite{https://doi.org/10.1111/j.1551-6709.2010.01106.x}. Furthermore, as noted in \cite{milajevs-etal-2014-evaluating}, tensor-based models can only efficiently handle sentences of a fixed structure. Unfortunately, in most practical applications, this is not the case.

Some methods try to solve the mentioned problems by sticking to the original vector dimensionality. It can be accomplished using convolution methods \cite{jones2007representing, 33477}. For example, the circular convolution, as presented in \cite{377968}, achieves compression by summing along the transdiagonal elements of the tensor product. The compression is lossy, but the noisy version of the original vector can be recovered using circular correlation \cite{https://doi.org/10.1111/j.1551-6709.2010.01106.x} and matched to the original one by comparing with all known component vectors. Due to the same mathematical principles as light holography, these models are also referred to as holographic.

Some works related to tensor products stem from formal semantics in a way that syntax drives the compositional process. In particular, it was expected that one of the most important parts of the sentence is the pair of a noun and an adjective. Machine learning, in particular regression-based, can be used to find the composition method. The idea is that vectors for AN (adjective--noun) pairs can be learned in the same fashion as for regular words. Then, given the constituents and the final AN vector, a linear mapping is learned. In \cite{baroni-zamparelli-2010-nouns}, the mapping is learned as a separate matrix for each adjective, while in \cite{guevara-2010-regression}, a generic ``AN-slot'' function is trained. These methods do not increase the dimensionality, yet it is not clear how they would work with more words than only the AN pairs. A later work \cite{grefenstette-etal-2013-multi} extended the idea of the adjective matrix of \cite{baroni-zamparelli-2010-nouns} to other types (in addition to adjectives and nouns). They emulate formal semantics by representing functions as tensors and arguments as vectors. This way, subjects and objects are rank-1 tensors, while verbs are rank-3. However, as the authors trained only nouns, verbs, subject--verb pairs, and subject--verb--object triplets, their method is still limited to phrases no longer than three words.

Although tensor products share some exotic mathematical properties with quantum mechanics, for example, quantum entanglement \cite{33477}, the framework relies heavily on formal semantics (see \cite{baroni-etal-2014-frege}). Therefore, it also shares the same weaknesses and is outperformed by pure machine learning approaches.

\subsubsection{Averaging}\label{related_averaging}
As neural vectors became more effective than traditional approaches \cite{milajevs-etal-2014-evaluating}, it turned out that it was common to derive a vector for a sequence of words simply by summing or averaging individual vectors. In 2009, the authors of \cite{https://doi.org/10.1111/j.1551-6709.2010.01106.x} reported that for composing a phrase representation, ``averaging is the most common form of vector combination''. Despite its simplicity, the basic rule of averaging word vectors was shown to work very well \cite{7041633, https://doi.org/10.48550/arxiv.1412.1632, 10.1007/978-3-319-19581-0_3}. Later work, such as \cite{ritter-etal-2015-leveraging, 10.1145/2838931.2838932, shen-etal-2018-baseline}, even showed that simple pooling methods, such as basic vector addition or averaging, match or outperform much more sophisticated methods for encoding the meaning of a text sequence. The authors of
\cite{https://doi.org/10.48550/arxiv.1511.08198} reported that in out-of-domain scenarios, simple architectures such as word averaging outperform complex Long Short-Term Memory (LSTM) models. In addition, they are even competitive with systems tuned for particular tasks. Later work by \cite{aldarmaki-diab-2018-evaluation} evaluated various compositional models and found that word vector averaging performed reasonably well in most supervised benchmarks. According to the authors of \cite{reimers-gurevych-2019-sentence}, even for the currently trending BERT transformer model, to map the sentence to a single vector, the most popular approach is to average the word embeddings of the BERT output layer.

As averaging is well-suited to acquiring sentence optimization, some methods take averaging into their structure to further optimize it. One such example is the deep average network (DAN) \cite{iyyer-etal-2015-deep}. It takes the input as an averaged text sequence and passes it to one or more feedforward layers to finally perform linear classification. Another, C-PHRASE model of \cite{pham-etal-2015-jointly}, is trained to predict the contexts of phrases from the additive combination of their elements. Such a design results in a useful property of C-PHRASE, that summing word vectors yields sequence representation. The authors of \cite{weighting_words_rnn} found that one should use an average of all hidden states of LSMT, rather than using representation only from the last one. In a similar fashion, the averaging or summing is optimized in the works of the Siamese CBOW model \cite{kenter-etal-2016-siamese}, Sent2Vec model \cite{pagliardini-etal-2018-unsupervised}, and in other works \cite{gupta-jaggi-2021-obtaining, hill-etal-2016-learning-understand, joulin-etal-2017-bag}.

A model derived from BERT, specially designed for sentence embeddings, SBERT, also uses averaging to pool words from two sequences to a pair of fixed-size sentence embeddings. Later, using the pair and its element-wise difference, the representation is passed to the softmax layer for classification and regression tasks. The authors of \cite{9412102} tried to improve this configuration by replacing the averaging with a convolutional neural network. However, their improvement was limited to only better scores for the SALBERT model, which was still lower than SBERT based on averaging. Therefore, calculating the mean embedding from multiple word vectors is still a solid approach to the inner workings of current sentence models.

Averaging is used in various scenarios. Some authors employ it to derive static embeddings from contextualized models. In \cite{gupta-jaggi-2021-obtaining}, word vectors are learned by predicting them from the average of all contextual embeddings of words (except the target word) returned by the BERT encoder. Other authors of \cite{bommasani-etal-2020-interpreting} derived static word embeddings from contextualized transformer representations by averaging word embeddings for each word in 100k different contexts. Such derived static versions are of better quality than classic Word2Vec and share their benefits, such as having tens of millions of times lower computational cost than using standard contextual embedding models \cite{gupta-jaggi-2021-obtaining}. The other scenario is to compose a new more expressive meta-embedding from different models and domains. As a means of composition, the authors of \cite{yin-schutze-2016-learning} proposed concatenation with some trainable methods. However, a later work \cite{coates-bollegala-2018-frustratingly} found that averaging word vectors and padding them with zeros to compensate for dimensionality mismatches is a surprisingly effective method.

Despite its popularity, averaging has inherent deficiencies. The most prominent is the loss of word order information. For this reason, it is sometimes called a bag-of-words representation (that is, imagine that words lay in the bag in no specific order). As shown in \cite{iyyer-etal-2015-deep}, models based on averaging are weaker in tasks that require syntax information. The mean pooled static representations of the words ``dog chased human'' and ``human chased dog'' will be the same. In addition, simple summation can cause destructive interference, affecting valuable information. This is especially relevant in long documents. In addition to neglected sentence structure and word order, the authors of \cite{aldarmaki-diab-2018-evaluation} also note that individual word identities are lost and noninformative words are more prominently represented than essential ones.

However, to some extent, drawbacks can still be overcome. Most tasks do not usually rely on word order, and the number of occurrences of order-sensitive elements, such as double negations, is generally low. Word vectors have a rich representation and some related knowledge can still be found in them. For example, the authors of \cite{adi2016fine} showed that the average of the word vectors retains information about the original sequence length. It is even more the case with contextualized transformer representations, where starting from the second layer, each hidden vector is expected to ``know'' every other token vector and, therefore, to be aware of the word order.

\subsubsection{Weighted Average}

Not all constituent words are of the same importance; therefore, the corresponding vectors should be weighted. As a surprising event in Information Theory has higher information content than an expected event \cite{shannon1948mathematical}, some specific words should also add more weight to the sentence vector than others.

In an early work \cite{https://doi.org/10.1111/j.1551-6709.2010.01106.x}, we can already find different weights for the members of the pairs of adjective--noun, noun--noun, or verb--object phrases. Later, work went from formal semantics to statistics-based measures, in particular the tf-idf scores. They can be used as features on their own, such as in \cite{weighting_words}, but it is better to use them to weight neural word vectors. For example, the authors of \cite{singh-mukerjee-2015-words} proposed the Composite Document Vector as a concatenation of idf-graded weighted Word2Vec vectors and tf-idf features. The authors of \cite{arora2017simple} proposed another better weighting approach, called Smooth Inverse Frequency (SIF). Here, higher-frequency words are down-weighted smoothly. The authors of \cite{aldarmaki-diab-2018-evaluation} evaluated various compositional models and found that weighted averaging, in particular SIF, resulted in better performance in unsupervised similarity tasks that outperformed all other models. The author of \cite{ethayarajh-2018-unsupervised} presented the uSIF method, an improvement to SIF, omitting hyperparameter tuning and constructing weighting with both word frequency and word vector length information. The authors of \cite{ELABAPDB89093624} also showed that the mean pooled output of transformers using tf-idf weights is better for clustering than the only regularly averaged output.

A drastic case of weighting---a total removal of some words---is also shown to improve results. The authors of an older work \cite{https://doi.org/10.48550/arxiv.1412.1632} removed the stop words from the sequence before averaging. Meanwhile, as shown in \cite{yan-etal-2021-consert, jiang-etal-2022-promptbert}, the current performance of the BERT transformer model is also significantly improved in Semantic Text Similarity (STS) tasks if the most frequent tokens are removed. In other cases, the task itself focuses only on a few words, and others become redundant. The authors of \cite{9660801} found that for the relation extraction task, some sentence embedding methods work better with shorter spans of words than the original sentences. They performed sentence segmentation in a way that a (sub)sentence would cover the identified entity mentions. Other authors of \cite{9660801} also noted that SentenceBERT and Quickthought on spans or short segments containing two entity mentions are more clusterable than on the original sentences. These works also highlight the unequal contribution that each word has to the final sentence representation vector.

Some works use more complex word weighting schemes. For example, the authors of \cite{weighting_words_rnn} proposed a form of attention to weight each hidden state of LSTM. Other authors of \cite{yang-etal-2019-parameter} proposed the idea that each word brings a novel orthogonal basis to the sentence. Therefore, the length of a projection in this direction can be converted into a word's weight for use in the averaging. They come up with the final weight, consisting of three scores: novelty, significance, and corpus-wise %
uniqueness. Authors of \cite{9660801} found that for the methods analyzed, such weighting of GloVe embeddings made them the most clusterable. Similar work by \cite{10.1109/TASLP.2020.3008390} incorporated alignment and novelty scores and applied them to contextualized representations of SBERT. Here, the alignment measures how well the word aligns with the neighboring ones, i.e., a well-aligned one is less informative and should be weighted less. The novelty, similar to \cite{yang-etal-2019-parameter}, is expressed as the magnitude of the orthogonal component of the word to the subspace of neighboring words. Furthermore, the authors of \cite{10.1109/TASLP.2020.3008390} perform average weighting through all SBERT layers and also weight each layer-aggregated vector via $l_{1}$-normalized variance. This means that words that evolve faster across layers will receive higher weights since they have greater variance. Although complex to implement, such methods are reported to provide some minor performance improvements.

Surprisingly, the norm of word vectors has a large dispersion, as observed by the authors of \cite{https://doi.org/10.48550/arxiv.1508.02297, https://doi.org/10.48550/arxiv.1805.09209}, and during the average pooling operation, it acts as a weight of the word vector. The authors of \cite{pagliardini-etal-2018-unsupervised} observed that their trained word vector in this way down-weights frequent tokens by itself, and this weighting follows the hypothesis of Luhn \cite{5392672}, a well-known information retrieval paradigm, stating that mid-rank terms are the most significant for discriminating content. Using these insights, the authors of \cite{yokoi-etal-2020-word} used the norm of word vectors as a proxy for the importance of words.

\subsubsection{Clustering}

As shown by \cite{amiri-mohtarami-2019-vector}, averaging word vectors leads to a loss of information. It is in particular significant for longer text sequences, such as documents. Therefore, the idea was developed to aggregate vectors in a more smooth and information-preserving manner. The authors of \cite{gupta-etal-2016-product} clustered the word vectors into $k$ groups using $k$-means. Then, for a document, each cluster vector is obtained as a sum of its constituent vectors. The final representation is then the concatenation of the cluster vectors and the inverse cluster frequency (icf) values, which are calculated using the idf values of the words present in the document. Other authors of \cite{KIM2017336} proposed a similar method, but viewed it as one that reduces the dimensionality from words to concepts. Concepts (as clusters) are created by clustering word vectors generated from Word2Vec, and frequencies of these concept clusters are used to represent document vectors. Finally, similarly to tf-idf applied to bag-of-words, a weighting scheme, concept frequency-inverse document frequency (cf-idf), is applied to acquired bag-of-concepts. The authors of \cite{8938709} proposed the concepts of word containers and document containers to explain how such procedures bring benefits. Similarly to other works, they perform clustering of words into distinct clusters. Vectors belonging to the same cluster are averaged, while resulting representations among different clusters are concatenated. Using standard contextualized pre-trained transformer models with frozen weights, the authors of \cite{https://doi.org/10.48550/arxiv.2010.11351}, instead of a simple mean pooling, proposed training a categorical variational autoencoder and reported the performance improvement on some STS tasks. This way, similar to clustering, the lower intermediate compressed representation would carry the essential representation of the sequence. Overall, using clustering of independent word vectors allows the document representation to be expressed in a smaller space of concepts, whose concepts are otherwise erased due to destructive interference during the global averaging.

The authors of \cite{mekala-etal-2017-scdv} improved the aforementioned method of \cite{gupta-etal-2016-product} by using soft clustering and allowing a single word vector to participate in multiple clusters. More specifically, each word is represented as a $K \times d$ dimensional embedding, where each $k^\text{th}$ row corresponds to the original word vector weighted by its probability distribution in the $k^\text{th}$ cluster. The embedding of this word is weighted with the inverse document frequency of the word and summed with the embeddings of other words to form a document vector. As many values in such a vector were observed to be close to zero, sparseness is induced with a given minimal value threshold. The authors of \cite{mekala-etal-2017-scdv} named their method Sparse Composite Document Vector (SCDV).

The success of SCDV was supplemented by numerous other contributions. The authors of \cite{gupta2020p}, instead of the Gaussian Mixture Models (GMMs), used K-SVD (an algorithm for designing overcomplete dictionaries for sparse representation) \cite{1710377} for the topic modeling. Furthermore, they used a newer word vector algorithm Doc2VecC and a modern SIF weighting and removal of the top principal component of \cite{arora2017simple}. Other authors of \cite{gupta2020multisense}, with their SCDV-MS method, additionally performed word sense disambiguation to dissect polysemous words into distinct vectors. In this way, the quality of the clusters improved. To induce sparsity, the authors also applied hard thresholding in an earlier stage in word cluster assignments and trained word vectors with Doc2vecC. Next, with the emergence of transformer models, the authors of \cite{gupta-gupta-2021-unsupervised} utilized contextualized representations in their SCDV + BERT (ctxd) method. First, the corpus is contextualized following the technique of \cite{mekala-shang-2020-contextualized}. That is, each corpus word is clustered among its individual recurrence contexts into distinct clusters corresponding to the different meanings of the same word. After such a procedure is repeated for all words, the vector for each word will then be the centroid vector of the closest cluster. The authors of \cite{gupta-gupta-2021-unsupervised} then repeated the original SCVD procedure and the result was an average improvement for STS tasks. Therefore, the idea of turning global averaging to local inside-cluster averaging did not change; only newer representation techniques and models were employed.

\subsubsection{Spectral Methods} A text sequence comprising word embeddings can be interpreted as a multidimensional signal over time. Therefore, temporal summarization techniques can be employed.

One such method is the Discrete Cosine Transform (DCT), originally presented in \cite{1672377}. This invertible function maps an input sequence of $N$ real numbers to the coefficients of the $N$ number of orthogonal cosine basis functions. The DCT components are arranged in order of significance, with the first one being proportional to the simple average of the sequence. DCT is used in data compression by preserving only the most important coefficients.

In \cite{almarwani-etal-2019-efficient, almarwani-diab-2021-discrete}, DCT was shown to outperform simple averaging of word embeddings. First, it can be attributed to the fact that DCT is structure-sensitive, as it captures signal dynamics. Second, the DCT has multiple coefficients, with the first one already corresponding to averaging; thus, it should capture more information and enrich the performance of probing classifiers. The authors of \cite{almarwani-etal-2019-efficient} apply DCT along the sequence of words for each embedding dimension. Then, they retain lower-order coefficients and concatenate them to obtain overall feature patterns in the word sequence. One drawback of this technique is that short sentences must be padded with zeroed vectors. The following work \cite{almarwani-diab-2021-discrete} also showed good results for DCT in multilingual and cross-lingual settings.

Concerning the spectral decomposition of DCT, the authors of \cite{kayal-tsatsaronis-2019-eigensent} proposed EigenSent, which utilizes Higher-Order Dynamic Mode Decomposition (HODMD). The method summarizes transitions in a sequence of words into one representative sequence embedding. The authors found that the best performance is achieved when such an embedding is concatenated with simple word vector averaging. This way, information on both the dynamics and the scale of the sequence is captured.

\subsubsection{Using Special Tokens}

Modern transformer models have several special tokens that can play an important role during representation aggregation. For example, BERT has \texttt{[CLS]} designed for the next sentence prediction task, and it is supposed to carry the meaning of the input text sequence. The authors of \cite{kim-etal-2021-self} tried to use this property and trained the model to learn to aggregate everything in this \texttt{[CLS]} token. Other special tokens are \texttt{[SEP]}, used to separate two text inputs, \texttt{[MASK]}, to concentrate the representation of the masked word, padding, and various sentinel tokens (as in T5). Despite its intended purpose, special tokens are generally dismissed, as in \cite{10.1007/978-981-15-6168-9_13, huang-etal-2021-whiteningbert-easy, https://doi.org/10.48550/arxiv.1910.07973}, and a simple average is used instead. However, such tokens are important, as the authors of \cite{kovaleva-etal-2019-revealing}, after fine-tuning BERT, found a clear tendency for earlier layers to pay attention to \texttt{[CLS]} and for later layers to pay attention to \texttt{[SEP]}.

Recently, a prompt-based learning paradigm has emerged (see the survey \cite{https://doi.org/10.48550/arxiv.2107.13586}). Here, instead of fine-tuning the model to the downstream task, one reformulates it to look like the one that was being solved during the model pre-training process. Therefore, the main effort goes to prompt engineering, i.e., finding the most appropriate prompt for the given task. Such a strategy is especially suitable for very large, even colossal language models that are difficult to train.

The authors of \cite{jiang-etal-2022-promptbert} successfully used prompt engineering to better capture sentence embeddings with the BERT model. During the manual prompt search, the best prompt found was ``\texttt{This sentence: "[X]" means [MASK].}'', where \texttt{[X]} is replaced by the input sentence, and the output vector of token \texttt{[MASK]} is used as a final representation of the input sequence. Then, they further optimized the template by fine-tuning it on Natural Language Inference (NLI) data with the contrastive objective, while the BERT model weights were frozen. The authors used the continuous template technique of \cite{zhong-etal-2021-factual}, where each template token is treated as a vector and optimized by gradient descent. The final template was shown to outperform multiple untrained baselines. The authors also showed that different templates can be used effectively to represent the same sentence with different points of view during supervised contrastive learning. The following work by \cite{snsce} adapted the idea of using prompt instead of mean pooling and also used this representation extraction during contrastive learning.

\subsubsection{Aggregating through Layers}

One special aggregation that is possible with newer contextualized models is through deep neural network layers. For simplicity, we refer to transformer model blocks as layers.

One can imagine that ``each layer will increasingly magnify small but significant differences'' \cite{iyyer-etal-2015-deep}. Therefore, the early layers capture more fundamental and low-level information \cite{kovaleva-etal-2019-revealing}, which dominates the learning of shallow lexical and meaning-related knowledge \cite{dalvi2022discovering}. The authors of \cite{https://doi.org/10.48550/arxiv.2010.11351} reported that the word embeddings of the lower layers of BERT perform better than their upper layers on a word analogy task, and other authors of \cite{bommasani-etal-2020-interpreting} showed that the first quarter of the models' layers perform best in lexical semantic understanding. In the middle layers of the transformers, the hidden states are the most transferable \cite{liu-etal-2019-linguistic} and contain the most relevant information \cite{tenney2018what}. The authors of \cite{muller-etal-2021-first, hammerl-etal-2022-combining} report that the middle layers of the multilingual transformers are more multilingually aligned. According to the authors of \cite{dalvi2022discovering}, with the inclusion of context in the upper layers, the encoded concepts evolve into a linguistic hierarchy where the middle and upper layers have a better representation of the core linguistic and semantic concepts. Finally, as the author of \cite{ethayarajh-2019-contextual} explains, upper-layer representations become more context-specific.

A general tendency, as shown by \cite{carlsson2020semantic}, is that the STS performance increases until the middle layers before decreasing toward the final ones. The authors of \cite{carlsson2020semantic} even report that the final layer of most transformer models produces the worst-performing representations. Most scholars agree that this is due to the overspecialization of the last layers to the pre-training task \cite{chung2021rethinking}. Therefore, the final layers of BERT are the most task-specific \cite{rogers-etal-2020-primer} (but not as much as LSTM, as analyzed in \cite{liu-etal-2019-linguistic}). Even during the fine-tuning phase, the authors of \cite{kovaleva-etal-2019-revealing} determined that the last two layers encoded the highest share of task-specific features attributed to the score gain. Therefore, the last layers are specific to the training objective and cause problems if it differs a lot from the downstream task objective.

However, some work in \cite{timkey-van-schijndel-2021-bark, https://doi.org/10.48550/arxiv.2104.07456} found that it was beneficial to apply post-processing to the final layers. This enabled them to restore the apparent drop in representational quality in later layers and gave large performance improvements. It is thought that anisotropy, known as the narrow cone of embeddings in vector space, may be to blame, which in \cite{ethayarajh-2019-contextual} was found to increase from the earlier to the later layers. However, successful post-processing reveals that the last layers also possess relevant knowledge, but it is somewhat obscured in the raw outputs.

The question is how to compose a universal representation from the relevant pieces in multiple layers. One way is to try various combinations. The authors of \cite{huang-etal-2021-whiteningbert-easy} analyzed all possible two-layer combinations for BERT. They found that combining the top and bottom layers is better than using only the top layers. They tried to expand it to more layers but remained on a combination of three layers (L1 + L2 + L12). From the older works, we can mention the AdaSent model \cite{10.5555/2832747.2832816}. It operates on a hierarchy of representations derived from a pyramid-like multilayer network structure. The gating network is trained to adaptively select the most appropriate representation in the hierarchy for the given task. A similar gating system for dynamically deciding which intermediate transformer layers to use was proposed by the authors of \cite{yang2021deepening}.

The contribution of each layer can also be accounted for by weighted average. The authors of ELMO \cite{peters-etal-2018-deep}, a contextualized model based on LSTM, present this configuration with additional weight parameters for each layer. Other authors investigating representations \cite{michael-etal-2020-asking, liu-etal-2019-linguistic, tenney2018what} call this weighting a scalar mix technique. The authors of \cite{10.1109/TASLP.2020.3008390}, instead of learning layer weights for each word, compute them using complex alignment and novelty measures. For a few-layer model like ELMO, other authors of \cite{shi-etal-2019-retrofitting} found it useful to concatenate representations across layers while averaging across tokens, rather than using only the top layer. Other authors of \cite{kim-etal-2021-self} extract relevant information from intermediate hidden representations of BERT by treating them as positive samples during contrastive learning and determining which final sentence embedded in the tuned model should be close.

In conclusion, there is no generally accepted method for taking advantage of representations across all layers in an unsupervised setting. In the other, supervised case, the whole network is fine-tuned, and the whole hierarchy of representations is taught to surface representations according to the task at hand.

\subsubsection{Other Means of Composition}

Many other composition functions can be used to derive sentence embeddings. In an early work \cite{https://doi.org/10.1111/j.1551-6709.2010.01106.x} on phrase composition, the authors also found that multiplicative and dilation models perform well. However, there are caveats; element-wise multiplication of sparse representations results in loss of information (this effect is especially strong for more than two words), while dilation, just like weighted average, requires additional parameters. It should also be mentioned that max pooling is quite a popular operation to aggregate the output of the LSMT encoder, as shown in \cite{10.1162/tacl_a_00288, conneau-etal-2017-supervised}, and also by the authors of \cite{kim-etal-2021-self}, who found it to work best for all BERT layers for the STS benchmark task. According to the authors of \cite{toshniwal-etal-2020-cross}, max pooling is also competitive with boundary-based methods (such as those that use embeddings of words in the boundaries) for the representation of the text span.

The input text sequence can be combined hierarchically. The authors of \cite{socher-etal-2011-semi} used a recursive autoencoder to contract two child words to a single parent one repeatedly, to arrive at a final single common parent representation of the whole sequence. Later, the authors of \cite{socher-etal-2013-recursive} extended this combination to specifically follow the structure of the parse tree. Such a recursive model allowed them to compose word vectors in a bottom-up approach up to a sentence level and outperform the simple averaging method. However, a drawback is that such a composition is limited to only sentences. Other authors of \cite{shen-etal-2018-baseline} used similar ideas for combinations, but did not learn any composition function. Instead, given a specific window length, the vectors of words in these windows are averaged, and the resulting embeddings are then max-pooled. Such a setup better preserves the spatial information.

The common average operation can be generalized as a power mean, as proposed for information retrieval in \cite{10.1145/182.358466}. The authors of \cite{rueckle:2018} introduced such an idea for combining word vectors. They showed that using the document vector as a concatenation of the power mean with different power values, particularly $p=\pm \infty$ (maximum and minimum), $p=1$ (regular arithmetic mean), and other $p > 0$, substantially improves the representation of the document. Other authors of \cite{lin2017a} also proposed a long vector representation. They employ self-attention to replace the max (or average) pooling from either RNN hidden states or convolved n-grams, resulting in the sentence embedding of the matrix form. However, as the authors of \cite{eger-etal-2019-pitfalls} criticize, concatenating representations is problematic. According to them, bigger embeddings will always increase the performance of the linear model on top of them. Therefore, embeddings of the same size should always be compared. In light of this, concatenation is an undesirable operation. 

\subsection{Reshaping Representation Spaces}\label{reshaping}

Word and sentence vectors obtained by older Word2Vec \cite{NIPS2013_9aa42b31}, GloVe \cite{pennington2014glove}, and current state-of-the-art transformer techniques are not ideal. Therefore, additional processing techniques are applied to eliminate side effects and improve the performance of downstream tasks. In this subsection, we will describe anisotropy, a phenomenon believed to be responsible for this, the reasons for its appearance, and the techniques that attempt to improve the representation space.

\subsubsection{Isotropy}

Isotropy is defined as uniformity in all orientations. In this work, we talk about embedding spaces of words and sentences. According to the authors of \cite{rudman-etal-2022-isoscore}, ``distribution is isotropic when the variance is uniformly distributed across all dimensions''.

The anisotropic properties of independent word embeddings were first observed by the authors of \cite{mu2018allbutthetop}. They analyzed common embedding models and found that, in all cases, word vectors share a nonzero common vector; therefore, they are not zero-centered. The authors also found that the variance ratios of the first few components derived by PCA decay nearly exponentially. Furthermore, the variances explained by the leading components ``encode the frequency of the word to a significant degree''. Therefore, they proposed eliminating the common mean vector and then removing the top principal components, computed on representations from the entire vocabulary. A similar procedure was also shown in an earlier work \cite{arora2017simple} to be a very good baseline. Here, words were combined by a frequency-weighted average, and then just the first principal component (yet dataset-specific, as computed on the entire dataset) was removed. As shown by the All-But-The-Top (ABTT) method of \cite{mu2018allbutthetop}, such a simple procedure, eliminating the common mean vector and a few top dominating directions from the word vectors, greatly improves both the isotropy and the downstream task performance (for sentence tasks, a simple average of preprocessed word vectors is used). As found by \cite{ijcai2019-761}, it is already known that the isotropy of the target embedding is critical for the alignment of embeddings, which is important for areas such as domain adaptation, word embedding evaluation, and machine translation. However, the work \cite{mu2018allbutthetop} became the main stimulus for the research direction that attempts to improve the isotropy of word embeddings.

Word representation anisotropy is also observed in the latest contextualized models. The authors of \cite{gao2018representation} first observed this effect on the transformer machine translation model \cite{NIPS2017_3f5ee243}, while the author of \cite{ethayarajh-2019-contextual} found that this is the case for all ELMo \cite{peters-etal-2018-deep}, BERT \cite{devlin-etal-2019-bert}, and GPT-2 \cite{radford2019language} layers. All representations for randomly sampled words from these contextualized anisotropic spaces were observed to be highly similar, far from zero average cosine similarity. Furthermore, the embeddings of any two words are positively correlated. Finally, it was observed that all word vectors in the representation space tend to occupy a narrow cone.

The research community offered some explanations for why anisotropy may occur. We will cover them in more detail below.

\paragraph{Word Frequencies} 

The authors of \cite{schnabel-etal-2015-evaluation} were among the first to observe that word embeddings encode a surprising degree of information about word frequencies. In \cite{arora2017simple}, it was shown that simply representing sentences as a weighted average of words by their frequencies is a very competitive method. In particular, the authors computed the weighted average of the word vectors in the sentence (the weight of a word $w$ is $a/(a + p(w))$, with $a$ being a parameter and $p(w)$ the frequency) and then removed the projections of the average vectors on their first singular vector (``common component removal''). Later, the authors of \cite{mu2018allbutthetop} suggested removing more than one component. They found that the top PCA directions encode word frequency information; therefore, the method was proposed to null these directions. As this bias is very strong in representation space, it is inevitably related to the anisotropy phenomena.

Similar encoding of frequency information is also evident in contextualized embeddings of modern transformer models. The authors of \cite{li-etal-2020-sentence} investigated BERT embeddings and found that high-frequency words are all close to the origin and densely concentrated, while low-frequency words are far away and scattered sparsely. The work in \cite{bis-etal-2021-much} highlighted that anisotropy is the most pronounced in rare words. The authors of \cite{rajaee-pilehvar-2022-isotropy} also showed that the same applies to multilingual BERT embeddings; they have a biased structure towards word frequency.

This bias to word frequency reveals a weakness. The authors of \cite{Schick_Schutze_2020} constructed a dataset using semantic relations extracted from WordNet \cite{10.1145/219717.219748} to test the semantic properties of words. Using that, they showed that the ability of BERT to understand words depends highly on their frequency, and therefore rare words are neglected. It was depicted in \cite{yan-etal-2021-consert} that removing the 34 most frequent tokens before averaging can greatly boost the performance on semantic textual similarity tasks. The authors of \cite{bis-etal-2021-much} conducted a simple experiment on the CNN News corpus with popular WordPiece tokenization and found that 30\% of the corpus can be accounted for using the 13 most frequent tokens, while to cover at least 98\% of the corpus, nearly 15,000 tokens are needed. Therefore, rare words are a constituent part of modern NLP pipelines and, unfortunately, induce deficiencies.

The authors of \cite{gao2018representation} proposed an explanation for generation models that related observations of the narrow cone form in the representation space and the frequencies of words. They argue that the main culprit is the way models are trained on the language modeling task. During the training process, the ground-truth word embedding will be pushed toward the direction of the hidden state to obtain a larger likelihood. Meanwhile, the embeddings of most words in the vocabulary (with non-appearing and rarely appearing frequencies) will be pushed towards similar directions negatively correlated with most hidden states and thus end up grouped together in the local region of the embedding space. The following work \cite{bis-etal-2021-much} complemented the given explanation with the ``common enemies effect''---the effect of the target words producing gradients of the same direction for all of the non-target words at each step of training with cross-entropy loss, and rare tokens are the most affected by it. The authors also found that the embeddings learned by GPT-2, BERT, and RoBERTa do not degenerate into a narrow cone (they only appear as a cone when projected to a lower-dimensional space), but instead drift in one shared direction. Therefore, the ``common enemies effect'' is argued to be the main culprit behind the anisotropy of the representation space.

\paragraph{Outliers}

There is evidence from multiple authors that contextualized representations of transformer models (in particular, the BERT family) contain undesirable outliers (interestingly, abnormal dimensions were also previously observed for GloVe vectors in \cite{10.1145/2872518.2889381}). These are certain positions in word vectors with unusually high values. The authors of \cite{kovaleva-etal-2021-bert} called it outlier dimensions; in \cite{luo-etal-2021-positional}, it is referred to as outlier neurons, while the authors of \cite{timkey-van-schijndel-2021-bark} call it rogue dimensions. In all of these works, it is agreed that it is a significant contributor to the anisotropy of the representation space.

There is debate about what causes outliers. The authors of \cite{kovaleva-etal-2021-bert} found that outliers are essential for good downstream performance. They regard it as a distinct model property that emerges during training and even encourage one to consider it during weight initialization. In contrast, in \cite{timkey-van-schijndel-2021-bark}, it was found that the behavior of the model is not driven by outliers; rather, only a small subset of linguistic abilities is handled there (such as positional information; its relationship with outliers was revealed in \cite{luo-etal-2021-positional}). Even more strangely, the authors of \cite{rajaee-pilehvar-2022-isotropy} found that the embedding space of the multilingual BERT model does not have outliers as the English BERT does, but is still anisotropic.

There are several options to account for outliers. The most obvious solution is to remove them. The authors of \cite{luo-etal-2021-positional} showed that this brings improvements for tasks that directly use the geometry of the embeddings, such as semantic textual similarity; however, {ref. \cite{kovaleva-etal-2021-bert}} showed that disabled outliers significantly degrade both Masked Language Modeling (MLM) loss and downstream task performance. The authors of \cite{timkey-van-schijndel-2021-bark} suggest that rogue dimensions be accounted for through the standardization transformation. Both options aim to reduce the dominating effect of the highest-value dimensions during the similarity measure, such as cosine similarity or Pearson correlation. On the other hand, one can argue that the similarity measure we use, not the representation space, needs improvements.

\paragraph{Criticism} 

Many works try to improve the isotropy of representation spaces, but is that truly the right way to seek performance improvements? Some recent publications, such as \cite{bis-etal-2021-much}, expressed doubts about the role of isotropy in model performance, and the authors of \cite{ding-etal-2022-isotropy} even observed a negative relation between isotropy calibration and downstream performance. The works in \cite{bis-etal-2021-much, rudman-etal-2022-isoscore} ask whether there is truly a ``narrow cone'' in the embedding space. Other authors of \cite{jiang-etal-2022-promptbert} argue that the main problems are various biases unrelated to isotropy. Furthermore, the authors of \cite{rudman-etal-2022-isoscore} criticized multiple popular measures of isotropy and warned that they are misleading and inaccurate.

The authors of \cite{cai2021isotropy} analyzed the contextual embedding space and found isolated clusters and low-dimensional manifolds. They concluded that although the embeddings are globally anisotropic, local isotropy exists. The authors of \cite{rajaee-pilehvar-2021-cluster} continued this observation and applied the known preprocessing techniques of \cite{mu2018allbutthetop}, not globally but locally, in clustered regions of representations. They observed that sense-level information is shadowed by structural, syntactic, and tense biases (for verbs), which their method of dominant direction removal helps to reduce. Another work \cite{rajaee-pilehvar-2021-fine-tuning} showed that during the fine-tuning phase, the anisotropy becomes even worse, while, in contrast, the performance of STS increases. They found that removing the top dominant directions for fine-tuned representations becomes detrimental, as the most essential information resides there. These findings indicate that the representation space of transformer models is quite complicated and that a simple fixation on isotropy can destroy its native features. These also include biases, the usefulness of which may not be obvious to us, but models have found them helpful during the pre-training. In the end, brute-forcing the plain isotropy may harm some intricate semantic details, which we seek to capture.

\subsubsection{Post-Processing Methods}

Many post-processing methods have been proposed to improve the embeddings. Whether they improve the isotropy of the representation space or help surface-relevant information \cite{artetxe-etal-2018-uncovering}, multiple authors show the benefits of using them. However, these methods generally favor unsupervised tasks, such as semantic textual similarity, with limited improvements for downstream tasks. If enough data on the target task are available, supervised training will inevitably be a better choice, as the transformation relevant to the task is learned. However,  post-processing may be the only option if the training samples are obscure.

The simplest is the centering operation. It is especially important for the STS task that involves cosine similarities. The authors of \cite{bis-etal-2021-much} found that it restores a nearly perfectly isotropic distribution. Subtracting the mean is also the first step in several other post-processing methods. In some sense, such operations as centering can be viewed as fine-tuning to the target domain, as all target embeddings participate in providing the mean coordinates.

We already mentioned some operations related to isotropy enhancement in the previous section, such as zeroing target outliers or ABTT. In the following, we will mention the remaining important ones.

\paragraph{Z-Score Normalization} 

The Z-score describes the position of a point in terms of its distance from the mean when measured in units of standard deviation (in the distribution of the target samples). Z-score normalization, also called standardization, transforms the distribution of embeddings to have a zero mean and a unit standard deviation in all dimensions. The works in \cite{timkey-van-schijndel-2021-bark, https://doi.org/10.48550/arxiv.2104.07456} recommend that you consider it as a post-processing step. The authors of \cite{eger-etal-2019-pitfalls} advise using normalization, such as the z-score, in supervised settings as a binary hyperparameter, because it can lead to rank changes.

\paragraph{All-But-The-Top (ABTT)} 

Many works are based on the influential ABTT algorithm \cite{mu2018allbutthetop}. It has a hyperparameter, which tells the number of dominant directions to remove. Principle components are either completely removed or left intact. As a result, the main weakness is finding the right number of top components to remove, which can either cause information loss or eliminate an insufficient amount of noise. The authors of \cite{8784743} proposed post-processing through variance normalization (PVN) to normalize the variance of leading principal components to the same level instead of total component disposal. The authors of \cite{conceptor_negation} performed the removal in a softer way by employing matrix conceptors \cite{Jaeger14} in an unsupervised way. A similar technique that employs conceptors can also target specific components, such as those that incorporate social biases, and diminish them, as in \cite{karve-etal-2019-conceptor}. Other authors of \cite{10.1007/978-3-030-86383-8_36} proposed learning weights for each dominant direction removal. All these approaches try to preserve useful information residing in the top principal components while narrowing the target noise.

Authors of \cite{raunak-etal-2019-effective} combined ABTT and dimensionality reduction. They found that the best pipeline is to perform ABTT twice while performing dimensionality reduction in between. In another line of work, the authors of \cite{rajaee-pilehvar-2021-cluster, rajaee-pilehvar-2022-isotropy}, instead of global post-processing, remove local dominant directions in separate clusters in the representation space. In this way, the structure of the representation space is accounted for. Similarly, the authors of \cite{yang-etal-2019-parameter}, in light of the predecessor method of ABTT \cite{arora2017simple}, proposed to eliminate the sentence-dependent principal component, where they rerank the top principal vectors based on correlation with each sentence. This individual removal of dominant directions was shown to improve performance on the STSB task.

\paragraph{Whitening} 

The authors of \cite{whitening_su} proposed the whitening approach to alleviate the anisotropy problem of sentence embedding. The whitening operation involves centering embeddings at the origin and making different dimensions have a unit variance and be uncorrelated, turning their covariance matrix into the identity matrix. It was also found to be useful in earlier work for the alignment of bilingual embeddings \cite{Artetxe_Labaka_Agirre_2018}. As a side effect, the authors of \cite{whitening_su} showed that whitening can also be used with a dimension reduction operation. A concurrent work \cite{huang-etal-2021-whiteningbert-easy} also used a whitening algorithm to improve performance on STS tasks. In that work, the authors combined the first and last layers of BERT embeddings and then normalized them with whitening. Despite the initial success, subsequent work criticized the whitening operation. The authors of \cite{gao-etal-2021-simcse} showed that whitened representations greatly improve uniformity, but also suffer degeneration in the alignment property (for alignment and uniformity properties, see \cite{pmlr-v119-wang20k}). To make matters worse, the authors of \cite{wang2022just} called whitening a ``trick'' that only helps with similarity tasks (by partially overfitting) and harms downstream task performance.

\subsubsection{Retrofitting}

This word vector space specialization method can incorporate semantic knowledge from external resources into word embeddings. The authors of \cite{faruqui-etal-2015-retrofitting} were the first to use the term ``retrofitting'' for their post-processing step of vector space word representations. Their idea was to incorporate rich relational information from semantic lexicons to word vectors, which are trained in a data-driven fashion using plain texts. The authors showed that the new vectors are similar in both their purely distributional representations and related word types. In addition, they showed improvements in several benchmark tasks. 

Retrofitting can be useful for fitting a desired domain that is different from the original one used to pre-train the word vectors. An example is \cite{yu-etal-2016-retrofitting} where the authors retrofitted to linkage information in biomedical taxonomy. Another related field is the alignment of cross-lingual word embeddings. The authors of \cite{zhang-etal-2020-overfitting} used retrofitting to align the vectors of the source and target languages in the dictionaries. This way, translation pairs are pulled closer while minimizing deviation from the original embeddings and preserving the existing representation. In all these cases, the retrofitting is based on external relational data.

According to the authors of \cite{glavas-vulic-2018-explicit}, the weakness of retrofitting is that only the vectors of words present in the external constraints (resources) are modified. Therefore, they proposed an explicit retrofitting model in which external knowledge relations are turned into supervised training examples. Here, the distance between synonyms is supposed to be as low as possible, and between antonyms as high as possible, while between remaining words not present in the external knowledge base, it should remain the same as in the original representation space. This way, on top of word vectors, a deep feedforward neural network is trained to retrofit all the word embeddings.

Some work also found success in applying retrofitting to contextualized representations. The authors of \cite{shi-etal-2019-retrofitting} observed that for ELMo \cite{peters-etal-2018-deep} representations, the distance between the shared word in the paraphrases is even greater than the distance between ``large'' and ``small'' in random contexts. Therefore, to improve the representation capabilities, they minimized the difference in contextualized representations of the shared word in paraphrased contexts while differentiating between those in other contexts. This retrofit resulted in improved performance of downstream applications. Other authors of \cite{zheng-etal-2022-using} proposed a two-step process: first to train static vectors from contextualized ones, and then to perform retrofitting. They showed that compared to baselines, such an approach gives the best results in a range of intrinsic and extrinsic tasks.

\subsubsection{Other Methods}

Some methods are proposed to reduce anisotropy during the training process. In particular, the authors of \cite{gao2018representation} added the specific regularization. According to them, the aperture of the narrow cone of representations can be improved by minimizing the cosine similarities between any two word embeddings. Such regularization encourages the vectors to be more evenly spread and expressive. This method is highly related to the now-popular contrastive learning approach (see Section~\ref{contrastive}). Other authors of \cite{Wang2020Improving} improve the isotropy of the output embedding matrix using the spectrum control method. They guide the singular value distribution of the embedding matrix throughout the training process and control the decay rate of these singular values. 

The authors of \cite{li-etal-2020-sentence} suggest learning transformation of the embedding space of the transformer, sometimes called a flow technique. Their method transforms the BERT sentence embedding distribution into a smooth and isotropic Gaussian distribution. During unsupervised training, only the flow network is optimized, while the BERT parameters remain unchanged. Although BERT-flow showed performance improvements on multiple tasks, the technique is criticized in the literature. First, it needs a specialized implementation and has multiple additional parameters. The authors of \cite{ding-etal-2022-isotropy} report that BERT-flow requires on average 4.2 times more time per training epoch. Furthermore, for the STS benchmark task, the authors of \cite{wang2022just} report that BERT-flow with the $l_{2}$ similarity metric performs even worse than the baseline of the BERT average.

Different post-processing can help surface the different information residing in the embeddings. The authors of \cite{artetxe-etal-2018-uncovering} proposed a simple unsupervised singular value decomposition to reassign the feature weights. By changing the similarity order of their transformation, they tailored word embeddings in the semantics/syntax (tasks focusing on sing--chant or sing--singing) and similarity/relatedness (tasks solving car--automobile or car--road relationships) axes. Other authors of \cite{NEURIPS2020_3acb2a20} used spectral filters to dissect BERT representations at different temporal scales. They showed that a low-pass filter yields the highest probing accuracy in topic classification, a high-pass filter in speech tagging, and a middle-pass filter is best in dialog act speech tasks.

\subsubsection{Similarity Measures}

Some most basic tasks for sentences require measuring the distances between the corresponding embeddings. Usually, cosine similarity is employed. However, if the continuous representation space is curved or a text sequence is treated as a set of words, different metrics or representation space manipulations can be beneficial. The authors of \cite{zhelezniak-etal-2020-estimating} proposed Mutual Information (MI), well known in information theory and statistics, as a candidate for a similarity measure. They managed to successfully estimate MI for continuous random variables by the use of Kraskov--St{\"o}gbauer--Grassberger (KSG) estimator \cite{PhysRevE.69.066138}, which is based on elementary nearest-neighbor statistics. The other authors of \cite{zhelezniak-etal-2019-correlation} advised comparing sentence embeddings consisting of word vectors averaged by rank correlation, such as Kendall's $\tau$. They argued that such measures enable mean pooled representations to rival modern deep ones, used with cosine similarity. The authors of \cite{zhelezniak2018dont} proposed treating sentences as fuzzy sets of words and showed good performance with the specially adapted DynaMax similarity measure. They showed that word vectors alone are sufficient to achieve excellent performance on semantic textual similarity tasks when sentence embeddings and similarity measures are derived using ideas from fuzzy set theory. Other authors of \cite{min-etal-2021-locality-preserving} suggest processing each sentence with respect to its found $k$-nearest neighbors. Then, they find an optimal Euclidean subspace of the sentence manifold where cosine similarity would work best. The authors of \cite{kayal-2021-unsupervised} also project the sentences onto a fixed-dimensional manifold with the objective of preserving local neighborhoods in the original space. All these works mentioned employ various strategies to enhance the comparison of document representations, tailoring their approaches to the given task.

Many similarity measures between sets employing earth mover's distance were proposed. The authors of \cite{pmlr-v37-kusnerb15} were the first to frame the similarity of documents as a transportation problem. The idea is that the distances between similar but different words in two documents should be small. Their Word Mover’s Distance (WMD) is the cost of transporting a set of word vectors in an embedding space. This approach was effective, as it managed to exploit similarities between different Word2Vec word vectors, such as the relation of analogies. However, the main deficiency of WMD was that the distances were expensive to compute. Moreover, an output, the single number of a distance between two given documents, can only be combined with $k$-nearest neighbors or $k$-means, while applications usually require a whole feature vector. The authors of a subsequent work \cite{wu-etal-2018-word} managed to derive vectors for documents using WMD. They constructed a positive-definite word mover’s kernel using a feature map given by the WMD to random documents and then derived document embedding via a Random Features approximation of the kernel. Other authors of \cite{yokoi-etal-2020-word} proposed Word Rotator’s Distance (WRD). It is designed so that the norm and angle of word vectors correspond to the probability mass and transportation cost in the earth mover's distance, respectively. Therefore, the norm of vectors, which is associated with the vector's significance, does not interfere with the calculation of transportation cost, as in WMD. Finally, the authors of \cite{https://doi.org/10.48550/arxiv.2002.00745} attempted to include structural information absent in the WMD's distance estimation between two sets of words. They represented the sentence vector as a weighted average of substructure vectors at a lower level in a recursive way, while a transport plan at a different level explains how the different substructures align.

\subsection{Learning Sentence Embeddings Directly}\label{sentnece_level}

We already mentioned some models that learn the aggregation of tokens or their combination rules. However, in this section, we will look at more direct approaches to learning a representation vector for a given text.

\subsubsection{Paragraph Vectors}  

The method presented in \cite{pmlr-v32-le14}, and sometimes called Doc2Vec, is one of the first successful Word2Vec \cite{https://doi.org/10.48550/arxiv.1301.3781} adaptations to sequences of tokens. The authors presented two methods of how document representation could be trained.

In the Distributed Memory model of Paragraph Vectors (PV-DM), every paragraph is mapped to a unique vector, which, together with context words, participates in predicting the next word. This way, such a unique vector acts as a memory that stores the topic of the paragraph and bears its representation. The second is the Distributed Bag Of Words version of the Paragraph Vector (PV-DBOW) model. It leaves only a unique paragraph vector in the input and predicts words randomly sampled from that paragraph in the output. However, unlike the first method, it does not account for word ordering.

In the original article \cite{pmlr-v32-le14}, authors represented sentences as a concatenation of DM and DBOW vectors, as they saw such a setup as more consistent. Later works \cite{lau-baldwin-2016-empirical, hill-etal-2016-learning} also reported that individual model performance had only marginal differences, yet deviations were usually task-dependent. 

The authors of \cite{chen2017efficient} proposed a modification to Paragraph Vectors and called it Doc2VecC. Differently from Doc2Vec, a document vector is derived not as a unique vector, but as an average of sampled constituent word embeddings. The other notable contribution is the added data-dependent regularization that favors informative or rare words while forcing the embeddings of common and non-discriminative ones to be close to zero. At the time, the authors showed Doc2VecC to match state-of-the-art in multiple tasks.

\subsubsection{To RNN- and Transformer-Based Models}

Shallow Doc2Vec-like models soon met competition from more complicated Recurrent Neural Networks (RNNs). In a sequence-to-sequence \cite{NIPS2014_a14ac55a} (like machine translation) setting, such a model usually has two parts: encoder and decoder. The encoder, going through each token one by one, encodes the input sequence into a fixed hidden representation, which is later used by the decoder to generate the target sequence (translation in NMT) token by token. Meanwhile, only the decoder part is needed to generate sentence representation. Words of the sentence are sequentially fed as input to the RNN, and the final hidden state is interpreted as its representation.

The benefit of RNN models is that the sequential consumption of tokens allows the final representation to account for the word order and process the arbitrary number of tokens. However, the effectiveness of this mechanism is questionable. It turns out that information has a hard time propagating all the way to the final hidden state. As compensation for that, a bidirectional \cite{650093} setting was used, which connects two hidden layers of opposite directions to the same output, simultaneously getting forward and backward information. The second solution was to use an attention mechanism \cite{BahdanauEtal15}, which, during decoder token generation, takes a weighted average of the encoder hidden states from all the input tokens. But it was only helpful for sequence-to-sequence tasks, like translation, as such a mechanism removed the last hidden state bottleneck, which was required to compress a long sequence of tokens into a single vector for a whole text representation.

One of the first famous recurrent models for sentence representation is the SkipThought \cite{NIPS2015_f442d33f}. This model is trained to predict neighboring sentences from the source one. The center sentence is encoded by a bidirectional GRU (Gated Recurrent Unit \cite{cho-etal-2014-learning}) that concatenates the last hidden state of a forward GRU and the last hidden state of a backward GRU, and then decodes it into the two target sentences. This way, the encoder is trained to map a sentence to a single representative vector. The authors showed that the model is robust and performed well on all tasks considered. The main drawbacks of this model are the month-long training, huge vector size of 4800 (resulting from the concatenation of 2400 vectors from two separate models), and support of only sentence-level embeddings, as well as the need for training text with a coherent inter-sentence narrative.

As recurrent GRU or LSTM \cite{lstm_paper} architectures became the standard, the most recent advances in text representation were due to how data and training objectives were chosen. Here, we have to mention a famous InferSent \cite{conneau-etal-2017-supervised} model. The authors showed that the high-quality supervised data, although in low quantity, can hugely increase the performance. They trained an encoder based on a bi-directional LSTM architecture with max pooling, on the Stanford Natural Language Inference (SNLI) \cite{bowman-etal-2015-large} dataset. The resulting model outperformed SkipThought after less than a day of training on a single GPU. Other authors of \cite{logeswaran2018an} modified the SkipThought model to QuickThoughts by framing the training task as a classification. Instead of generating the target sentence, the model encodes the likely candidates and chooses one of them. Authors of \cite{subramanian2018learning} leveraged several data sources with multiple training objectives. Their GenSen model training tasks included Neural Machine Translation, Constituency Parsing, Natural Language Inference, and SkipThought-like training. Finally, authors of \cite{nie-etal-2019-dissent} proposed even greater utilization of training data by constructing a discourse marker prediction task, predicting tokens such us \texttt{because}, \texttt{and}, \texttt{if}, etc. Such framing of the task allowed the authors to mine vast amounts of text pairs together with the connecting markers. Their DisSent model performed similarly well to InferSent on various transfer tasks. Overall, one can see that how the model is trained and data quality and quantity play a huge role in the resulting text representation performance.

Many drawbacks of the recurrent models were solved by the transformer architecture~\cite{NIPS2017_3f5ee243}. The authors decided to get rid of recurrence and leave only the attention mechanism \cite{BahdanauEtal15} itself. The resulting model (1) is highly parallelizable; (2) only after the first layer, each token has already attended to every other; and (3) it is faster than RNN models when the sequence length is smaller than the representation dimensionality. These properties allowed us to train transformer models with staggering amounts of text much faster and they became state-of-the-art models on multiple tasks.

Previous advances in sentence representation using RNNs were quickly applied to the new architecture. The authors of \cite{cer-etal-2018-universal} presented a Universal sentence encoder model. It used only the encoder part of the original transformer model trained with multiple tasks: Skip-Thought-like training, SNLI, and conversational response prediction \cite{henderson2017efficient}. After the appearance of the pre-trained BERT model, the authors of \cite{reimers-gurevych-2019-sentence} proposed SBERT: a pre-trained BERT model further fine-tuned with NLI data. It was a huge success, similar to the earlier InferSent model, yet this time both the new architecture and the general pre-training were the new factors for additional performance advancement. Other authors of \cite{cao-etal-2021-pause} tried to further utilize the unsupervised data. Their PAUSE model learned sentence embeddings from a partially labeled dataset and showed that this way, the same performance can be achieved with a smaller fraction of labeled NLI data. Here, we see the same tendency as with the RNN models: new efforts to construct better fine-tuning tasks and better utilize existing data.

\subsubsection{Contrastive Learning Approaches}\label{contrastive}

The search for better training tasks and data utilization culminated in a new area of contrastive learning.

Distance-based contrastive loss \cite{1640964} is more attractive for learning sentence-level embeddings than the more conventional error prediction losses. It allows for a simple construction of self-supervised learning using pairs of positive and negative examples. Embeddings of the first group of samples are encouraged to be the same (by utilizing a loss on a distance function between the text vectors), while the negative ones are encouraged to be different. 

In this way, the burden of expensive labeling can be relieved. Moreover, it can be used as an intermediate step between pre-training and fine-tuning to inexpensively align the model to the target task domain. These properties are especially useful for the ``data-hungry'' transformer models.

The models in this class differ mainly in how they construct the positive example pairs. This can be achieved in many different ways. We will describe them below.

\paragraph{Feature/Vector/Embedding-Level Augmentations}

These are the modifications to the sentence in its vector representation space.

Dropout \cite{dropout} is one of the most popular feature-level augmentation techniques. It was originally used as a regularization of neural networks to increase the robustness to noise from the vector space. However, the same can be applied to derive an augmented text sequence representation.

Popular BERT and RoBERTa pre-trained transformer models already have dropout layers and require almost no modifications to employ the dropout augmentation. Two positives are acquired by passing the same sample through the dropout-enabled network twice \cite{liu-etal-2021-fast, gao-etal-2021-simcse, esimsce}. This does not require additional preprocessing and can be applied on the fly during the training.

Despite the effectiveness reported, dropout augmentation has the disadvantage of being biased toward sentence length. This was observed in \cite{esimsce}. Two augmented versions of the same sentence are of the same length and can be easily discriminated against randomly drawn negatives of varying lengths. Therefore, instead of learning the general sentence representation to match the same sentence samples, the model now just takes the shortcut by only comparing their lengths. To counteract this, the authors of \cite{esimsce} added token-level augmentations in addition to dropout so that the length of positives would be different. Other authors of \cite{klein2022scd} also managed to avoid this problem by using negatives produced by a much higher dropout rate than those used for positives.

Another feature-level (also regarded as token-level) augmentation technique is the shuffle of tokens. Usually in pre-trained language models, positional information of each sequence element is brought about by the addition of special positional embeddings to the existing token ones. As a result, the shuffling of positions can be implemented simply as the shuffling of position IDs and the corresponding positional vectors.

Similarly to dropout, there is a cutoff augmentation \cite{cutoff}. Dropout can also be considered as random erasing of weights in $L \times d$ text sequence embedding matrix with $L$ tokens of length $d$ vectors. The token cutoff erases randomly selected rows, while the feature cutoff erases columns of the $L \times  d$ matrix. In the work of ConSERT \cite{yan-etal-2021-consert} it was found that the shuffle and token cutoff are the two most effective augmentation strategies, significantly outperforming the feature cutoff and dropout.

There are more sophisticated approaches to alter embeddings than a dropout randomly replacing weights with zero values. Specifically, adversarial attacks were shown to be successful. They aim to add worst-case perturbations to the input samples. The authors of \cite{yan-etal-2021-consert} used the Fast Gradient Value (FGV) \cite{rozsa2016adversarial} method, which unfortunately relies on supervised loss to compute adversarial perturbations. Other authors of \cite{2202.13093} employed the Fast Gradient Sign Method (FGSM) \cite{GoodfellowSS14}. Perturbation is obtained by applying the sign function to the derivative of the model with respect to the input by contrastive loss. The authors of \cite{2202.13093} composed their best augmentation mix as a dropout with FGSM.

\paragraph{Token-Level Augmentations} 

Language gives us many possibilities to convey almost the same meaning through similar or very different sequences of words. This can be easily utilized for augmentation purposes.

Using synonym replacement, random swapping of two consequent words, random insertion and deletion, the so-called Easy Data Augmentation (EDA) the authors of \cite{wei-zou-2019-eda} managed to reduce the required dataset size by half for the same performance in multiple classification tasks. In contrastive learning experiments for the biomedical relation extraction task, the authors of \cite{su-etal-2021-improving} found that synonym replacement outperformed random swap and random deletion. Meanwhile, one of the first contrastive works for text \cite{CERT_fang} showed that back-translation \cite{edunov-etal-2018-understanding} augmentation outperformed the four mentioned techniques. The authors took a pre-trained BERT \cite{devlin-etal-2019-bert} and pre-trained it further on the input text of GLUE \cite{wang-etal-2018-glue} tasks in a contrastive fashion, to be later fine-tuned and evaluated in GLUE. The resulting CERT model achieved slight improvements in addition to the regular BERT. CLEAR \cite{wu2020clear} was also an attempt to improve the regular BERT and RoBERTa pre-training by jointly using MLM loss and constrastive loss. The authors found that the different combinations of different augmentation strategies (in particular, substitution, span-deletion, and reordering) favor different improvements for GLUE and various SentEval tasks. A recent work \cite{2202.13093} tried token-level augmentations such as typos, synonym replacement, paraphrasing, and back-translation but found that they underperformed feature-level ones.

As the popular BERT model is pre-trained with a masked language modeling task, i.e., predicting the true masked token behind the special input \texttt{[MASK]} token, the same \texttt{[MASK]} can also be used to induce augmentations by randomly masking tokens. The mirror-BERT model of \cite{liu-etal-2021-fast} used such random span masking for input augmentation, which, together with feature augmentation, showed gains over off-the-shelf models in both lexical and sentence-level tasks, across different domains and different languages.

Token-level augmentations can greatly contribute to existing vector-level ones. Feature-level augmentations do not affect the length of the text sequence and are therefore susceptible to length bias. ESimCSE \cite{esimsce} showed that simple word repetition augmentation can effectively counteract it. The authors also tested the insertion of the \texttt{[MASK]} token with small improvements, while the insertion of stop words slightly decreased the effect.

A better distinction of negatives can also be obtained by token augmentations. The authors of \cite{snsce} proposed Bidirectional Margin Loss (BML) to incorporate soft-negative samples that are generated using a simple rule-based method. According to its dependency syntax tree, the positive sentence is converted to its negation with correct syntax and clear semantics. The proposed setup saw improvements in the semantic textual similarity tasks from SentEval.

One drawback of token-level augmentations is the chance of producing false positives. It is especially risky in stochastic modifications, such as back-translation, deletion, or insertion, where exact output cannot be controlled. Furthermore, token-level augmentations are more complicated than feature-level ones, and thus they cannot be performed on-the-fly during the training and must be prepared in advance.

\paragraph{Positives by Relative Placement} 

Similar to the older idea of skip-thoughts \cite{NIPS2015_f442d33f}, the placement of sentences in a text can be exploited. In such a setting, text segments near or overlapping each other in the long text should also be near each other in the sentence embedding space. Therefore, text parts near the same anchor in the text can be regarded as positive, while further away as negative pairs for contrastive learning. DeCLUTR \cite{giorgi-etal-2021-declutr} for each document in a mini-batch samples multiple anchor spans. The corresponding positives are further sampled for each anchor and can be partially overlapping, adjacent, or subsumed by the anchor. This is accomplished by sampling the anchor span length to be mostly longer than the positive spans. At the time, DeCLUTR showed improvements in SentEval \cite{conneau-kiela-2018-senteval} transfer tasks in an unsupervised setting. Location exploitation is a simple approach, but it has some drawbacks. As the authors used the dataset consisting of documents of at least 2048 tokens in length, it clearly reveals that sentence placement methods have limited applications for short text domains. Moreover, random spans of DeCLUTR are subject to fragmentation in semantics.

\paragraph{Positives by Two Networks} 

The augmented version of the sample can also be constructed using two encoders. Different weight initialization and training data order are the most common causes of fluctuations in neural network performance. Although the performance is generally comparable, each model learned differently represents a different local minimum. Such a subtle difference can act as an augmentation. One of the first such setups was the CT model \cite{carlsson2020semantic}. The simple objective required two models to retain similar representations for identical sentences while distinguishing their representations from different ones.

The authors of \cite{kim-etal-2021-self} presented an approach of two BERT models, one fixed and the other tunable. The authors imposed a contrastive tension between the \texttt{[CLS]} token of the tunable model and a holistic representation of the fixed one. The latter is aggregated from all BERT layers, thus encouraging the tunable model to learn to concentrate all the relevant information to the \texttt{[CLS]} token.

The authors of \cite{pcl} proposed using multiple diverse positives instead of the usual two. The authors claim that this increases the probability of ``at-least-one'' effective positive during training. To implement diverse positives, the agreement of two similarity distributions of samples between two encoder models had to be maximized.

Another setup incorporating two networks can be used for Knowledge Distillation (KD). In \cite{Disco_DistilCSE}, after the regular KD step with the original pre-training task, an additional contrastive pre-training was added. The pair of positives for a single text segment consisted of a representation from the teacher model and another from the student. After the third stage of fine-tuning on semantic textual similarity tasks, \cite{Disco_DistilCSE} showed that the 110 M model outperformed the one with 11B parameters.

MoCoSE \cite{2202.13093} combines both feature-level augmentations and two branches based on asymmetric BERT encoders. While the online branch is updated through the loss gradient, the second, the target branch, is updated by the Exponential Moving Average (EMA). These discrepancies between the two branches prevented the model from collapsing and allowed the achievement of competitive results in SentEval.

Although two-network approaches are conceptually similar, they require extra memory or time, which is a big drawback. Transformer models already require the latest state-of-the-art GPU with the largest available memory. Techniques such as gradient accumulation are often used to process large mini-batches sequentially, but in the contrastive learning approach, the requirement for large in-batch negatives gives additional complexity. 

\paragraph{Positives and Negatives from Supervised Data}

Up to now, we have described various augmentation techniques that modify existing unlabeled samples for use in contrastive learning. However, there are several labeled datasets that can also be turned into positives or negatives by reusing the label information.

One of the first works to implement this idea was SimSCE \cite{gao-etal-2021-simcse}. The authors analyzed six candidate datasets. They found that the Natural Language Inference (NLI) datasets, SNLI \cite{bowman-etal-2015-large} and MNLI \cite{williams-etal-2018-broad}, together performed the best. The combined dataset contains sentence pairs in the form of \texttt{(premise, hypothesis, label)}, labeled as entailment, neutral, or contradiction, 314k for each class. The SimSCE authors constructed positive pairs from entailment samples and negatives using contradiction ones. The resulting supervised approach achieved state-of-the-art results at the time for SentEval tasks. The authors also tried to incorporate neutral pairs as less weighted negatives but did not observe improvements.

The authors of \cite{zhang-etal-2021-pairwise} note that the use of NLI labels to construct positive and negative pairs can contradict apparent semantic information. Elements of the negative pair may not be negative in semantic space. To address this issue, the authors train the PairSupCon model by incorporating an instance discrimination objective, which is claimed to have an implicit grouping effect. The objective discriminates both hypothesis and premise sentences from the positive pair separately from all other sentences in the batch. The authors also incorporated importance weighting on negatives to facilitate the better effect of hard ones. In total, the overall loss consisted of two instances of discrimination (one for hypothesis and the other for premise) with negative weighting and a third cross-entropy predicting NLI labels. Significant improvements were shown for clustering tasks, while for semantic textual similarity tasks in SentEval, only moderate gains were shown.

A novel use of NLI dataset labels was implemented in the PairSCL \cite{pair_level} model. First, the pair of hypothesis and premise sentences from the NLI dataset is passed through an encoder, and the unified representation of such pair is aggregated by a cross-attention module. Then a supervised contrastive loss is applied between positive and negative samples. What is interesting is that positives are regarded as hypothesis and premise pairs from the same class, i.e., whether contradiction, neutral, or entailment. Pairs labeled in different classes are considered negatives. In addition to such supervised contrastive loss, an additional cross-entropy loss is applied that predicts the actual class label. This setup showed improvements in both the NLI and SentEval transfer tasks.

ST5 \cite{sentenceT5} adopted the two-stage training strategy. In the first, 2B mined question-answering data from Community QA sites were used, framing the question and answer into the positive pair. During the second stage, NLI data were used, similar to SimSCE \cite{gao-etal-2021-simcse}. Such a more data-rich training allowed the authors to outperform the previous approaches for the SentEval task. Furthermore, using larger models with up to 11B parameters brought even further gains.

Supervised datasets are difficult to produce. If the domain of the task in question differs from publicly available NLI datasets, the easiest solution is to use unsupervised methods. Otherwise, labeled sources are indispensable. Exploiting multiple datasets and label information can make the benefits even more obvious.

\paragraph{Direction of Contrastive Learning for NLP}

The classic setup of contrastive learning is becoming more nuanced. The usual setup of two positives and multiple in-batch negatives (just any other sentence) is not perfect. As a result, the exact distance to the negatives or between the positives varies, and learning produces only a coarse approximation. The distinction between soft, hard, and weighted negatives begins to arise \cite{snsce, gao-etal-2021-simcse, zhang-etal-2021-pairwise}, as well as the use of multiple positives \cite{pcl}. The classic NT-Xent (also called InfoNCE) loss is thus often modified or additional losses added to incorporate finer training signals. The desire to scale models and datasets is also observed, as in \cite{sentenceT5}, yet we think that more memory-efficient approaches, as in \cite{Disco_DistilCSE}, should be prioritized. Nonetheless, contrastive learning currently produces state-of-the-art sentence embeddings.

\section{Methods}\label{methods}

Our extensive literature review allowed us to see the big picture of the sentence embedding research.

Currently, the evolution of models for sentence embeddings and related NLP tasks is settled at transformers. In this work, we use existing models to source the raw, token-level embeddings. In Section \ref{methods_models}, we describe the main model that we use in detail, some baselines we used for comparisons, as well as some of our original extensions. In particular, in T1--T4 models, we extend original prompting templates by incorporating more than one \texttt{[MASK]} token. Next, we present a new Avg. model where we derived sentence embeddings first by averaging tokens in different contexts and then by averaging the tokens themselves. Finally, we present our BERT + Avg. model, which combines both contextual and multiple-context averaged representations, all derived from the same BERT transformer model.

In addition to these extensions, we found two main directions that can be used to improve sentence embeddings from transformer models: token aggregation and post-processing techniques. Note, however, that our main contribution here is not the methods, as we reuse most of them from the existing works, but the combinations of them on the transformer model and extensive evaluation on multiple tasks. We thoroughly describe the techniques used (and minor extensions) for token aggregation in Section
\ref{methods_aggregating_tokens} and post-processing of embeddings in Section \ref{methods_post_processing}.

We have noticed that most works confine themselves to a small subset of evaluation tasks, which limits their results' comparability to others. Papers from top conferences always include classification tasks on top of semantic textual similarity, which is usually the only evaluation. In this work, we evaluate sentence embeddings on three different types of tasks: semantic textual similarity, downstream classification, and clustering. We present these tasks in Section \ref{evaluation}.

We now proceed with the formal definition of the problem we address in this work.

\subsection{Problem Formulation}

We are given a text sequence $s_{0}$, which was tokenized into $N$ individual pieces (i.e., tokens)
\begin{equation}
\begin{array}{cccc}
t_{1} & t_{2} & \ldots & t_{N}
\end{array},
\end{equation}
and a model representing each token as a $d$-length vector $v \in \mathbb{R}^{d}$ in each of its $L$ layers. We also use static token embeddings denoted as the -1st layer and input embeddings denoted as the 0th layer. They are sums of token, segment, and position vectors, with layer normalization applied on top. Therefore, we obtain the following 3-dimensional representation space $\mathbb{R}^{(L+2) \times N \times d }$:
\begin{equation}
\begin{array}{cccc}
   v_{1}^{-1} &v_{2}^{-1} &\ldots & v_{N}^{-1} \\
   v_{1}^{0} & v_{2}^{0} &\ldots & v_{N}^{0} \\
   v_{1}^{1} & v_{2}^{1} &\ldots & v_{N}^{1} \\
   \vdots & \vdots & \ddots & \vdots\\
   v_{1}^{L} & v_{2}^{L} &\ldots & v_{N}^{L} \\
\end{array}.
\end{equation}
{We want}  to find such an aggregation function $f$ that the $\left((L+2) \times N \times d\right)$-dimensional output from a chosen transformer model would be reduced to as meaningful a $d$-dimensional vector as possible in $\mathbb{R}^{d}$.

As a baseline for the \texttt{bert-base-uncased} model with $L=12$ layers, we consider the average over the $N$ tokens in text and over the first and last layers.
\begin{equation}
f_{\text{first+last}}(s_{0}) = \sum_{n=1}^{N}\frac{ v_{n}^{1}+v_{n}^{12}}{2N}.
\end{equation}

\subsubsection*{Additional Context Data}

Generally, a corpus has $S$ number of text samples. Therefore, other text sequences can also be used to derive the representation of the target sentence. In this way, our aggregation function $f$ operates on an $\left(S \times (L+2) \times N \times d\right)$-dimensional output from the model. $S$ can also be enlarged by using additional datasets. In particular, we used Wikitext-2 from \cite{merity2017pointer}. This can be used for various post-processing techniques such as centering, standardization, PCA, and others, where transformations are learned on corpora other than the target corpus.

\subsection{Models}\label{methods_models}

We use multiple text representation methods, focusing mainly on BERT-based ones. Prompting method T4, Averaged BERT (Avg.), BERT + Avg., B2S-100, and Random embeddings (RE) are our proposed models or modifications, while BERT, T0, and B2S are plain adaptations of existing ones. We tested token aggregation and sentence representation post-processing techniques on all eight models. We will describe them in more detail below.

\subsubsection{BERT}

BERT is a very popular transformer model. When first introduced in \cite{devlin-etal-2019-bert}, it spectacularly outperformed multiple other models on a wide range of tasks.

BERT is based on the original transformer architecture of \cite{NIPS2017_3f5ee243}. The main difference is that instead of sequence-to-sequence workings and encoder--decoder structure, it is composed only of an encoder side. This allows it to solve multiple classification and regression tasks, maintaining the same pre-trained base while only changing the classifier heads on top.

In this work, we use the BERT version called \texttt{bert-base-uncased}, a 110 M parameter model containing 12 layers (blocks) and working with lowercase text. We employ the Hugging Face implementation \cite{wolf-etal-2020-transformers} of BERT. It was pre-trained on BooksCorpus (800 M words) \cite{7410368} and English Wikipedia (2500 M words) datasets, which take 13 GB of plain text combined. The model was trained for 40 epochs (or passes through the corpus). Back then, the training time, model size, and data used were considered to be very large, yet now they are only a small fraction of what the current state-of-the-art models use.

\paragraph{Input} 

Input to the BERT encoder consists of token, positional, and token type embeddings. BERT uses WordPiece tokenization \cite{wu2016googles}. Splitting less frequent words into sub-words (e.g., ``transformer'' $\rightarrow$ ``transform'' and ``\#\#er'') rather than splitting everything on word boundaries allows one to reach a manageable vocabulary size of 30,000 tokens. BERT, like other transformer models, perceives input as a set; therefore, order information must be supplied in addition. To achieve this, additional positional embeddings are used, with a unique value for each position in a token sequence and feature dimension. Token type embeddings allow the model to distinguish between the two (if there are two) separate sentences (e.g., $\langle$ Question, Answer $\rangle$) in one token sequence. Overall, token, positional, and token type embeddings are added, and then layer normalization \cite{ba2016layer} and dropout \cite{JMLR:v15:srivastava14a} are applied. This results in the input to the BERT model blocks (for simplicity, we will call them layers).

There are 3 special tokens. \texttt{[CLS]} starts every token sequence, \texttt{[SEP]} ends token sequences and acts as a separator in a pair of sentences, and \texttt{[MASK]} is used during pre-training to mask some percentage of the input tokens at random for the model to predict (known as a ``Cloze'' task \cite{cloze}). The \texttt{[CLS]} token is also used for the next sentence prediction task---predicting whether the second sentence in a pair actually follows the first one in the training dataset or is a random one---the second unsupervised pre-training task of BERT.

\paragraph{Transformer Block}

Each transformer block (or layer) has two sublayers. The first is a multi-head self-attention mechanism and the second is a position-wise fully connected feed-forward network. The output of each sublayer \texttt{Sublayer(x)} is added to the input \texttt{x} that bypasses the sublayer through residual connection \cite{7780459} and the resulting signal ends in layer normalization \cite{ba2016layer}, \texttt{LayerNorm(x + Sublayer(x))} becoming the input to the second sublayer or the next layer.

\paragraph{Multi-Head Self-Attention} 

An input tensor $X \in \mathbb{R}^{b \times t \times f}$ with $b$ samples in a batch, $t$ tokens, and $f$ features is first linearly projected into queries $Q \in \mathbb{R}^{b \times t_{Q} \times h \times f_{Q}}$, keys $K \in \mathbb{R}^{b \times t_{K} \times h \times f_{K}}$, and values $V \in \mathbb{R}^{b \times t_{V} \times h \times f_{V}}$ tensors with $h$ heads using learned weights. Note that in the standard case, $t=t_{Q}=t_{K}=t_{V}$ and $f=hf_{Q}=hf_{K}=hf_{V}$. Then, for each sample text in a batch and each head dot, products between feature vectors of every combination of query and key tokens are calculated, resulting in an attention tensor $(QK) \in \mathbb{R}^{b \times h \times t_{Q} \times t_{K}}$. It is then scaled by dividing it by $\sqrt{f_{K}}$, and softmax is applied along $t_{K}$ so that all the dot products for any given query token and all the key tokens would sum to 1. Now, a dot product is calculated again for the probability-like scores and values tensor, resulting in a weighted selection of V, with the form of $\mathbb{R}^{b \times h \times t_{Q} \times f_{V}}$. Finally, the heads are concatenated back and a linear transformation layer $W^{O} \in \mathbb{R}^{hf_{V} \times f}$ is applied. The resulting tensor is the same shape as the input one: $\mbox{MultiHead}(X) \in \mathbb{R}^{b \times t \times f}$. %

For a more detailed explanation of the BERT and transformer architecture, please read the original papers \cite{devlin-etal-2019-bert, NIPS2017_3f5ee243}.

\subsubsection{Prompting Method (T0, T4)}

The classical use of the BERT model involves two steps: (1) a general pre-training on a very large corpora with unsupervised tasks, and (2) a supervised fine-tuning step on a small target task dataset. The first step is expensive, it requires a lot of data, time, and computing resources, but it is carried out only once. Weights that are produced in an unsupervised way contain a lot of useful representations for the fine-tuning to only perfect them.

Prompt-based methods take advantage of the first step and can completely avoid the fine-tuning stage. The idea is to frame the target task in the original pre-training format, for which the model is essentially optimized. To extract a general sentence representation, the authors of \cite{jiang-etal-2022-promptbert} proposed using ``\texttt{This sentence: "[X]" means [MASK]}'' template (which we will call T0), where the target sentence is placed instead of \texttt{[X]} and the final layer representation of \texttt{[MASK]} is used. This way, the model itself tries to predict the meaning of a given sentence, and we can extract prediction weights before they are turned into probabilities over tokens.

During initial experiments, we manually searched for other more complicated templates than T0, presented in Table~\ref{templates}. We will also use the T4 template, which is much longer and has 3 \texttt{[MASK]} tokens that have to be averaged.

\begin{table}[h]\small
\caption{Manual template search by adding additional text and [MASK] ([M]) tokens. Target sentences are inserted into the place of [X]. Only the average of 12th layer [M] representations are used. Bolded results are the best.}
\label{templates}
\begin{tabular}{l p{8.6cm} r r r} %
\toprule
\textbf{No.} & \textbf{Template} & \textbf{STS} & \textbf{Clust.} & \textbf{Class.} \\
\midrule
T0 & This sentence: "[X]" means [M]. & 63.4 & 45.5 & \textbf{78.7} \\
T1 & This sentence: "[X]" means [M][M]. & 63.4 & 46.6 & 78.6 \\
T2 & This sentence: "[X]" means "[M][M]" and is about [M]. & \textbf{70.4} & 52.8 & 78.3 \\
T3 & This sentence from the paraphrase dictionary: "[X]" means "[M]", which is about [M]. & 69.6 & \textbf{54.2} & 77.4 \\
T4 & This sentence from the dictionary: "[X]" means "[M]" and is about [M], which is a synonym for [M]. & 69.3 & 54.2 & 76.8 \\
\bottomrule
\end{tabular}
\end{table}

\subsubsection{Averaged BERT (Avg.)}

We follow the idea of \cite{bommasani-etal-2020-interpreting} to average the representations of a token in its different contexts to acquire its static embedding. Yet, we take it even further and construct vectors for sentences by again averaging such static token embeddings over the sentences.

We construct static tokens using the Wikitext-103 \cite{merity2017pointer} dataset and use the same tokenization as of BERT version \texttt{bert-base-uncased}. For each of all 28,807 tokens occurring in the Wikitext-103, we sum all vectors produced by the BERT model (which differ due to different contexts) and divide them by their count. Note that some tokens occur very often, while others are rare (for example, the most frequent, ``the'', is repeated 6,470,356 times). The process is repeated for every BERT layer, as well as combinations of layers, so that we can observe the performance dependence on this factor as well.

\paragraph{Combining Averaged BERT and Regular BERT (BERT + Avg.)} 

We also wanted to see how sentence embeddings derived from static averaged BERT tokens can contribute to the original BERT representations. Therefore, we averaged sentence embeddings from the two methods mentioned above. 

\subsubsection{BERT2Static (B2S, B2S-100)}\label{b2s}

There are more advanced distillation methods than simple BERT token averaging. For example, the authors of \cite{gupta-jaggi-2021-obtaining} trained static word vectors using BERT's contextualized representations. They adapted the Sent2Vec \cite{pagliardini-etal-2018-unsupervised} model for words and trained it to predict the word given the context element of it by the contextual representation of the BERT model. This way, the authors obtained vectors for the 750,000 most frequent words. They also showed that this approach results in vectors that perform better than existing static embeddings trained from scratch, while still enjoying a small memory and computational footprint.

In our experiments, we used the \texttt{bert\_12layer\_para} model (downloaded from %
\url{https://github.com/epfml/X2Static}, provided in \cite{gupta-jaggi-2021-obtaining}, accessed on 1 May 2023), which was trained from the contexts of a paragraph rather than only a sentence. We evaluated two versions of BERT2Static: a regular one (B2S) and one with the 100 most frequent words filtered out (B2S-100).

\subsubsection{Random Embeddings (RE)}

Instead of using a complex model to derive embeddings for words, we also tested random vectors as token embeddings. More specifically, we assign each token a random vector drawn from the normal (Gaussian) distribution, centered at 0 with 0.1 standard deviation. To facilitate better comparisons with the BERT model, we use the same tokenizer from the BERT version \texttt{bert-base-uncased} and make the vectors the same 768-size length. The whole sentence representation is then computed as the average of its constituent token vectors.

\subsection{Aggregating Tokens}\label{methods_aggregating_tokens}

Every text contains multiple tokens, each with a corresponding embedding vector. To obtain one vector for the whole text out of the many, usually, a simple average is calculated, as discussed in Section~\ref{composing}. We use it as a baseline here for multiple methods.

We investigate different methods of token weighting and filtering based on their frequencies. In classical bag-of-words approaches, it is usually accounted for with the tf-idf weighting. We, however, use only idf weighting, because the same tokens in BERT cannot be simply counted due to different contextualization. Given a dataset with $N$ documents and document frequency $\mbox{df}_t$, defined to be the number of documents in the given dataset that contain a token $t$, we calculate $\mbox{idf}_t$ for a token $t$ as 
\begin{equation}
\mbox{idf}_t = \log {\frac{N}{\mbox{df}_t}}.
\end{equation}
{To account} for long/short text differences, we scale $\mbox{idf}_t$ token weights for each text so that they would sum up to 1. This way, to calculate the average embedding for a given text, one need only sum up the weighted vectors. We calculate two inverse document frequencies: $\mbox{idf}_t^W$ for the Wikitext-2 dataset from \cite{merity2017pointer} and $\mbox{idf}_t^T$ for all samples from the given target task. 

We also adopt the method in \cite{yan-etal-2021-consert, jiang-etal-2022-promptbert} to drop the most frequent tokens. We choose 33 tokens, which were depicted in the appendix of \cite{jiang-etal-2022-promptbert}. Following that article, we also investigate dropping punctuation and subword tokens. We found that dropping all three parts---the most frequent, punctuation, and subword tokens---has the best effect, and following the original work, we name such token aggregation as with removed biases (``-biases'').

\subsection{Post-Processing Embeddings}\label{methods_post_processing}

It is common in machine learning to standardize datasets, as most methods are designed to work best with normally distributed data: Gaussian with zero mean and unit variance. Yet it is not that trivial, as data may not follow a smooth distribution and may contain disturbing outliers. %
In the context of the best practices reviewed in Section~\ref{reshaping}, we investigate the following processing methods of embeddings.

\subsubsection{Z-Score Normalization}
The most basic is the standard score, also called the z-score. It is the number of standard deviations $\sigma$ by which the value $x$ of a raw score is higher than or below the mean value $\mu$ of the raw scores of all samples. We normalize our vectors to have z-score = 1:
\begin{equation}
z = \dfrac{x-\mu}{\sigma}.
\end{equation}

\subsubsection{Quantile Normalization to Uniform Distribution (quantile-u)}
This technique works by making two distributions identical in statistical properties, thus reshaping given data values according to some known distribution function. In our case, we found that the uniform distribution worked very well as a reference. For a more detailed description of the technique, see \cite{10.1093/bioinformatics/19.2.185}.

We also tried other methods that are more robust to outliers; however, we found their performance marginally below quantile-u. This includes quantile normalization using a normal distribution and RobustScaler (as it is called in the scikit-learn preprocessing library \cite{scikit-learn}, which we used). The latter method removes the median value instead of the mean and scales the data according to the selected quantile range.

\subsubsection{Whitening}
It is a transformation that produces uncorrelated components, each with a variance of 1.

\subsubsection{All-But-The-Top (ABTT)}%

It is a method introduced in \cite{mu2018allbutthetop}. We start with the given embedding matrix $A \in \mathbb{R}^{b \times f}$ of $b$ sentences, each with $f$ features. First, it is centered by its mean $\mu \in \mathbb{R}^{f}$ into $\tilde{A} \in \mathbb{R}^{b \times f}$. Using the centered $\tilde{A}$, and given the number $d$ of the top principal components to remove, we then calculate PCA components $U \in \mathbb{R}^{d \times f}$. Now, we project our data into these components to acquire $ A^{PCA}\in \mathbb{R}^{b \times f}$:
\begin{equation}
A_{bf}^{PCA} = \sum_{d}\tilde{A}_{bf}U_{df}.
\end{equation}
{The final} processed embedding matrix will be:
\begin{equation}
A' = \tilde{A}-A^{PCA}.
\end{equation}
{The only} difference of our approach from the original authors of \cite{mu2018allbutthetop} is that instead of post processing  words, we use all-but-the-top to post-process documents.

\subsubsection{Normalization}
We also experiment with normalization: scaling individual sentence vectors to have a unit norm.

\subsubsection{Learning Post-Processing}
Some target task datasets can also be too scarce to calculate accurate statistics, such as mean, standard deviation, or others, used in post-processing. Therefore, instead, we also experiment with learning these weights on the Wikitext-2 dataset. We indicate such techniques with a superscript $\cdot^{W}$.

\subsection{Evaluation}\label{evaluation}

We evaluate the investigated methods on multiple clustering, semantic textual similarity, and classification tasks.

\subsubsection{Clustering Tasks}\label{cluster_data}

We assess the performance on 6 benchmark datasets for short text clustering. Compared to the usual ones, short datasets impose a challenge due to the weak signal caused by sparsity, which is a big problem for classic count-based approaches such as bag-of-words or tf-idf. Table~\ref{tab:clustering_dataset} provides an overview of the main statistics, and the details of each dataset are as follows.

\begin{table}[h]\small
\caption{Dataset statistics for the short text clustering datasets. N is the number of text samples, C is the number of clusters, L/S is the imbalance number defined as the size of the largest class divided by that of the smallest class, $\|V\|$ is vocabulary size, Len is the average number of tokens in each text sample, Alpha is the percent of tokens that are alphabetic (all token characters must be defined in the Unicode character database as ``Letter'', while tokens with numbers, punctuation, or BERT continuation tokens such as ``\#\#ing'' are excluded). Statistics with plain word tokenization are marked with ``W'' and with \texttt{bert-base-uncased} model tokenizer as ``B''.}
\label{tab:clustering_dataset}
\setlength{\tabcolsep}{3.2mm}
\begin{tabular}{lrrrrrrrrr}
\toprule
\multirow{2.5}{*}{\textbf{Dataset}} & \multirow{2.5}{*}{\textbf{N}} & \multirow{2.5}{*}{\textbf{C}} & \multirow{2.5}{*}{\textbf{L/S}} & \multicolumn{2}{c}{\boldmath{$\|V\|$}} & \multicolumn{2}{c}{\textbf{Len}} & \multicolumn{2}{c}{\textbf{Alpha, \%}} \\
 \cmidrule(r){5-6}  \cmidrule(lr){7-8} \cmidrule(l){9-10} 
 & & &&\textbf{W}  & \textbf{B}  & \textbf{W} & \textbf{B} & \textbf{W} & \textbf{B}   \\
\midrule
agnews & 8000 & 4 & 1 & 21,062 & 16,140 & 23 & 26 & 100 & 86 \\
biomedical & 20,000 & 20 & 1 & 18,888 & 9326 & 13 & 20 & 98 & 64 \\
googleTS & 11,109 & 152 & 143 & 19,508 & 14,763 & 28 & 33 & 100 & 85 \\
searchsnippets & 12,340 & 8 & 7 & 30,643 & 16,334 & 19 & 24 & 93 & 77 \\
stackoverflow & 20,000 & 20 & 1 & 22,909 & 7332 & 8 & 12 & 87 & 71 \\
tweet & 2472 & 89 & 249 & 5098 & 5091 & 9 & 11 & 100 & 81 \\

\bottomrule
\end{tabular}
\end{table}

\paragraph{Agnews} 
It is a subset of news titles \cite{https://doi.org/10.48550/arxiv.1502.01710}, which contains 4 topics selected by \cite{10.1007/978-3-030-51310-8_10}.

\paragraph{Biomedical} 
It is a subset of PubMed data distributed by BioASQ ( %
{\url{http://participants-area.bioasq.org/}}, accessed on 1 May 2023), where 20,000 paper titles from 20 groups are randomly selected by \cite{XU201722}.

\paragraph{GoogleTS} 
It contains titles and snippets of 11,109 news articles related to 152 events \cite{7498276}. We use the full version of the dataset, which includes both titles and snippets, named GoogleNews-TS in \cite{10.1007/978-3-030-51310-8_10}. 

\paragraph{Searchsnippets} 
It is extracted from web search snippets and contains 12,340 snippets associated with 8 groups \cite{10.1145/1367497.1367510}.

\paragraph{Stackoverflow} 
It is a subset of the challenge
data published by Kaggle (%
{\url{https://www.kaggle.com/competitions/predict-closed-questions-on-stack-overflow/data}}, accessed on 1 May 2023), where \cite{XU201722} selected 20,000 question titles associated with 20 different categories.

\paragraph{Tweet} 
It consists of 89 categories with 2472 tweets in total \cite{7498276}.

\paragraph{} 
We perform clustering by running $k$-means with the scikit-learn \cite{JMLR:v12:pedregosa11a} package and reported the clustering accuracy, computed using the Hungarian algorithm \cite{10.2307/2098689} and averaged over 10 independent runs ({we used the codebase from %
\url{https://github.com/amazon-science/sentence-representations} and downloaded clustering datasets from %
\url{https://github.com/rashadulrakib/short-text-clustering-enhancement/tree/master/data}, both were accessed on 1 May 2023}).

\subsubsection{Semantic Textual Similarity (STS) Tasks}\label{sts_data}

STS assesses the degree to which two sentences are semantically equivalent to each other. A single sample consists of two sentences and a score ranging from 0 for no meaning overlap to 5 for meaning equivalence. The semantic textual similarity shared task has been held annually since 2012 up to 2017 \cite{agirre-etal-2012-semeval, agirre-etal-2013-sem, agirre-etal-2014-semeval, agirre-etal-2015-semeval, agirre-etal-2016-semeval, cer-etal-2017-semeval} and formed STS12-STS17 datasets. A total of 8628 carefully collected samples from these contests formed the STS benchmark \cite{cer-etal-2017-semeval}. Table~\ref{tab:class_dataset} shows details of the STS datasets, including SICK-Relatedness \cite{marelli-etal-2014-sick} and STS-B test sets, which we also use. Similarly to \cite{wang2022just}, we also add (STR) \cite{abdalla2023makes}, a recent semantic relatedness dataset created by comparative annotations.

STS is a very popular choice for evaluating textual embeddings in an unsupervised way. Without any fine-tuning, one can calculate the distance (usually cosine) between two vectors of two sentences, which should correlate to the target equivalence score. It is so widely adopted that almost all work on semantic representations assesses the model performance on this task.

\begin{table}[h]\small
\caption{Dataset statistics for STS and classification tasks. N is the number of text samples, C is the number of clusters, L/S is the imbalance number, defined as the size of the largest class divided by that of the smallest class (for STS tasks, the two classes are binned to 1 point label value length ones from the highest and lowest sides), %
$\|V\|$ is vocabulary size, Len is the average number of tokens in each text sample, Alpha is the percent of tokens that are alphabetic. Statistics with plain word tokenization are marked with ``W'' and with \texttt{bert-base-uncased} model tokenizer as ``B''. For MRPC, SICK-R/E, STS-B, and STS tasks, statistics are calculated for tokenized and then concatenated sentences in each pair.}
\label{tab:class_dataset}
\setlength{\tabcolsep}{2.45mm}
\begin{tabular}{llrrrrrrrrr}
\toprule
\multirow{2.5}{*}{\textbf{Dataset}} & \multirow{2.5}{*}{\textbf{Split}}  & \multirow{2.5}{*}{\textbf{N}} & \multirow{2.5}{*}{\textbf{C}} & \multirow{2.5}{*}{\textbf{L/S}} & \multicolumn{2}{c}{\boldmath{$\|V\|$}} & \multicolumn{2}{c}{\textbf{Len}} & \multicolumn{2}{c}{\textbf{Alpha, \%}} \\
 \cmidrule(r){6-7}  \cmidrule(lr){8-9} \cmidrule(l){10-11}
 &  &  & &  & \textbf{W} & \textbf{B} & \textbf{W} & \textbf{B} & \textbf{W} & \textbf{B} \\
 \midrule
\multicolumn{5}{l}{STS tasks}  \vspace{1mm}\\
STS12 &  & 3108 &  & 5.2 & 8127 & 7802 & 25 & 28 & 83 & 77 \\
STS13 &  & 1500 &  & 0.7 & 5152 & 5141 & 20 & 21 & 88 & 82 \\
STS14 &  & 3750 &  & 1.6 & 9117 & 8613 & 21 & 23 & 86 & 80 \\
STS15 &  & 3000 &  & 0.9 & 7364 & 7185 & 23 & 24 & 89 & 85 \\
STS16 &  & 1186 &  & 1.2 & 3971 & 4175 & 26 & 28 & 87 & 83 \\
STR &  & 5500 &  & 1.0 & 22,392 & 12,883 & 25 & 32 & 83 & 76 \\
\midrule
 \multicolumn{4}{l}{Binary classification} \vspace{1mm}\\
MR &  & 10,662 & 2 & 1.0 & 20,325 & 13,802 & 22 & 26 & 84 & 76 \\
CR &  & 3775 & 2 & 1.8 & 5675 & 5221 & 20 & 22 & 84 & 80 \\
SUBJ &  & 10,000 & 2 & 1.0 & 22,636 & 15,912 & 25 & 28 & 85 & 78 \\
MPQA &  & 10,606 & 2 & 2.2 & 6239 & 6248 & 3 & 3 & 97 & 88\vspace{4pt}\\
\multirow[c]{3}{*}{SST2} & train & 67,349 & 2 & 1.3 & 14,816 & 11,570 & 9 & 11 & 88 & 78 \\
 & dev & 872 & 2 & 1.0 & 4339 & 4542 & 20 & 23 & 85 & 76 \\
 & test & 1821 & 2 & 1.0 & 7053 & 6824 & 19 & 23 & 85 & 76\vspace{4pt}\\
\multirow[c]{2}{*}{MRPC} & train & 4076 & 2 & 2.1 & 16,112 & 12,061 & 44 & 50 & 81 & 75 \\
 & test & 1725 & 2 & 2.0 & 10,092 & 8471 & 44 & 50 & 82 & 75 \\
 \midrule
\multicolumn{4}{l}{Fine-grained classification} \vspace{1mm}\\
\multirow[c]{3}{*}{SST5} & train & 8544 & 5 & 2.1 & 16,579 & 12,395 & 19 & 23 & 84 & 75 \\
 & dev & 1101 & 5 & 2.1 & 5038 & 5168 & 19 & 23 & 85 & 76 \\
 & test & 2210 & 5 & 2.3 & 7929 & 7478 & 19 & 23 & 85 & 76\vspace{4pt}\\
\multirow[c]{2}{*}{TREC} & train & 5452 & 6 & 14.5 & 9437 & 8492 & 10 & 11 & 87 & 81 \\
 & test & 500 & 6 & 15.3 & 1100 & 1247 & 7 & 8 & 85 & 79\vspace{4pt}\\
\multirow[c]{3}{*}{SCICITE} & train & 7320 & 3 & 4.4 & 27,775 & 13,603 & 31 & 40 & 100 & 77 \\
 & dev & 916 & 3 & 4.4 & 7625 & 6918 & 31 & 40 & 100 & 78 \\
 & test & 1861 & 3 & 3.8 & 12,609 & 9217 & 31 & 41 & 100 & 77\vspace{4pt}\\
\multirow[c]{3}{*}{SICK-E/R} & train & 4500 & 3/\ldots & 3.8/3.6 & 2258 & 2277 & 19 & 20 & 99 & 96 \\
 & dev & 500 & 3/\ldots & 3.8/5.0 & 1122 & 1172 & 20 & 20 & 99 & 96 \\
 & test & 4927 & 3/\ldots & 3.9/3.9 & 2271 & 2291 & 19 & 20 & 99 & 96\vspace{4pt}\\
\multirow[c]{3}{*}{STS-B} & train & 5749 & \ldots & 1.3 & 12,430 & 10,792 & 22 & 25 & 86 & 80 \\
 & dev & 1500 & \ldots  & 0.7 & 6542 & 6511 & 26 & 28 & 85 & 80 \\
 & test & 1379 & \ldots  & 1.1 & 4888 & 4921 & 22 & 24 & 85 & 81 \\
 \bottomrule
\end{tabular}

\end{table}

We follow the STS evaluation settings from \cite{gao-etal-2021-simcse}. First, the evaluation is kept unsupervised by not applying any additional regressors; cosine similarity between embeddings in a pair is taken directly as a model score for similarity. To find the degree of correlation between the annotated and model-given labels, Spearman’s rank correlation is used because it measures the rankings instead of the actual scores. Finally, for annual STS challenges, we concatenate all subsets and report the general Spearman correlation for that year (referred to in \cite{gao-etal-2021-simcse} as ``all'').  As we used the SentEval toolkit \cite{conneau-kiela-2018-senteval}, we had to implement concatenation ourselves since it only had ``mean'' and ``wmean'' settings available.

\FloatBarrier

\subsubsection{Downstream Classification Tasks}\label{class_data}

Differently from STS, these tasks are evaluated in a supervised way. Following the SentEval \cite{conneau-kiela-2018-senteval} benchmark suite, the commonly used evaluation protocol is to train a logistic regression or an MLP classifier with a cross-validation setup on top of the frozen representations, and the testing accuracy is used as a measure of the representation quality. We went after the logistic regression classifier and the 10-fold cross-validation scheme, the setting which is the most commonly reported in the literature. We evaluated various binary, ternary, and fine-grained classification as well as regression tasks. A more detailed description of each is presented below, while statistics are presented in Table~\ref{tab:class_dataset}. %

\paragraph{Binary Classification} 
It includes sentiment prediction from Stanford Sentiment Treebank dataset SST2 \cite{socher-etal-2013-recursive}, movie reviews MR \cite{pang-lee-2005-seeing}, and customer product reviews CR \cite{10.1145/1014052.1014073}. In SUBJ \cite{10.3115/1219840.1219855}, binary subjectivity status is labeled for sentences from movie reviews and plot summaries, and in MPQA \cite{wiebe2005annotating, deng2015mpqa}, phrase-level opinion polarity from news articles is predicted. The last is MRPC \cite{dolan-etal-2004-unsupervised}, the Microsoft Research Paraphrase Corpus, from parallel news sources for the paraphrase detection task.

\paragraph{Ternary Classification} 
SCICITE \cite{cohan-etal-2019-structural} is a domain-specific classification task that assigns one of three intent labels (``background information'', ``method'', ``result comparison'') to sentences collected from scientific articles citing other articles. Meanwhile, the SICK-E \cite{marelli-etal-2014-sick} dataset has labels for sentence pairs as ``entailed'', ``contradiction'', or ``neutral''.

\paragraph{Fine-Grained Classification}
It includes SST5 \cite{socher-etal-2013-recursive}, a 5-level sentiment analysis dataset, and TREC \cite{li-roth-2002-learning}, comprising classification of 6 types of questions. We also evaluate regular semantic textual relatedness and similarity tasks SICK-R \cite{marelli-etal-2014-sick} and STS-B \cite{cer-etal-2017-semeval} with classification, by splitting the real-valued similarity targets into 5 discrete class labels. For example, a [0,5] score of 3.6 goes to class 3 with weight 0.4 and also to class 4 with weight 0.6. 

\subsubsection{Isotropy}

We use the IsoScore \cite{rudman-etal-2022-isoscore} metric to measure the uniformity of the utilization of the embedded space. As shown by the authors of \cite{rudman-etal-2022-isoscore}, IsoScore has stronger properties than other isotropy measures.

For each sentence embedding method, we calculated IsoScore for the Wikitext-2 dataset from \cite{merity2017pointer}. We split the dataset into individual sentences (now samples) and omitted texts with less than 10 characters.

\subsubsection{Alignment and Uniformity}\label{align_uniform}

Explaining the success of contrastive methods, the authors of \cite{pmlr-v119-wang20k} proposed using alignment and uniformity properties to better quantify the quality of representations. Alignment is calculated between semantically related positive pairs and therefore is an expected distance between embeddings of the paired instances:
\begin{equation}
\mathcal{L}_{\text {align}}(f) =  \mathop{\mathbb{E}}_{(x, y) \sim p_{\text{pos}}}\left[\|f(x)-f(y)\|^2\right].
\end{equation}
{Uniformity} is computed using representations of the whole space:
\begin{equation}
\mathcal{L}_{\text{uniform}}(f) = \log \mathop{\mathbb{E}}_{(x, y) \overset{\text{i.i.d.}}\sim p_{\text{data}}}\left[e^{-2\|f(x)-f(y)\|^2}\right].
\end{equation}
{Smaller} values of both alignment and uniformity indicate better quality of the representations. 

As a distribution of positive pairs $p_\text{pos}$, we used STS-B task training split pairs with a similarity of 5.0, and for $p_\text{data}$, we used sentences from all pairs.

\section{Results}\label{results}

Here, we present the results of our various experiments outlined in Section \ref{methods}. 

\subsection{Token Aggregation and Post-Processing Techniques}

The results of best-performing token aggregation and embedding post-processing techniques are presented in Table~\ref{tokens_sts}. The results here are averaged over all the different tasks in the class. See Appendix~\ref{appendix1} for the individual results. 

We see that both aggregation and post-processing methods have a great impact on STS and clustering performance. For semantic textual similarity tasks, the best model was improved from 62.3 to 71.6 average Spearman correlation, while for clustering, the best model was improved from 59.2 to 64.8 average accuracy.

\begin{table}[h]\small
\caption{Effect of different token aggregation and post-processing methods on multiple text embedding models. Both correlation and accuracy scores range from 0 to 100 (from the worst to the best). We also added results for SimCSE-BERT model \cite{gao-etal-2021-simcse}, which is fine-tuned on NLI data supervision. The best result for each model is bolded, while underlined results are the best across all the models.}
\label{tokens_sts}
\setlength\tabcolsep{5.3pt}
\begin{tabular}{lrrrrrrrr}
\toprule
\textbf{Model} & \textbf{T0} & \textbf{T4} & \textbf{BERT} & \textbf{BERT + Avg.} & \textbf{Avg.} & \textbf{B2S} & \textbf{B2S-100} & \textbf{RE} \\
\cmidrule(r){2-3} \cmidrule(lr){4-6} \cmidrule(l){7-9}
\textbf{Layer} & \multicolumn{2}{c}{\textbf{Last}} & \multicolumn{3}{c}{\textbf{First + Last}} & \multicolumn{3}{c}{\textbf{--}} \\
\midrule
\multicolumn{9}{l}{Avg. Spearman correlation (\%) of STS tasks}\vspace{1mm}\\
avg. & 61.3 & 62.3 & 61.3 & 61.4 & 56.4 & 56.7 & 60.6 & 50.3 \vspace{3pt}\\
$\mbox{idf}_t^W$ & 67.8 & 67.4 & 69.2 & 69.3 & 66.9 & 66.1 & 65.4 & \textbf{66.4} \\
\quad+ zscore %
 & 68.6 & 68.1 & \textbf{69.8} & \textbf{70.6} & \textbf{69.5} & \textbf{66.4} & \textbf{65.9} & 66.4\vspace{3pt}\\
$\mbox{idf}_t^T$ & 67.4 & 69.2 & 68.7 & 69.0 & 67.5 & 56.7 & 60.6 & 64.9 \\
\quad+ quantile-u & \textbf{69.5} & \underline{\textbf{71.6}} & 68.9 & 69.5 & 68.4 & 56.2 & 61.0 & 63.0\vspace{3pt}\\
-biases & 64.7 & 66.0 & 68.2 & 69.3 & 67.3 & 54.1 & 57.5 & 63.7\vspace{3pt}\\
\verb|[MASK]| & 63.4 & 69.3 &  &  &  &  &  &  \\
\quad+ quantile-u & 65.6 & 70.5 &  &  &  &  &  &\vspace{3pt}\\
\multicolumn{9}{l}{SimCSE performance: 81.5}\\
\midrule
\multicolumn{9}{l}{Avg. accuracy (\%) of clustering tasks}\vspace{1mm}\\
avg. & 53.5 & 55.0 & 57.0 & 59.2 & 55.2 & 53.8 & 55.3 & 36.3\vspace{3pt}\\
$\mbox{idf}_t^W$ & 53.0 & 53.0 & 55.2 & 57.1 & 53.7 & 53.0 & 53.7 & 44.3 \\
\quad+ normalize & 54.3 & 54.4 & 58.0 & 57.9 & 54.1 & \textbf{55.8} & 56.1 & \textbf{49.5}\vspace{3pt}\\
$\mbox{idf}_t^T$ & 53.1 & 57.6 & 60.7 & 62.5 & 58.7 & 53.8 & 55.3 & 42.8 \\
\quad+ quantile-u$^W$ & \textbf{57.6} & \textbf{60.0} & 63.1 & 64.4 & 60.2 & 54.6 & 56.0 & 44.4 \\
\quad+ normalize & 55.5 & 57.6 & \textbf{63.4} & \underline{\textbf{64.8}} & 59.8 & 55.5 & \textbf{57.2} & 47.0\vspace{3pt}\\
-biases & 54.3 & 56.2 & 61.1 & 62.2 & 61.6 & 52.6 & 54.1 & 40.7 \\
\quad+ normalize & 54.4 & 56.3 & 62.4 & 63.4 & \textbf{62.2} & 54.4 & 56.8 & 44.6\vspace{3pt}\\
\verb|[MASK]| & 45.5 & 54.2 &  &  &  &  &  &  \\
\quad+ quantile-u & 44.4 & 54.9 &  &  &  &  &  &\vspace{3pt}\\
\multicolumn{9}{l}{SimCSE performance: 59.8 and 64.0 (avg. + quantile-u$^W$, first + last layers)}\\
\midrule
\multicolumn{9}{l}{Avg. accuracy (\%) of classification tasks}\vspace{1mm}\\
avg. & \underline{\textbf{80.5}} & 79.9 & 79.9 & 79.7 & \textbf{77.3} & \textbf{76.9} & 75.1 & 69.5 \\
$\mbox{idf}_t^W$ & 73.1 & 67.1 & 78.3 & 77.9 & 75.3 & 76.8 & 74.1 & 67.6 \\
$\mbox{idf}_t^T$ & 80.1 & 79.7 & 78.9 & 78.4 & 76.1 & 76.9 & \textbf{75.1} & 68.1 \\
-biases & 80.1 & \textbf{79.9} & \textbf{80.1} & \textbf{79.8} & 77.2 & 76.3 & 74.2 & \textbf{70.3} \\
\verb|[MASK]| & 78.7 & 76.8 &  &  &  &  &  &\vspace{3pt}\\
\multicolumn{9}{l}{SimCSE performance: 86.5}\\
\bottomrule
\end{tabular}
\end{table}

Our techniques can even improve the dedicated SimCSE model \cite{gao-etal-2021-simcse}, which was fine-tuned on NLI data. Its main strength lies in semantic textual similarity tasks, where it leads with over 10\% difference. However, for clustering tasks, its average accuracy is similar to the other evaluated models at 59.8\% and improves up to 64.0\%, if we apply the best-performing techniques. This showcases a general tendency that the top-performing models are very good only in a narrow subset of tasks and highlights the importance of our more general methods.

For maximum performance, usually, both aggregation and post-processing techniques must be used. As an example, sentences placed within the T4 template with no techniques applied have a 62.3 average Spearman correlation. With a better aggregation method of $idf_{t}^{T}$, it is pushed up to 69.2, while only using quantile-u post-processing gives 67.3 (see Table~\ref{pps} with only post-processing techniques applied). However, a combined effort of both $idf_{t}^{T}$ and quantile-u gives the average Spearman correlation score of 71.6. 

Still, the improvements from both aggregation and post-processing do not exactly add up linearly. 
This indicates that the improvements to representations of this type may saturate below the perfect scores. 

We considered many post-processing techniques for sentence vectors and show the performance of the most popular and best-performing ones in Table~\ref{pps} separately. Indeed, almost every method somehow improves the average Spearman correlation for STS tasks. For all-but-the-top (abtt), we varied the number of components to remove up to a hundred but settled on removing only the first two, as it was slightly better than the rest. Although highly credited in the literature, abtt-2 still obtained smaller scores than the others. Plainly averaging the token representations, the highest scores for both STS and clustering tasks were achieved using quantile-uniform normalization. We would also like to mention the simple normalization of vectors to the unit length, which was often beneficial for clustering tasks, and whitening normalization, which was learned on Wikitext-2 dataset, and was favorable with the random embeddings model. We trained other post-processing methods on Wikitext-2 too, yet they resulted in similar or slightly smaller scores.

\begin{table}[h]\small
\caption{Performance %
of post-processing with plain token averaging. abtt-2 is the ABTT method with top 2 principal components removed. The best result for each model is bolded, while underlined results are the best across all the models.}
\label{pps}
\setlength\tabcolsep{5.3pt}
\begin{tabular}{lrrrrrrrr}
\toprule
\textbf{Model} & \textbf{T0} & \textbf{T4} & \textbf{BERT} & \textbf{BERT + Avg.} & \textbf{Avg.} & \textbf{B2S} & \textbf{B2S-100} & \textbf{RE} \\
\cmidrule(r){2-3} \cmidrule(lr){4-6} \cmidrule(l){7-9}
\textbf{Layer} & \multicolumn{2}{c}{\textbf{Last}} & \multicolumn{3}{c}{\textbf{First + Last}} & \multicolumn{3}{c}{\textbf{--}} \\
\midrule
\multicolumn{9}{l}{Avg. Spearman correlation (\%) of STS tasks}\vspace{1mm}\\
avg. & 61.3 & 62.3 & 61.3 & 61.4 & 56.4 & 56.7 & 60.6 & 50.3 \\
\quad+ zscore %
 & \textbf{65.4} & \underline{\textbf{67.3}} & \textbf{65.1} & \textbf{66.5} & 64.1 & 58.5 & \textbf{62.1} & 55.6 \\
\quad+ quantile-u & 65.0 & 67.3 & 64.3 & 65.7 & 63.0 & 56.2 & 61.0 & 53.0 \\
\quad+ quantile-u$^W$ & 62.8 & 64.6 & 63.0 & 64.7 & 62.1 & 55.5 & 60.5 & 52.2 \\
\quad+ abtt-2 & 64.7 & 66.7 & 64.3 & 66.3 & 64.3 & 58.3 & 62.0 & 54.4 \\
\quad+ normalize & 61.3 & 62.3 & 61.3 & 61.4 & 56.4 & 56.7 & 60.6 & 50.3 \\
\quad+ whiten & 63.5 & 64.2 & 64.7 & 65.6 & 65.5 & \textbf{62.1} & 61.2 & 63.2 \\
\quad+ whiten$^W$ & 60.9 & 63.7 & 63.1 & 65.8 & \textbf{67.0} & 61.4 & 61.0 & \textbf{64.3} \\
\midrule
\multicolumn{9}{l}{Avg. accuracy (\%) of clustering tasks}\vspace{1mm}\\
avg. & 53.5 & 55.0 & 57.0 & 59.2 & 55.2 & 53.8 & 55.3 & 36.3 \\
\quad+ zscore & 54.2 & 55.7 & 57.8 & 59.5 & 56.1 & 53.3 & 55.0 & 36.3 \\
\quad+ quantile-u & \textbf{54.6} & \textbf{56.3} & 58.7 & 60.7 & 57.3 & 54.2 & 55.8 & 36.8 \\
\quad+ quantile-u$^W$ & 54.2 & 56.3 & \textbf{60.1} & \underline{\textbf{61.1}} & \textbf{58.6} & 54.5 & 56.0 & 36.9 \\
\quad+ abtt-2 & 53.4 & 54.8 & 57.0 & 59.0 & 55.8 & 53.6 & 55.0 & 36.5 \\
\quad+ normalize & 53.4 & 55.0 & 59.7 & 59.7 & 56.1 & \textbf{55.5} & \textbf{57.2} & 38.5 \\
\quad+ whiten & 35.4 & 35.7 & 33.1 & 29.8 & 26.8 & 25.3 & 26.0 & 24.8 \\
\quad+ whiten$^W$ & 49.0 & 49.8 & 56.2 & 58.7 & 55.0 & 50.4 & 50.1 & \textbf{41.5} \\
\bottomrule
\end{tabular}
\end{table}

Unlike unsupervised ones, classification tasks do not benefit from the two representation-shaping techniques. There is only negligible improvement for classification tasks, when the biases (punctuation, most frequent, and subword tokens) are eliminated. Otherwise, performance is only decreased. 

This result is logical because similarity or clustering tasks are carried out with the resulting text representations directly, while the classification is a supervised task learned on top. In the first case, the most informative components of the representations must be present in the largest principal components (i.e., constitute most of the variance in the data) for high scores, while the supervised logistic regression classifier can learn to extract them from the small principal components or ill-shaped representations on its own. 

\subsubsection{Avg. {versus}  B2S and B2S-100}

An interesting result, as seen in Table~\ref{tokens_sts}, is the comparison of simply averaged BERT tokens from many contexts (the Avg. model) and word vectors from a specially trained Word2Vec-style model BERT2Static (the B2S model) using the same BERT contexts, as described in Section \ref{b2s}. If no token aggregation and post-processing techniques are used, B2S is of similar performance in STS tasks (with average Spearman correlation of 56.7 {versus 
 56.4)}, worse at clustering (with average clustering accuracy of 53.8 versus 55.2), and comparable in classification tasks (with average B2S accuracy of 76.9 versus 77.3) to the Avg. model. As we can see, the task performance differences are very small.

Apart from training, the second main difference between the Avg. and B2S models is tokenization. The Avg. model uses sub-word tokens, the same as BERT (tokens of which it averages), while B2S models use a vocabulary of the full set of the most frequent words, 20 times the size of the BERT's. Due to this discrepancy, token aggregations for BERT and B2S models are applied differently, as idf statistics for both tokenizations are different; also, B2S does not have the equivalence of biased tokens such as BERT. However, it is very straightforward for B2S to filter out (do not use during averaging) a portion of the most frequent words. During our experiments, we varied this number and found that removing the 100 most frequent words works best, which improved the average Spearman correlation of STS tasks from 56.7 to 60.6 and the average accuracy of clustering tasks from 53.8 to 55.3 compared to full B2S, both with plain token averaging and no post-processing applied.

Despite the additional option for B2S to remove the 100 most frequent tokens, using additional token aggregation and post-processing techniques allows our simple Avg. model to surpass both B2S and B2S-100 (see Table~\ref{tokens_sts}). Note that the removal of the most frequent words for B2S works in a similar way to weighted token aggregations or post-processing, the gains are not additive. Now the best average STS Spearman correlation score becomes 69.5 for Avg. versus 66.4 of B2S and the best average clustering accuracy of 62.2 for Avg. versus 57.2 of B2S-100. Additional techniques here helped a much simpler Avg. model outperform a much more complex learned B2S.

The success of such a simple Avg. model inspired us to seek further improvement by combining it with the parent BERT model. For the combined BERT + Avg. model results, please see Section~\ref{bert_and_avg}.

\subsubsection{BERT versus Random Embeddings}\label{re_results}

The most dramatic increase in performance due to the two techniques is observed for the model of random embeddings. For STS tasks, the average Spearman correlation rises from 50.3 to 66.4, while for clustering tasks, it increases from 36.3 to 49.5. Although accuracy for the latter tasks is still very low, for STS, it is just a mere 3.4 points below what BERT managed to accomplish with both techniques applied.

If we were to look at the detailed performance on individual tasks in Appendix~\ref{appendix_agg_post}, we would find several where random embeddings with the help of the two shaping techniques score higher than the also-shaped BERT representations. For STS14 (Table~\ref{STS14}), it is 68.8 versus 68.3 Spearman correlation, while for the googleTS (Table~\ref{googleTS}), stackoverflow (Table~\ref{stackoverflow}), and tweet (Table~\ref{tweet}) datasets, the clustering accuracy is, respectively, 69.5 vs. 68.5, 70.6 vs. 59.9, and 58.5 vs. 55.1. This may indicate a smaller complexity of these particular datasets, where the task can largely be solved based on a small set of keywords. Also, if the texts do not contain the natural language of the type the BERT was pre-trained on (e.g., they contain code), the model cannot properly contextualize the tokens, and the random embeddings without contexts work better. Yet it is important to note that it is achieved only with the help of aggregation and post-processing methods on top of the random embeddings. 

The largest clustering accuracy difference between BERT and random embeddings is for the agnews and searchsnippets datasets, with the best scores of 86.8 versus 43.4 and 82.9 versus 36.6, respectively. We think that the observed performance gap may be related to the vocabulary and text sizes of the datasets. Stackoverflow and tweet datasets, with random embeddings ahead of the BERT, have the smallest vocabulary sizes of 7332 and 8091 (see Table~\ref{tab:clustering_dataset}), while agnews and searchsnippets have the largest vocabulary sizes of 16,140 and 16,334 unique tokens, respectively. Because random embeddings rely only on distinctness of tokens, with the increased amount of them, the probability of having exactly the same keywords in two texts drops. BERT embeddings having non-identical but semantically similar tokens and similar representations helps in this case. 

Having shorter text lengths (average text sizes of 12 and 11 for stackoverflow and tweet, respectively, versus 26 and 24 for agnews and searchsnippets, respectively) may also help the random embeddings, because they do not average away so easily, and there is less context to consider. 

In contrast, other static models have much more comparable scores. For example, the Avg. clustering accuracy is 83.8 (agnews) and 74.2 (searchsnippets).

\subsubsection{Isotropy}\label{isotropy_results}

Many post-processing methods have previously been proposed to improve the isotropy of representations. It was argued that representations from the BERT model fall into the narrow cone and, therefore, are anisotropic. Thus, by raising isotropy, one can improve regular task performance.

\begin{figure}[h]
{\includegraphics[width=\textwidth]{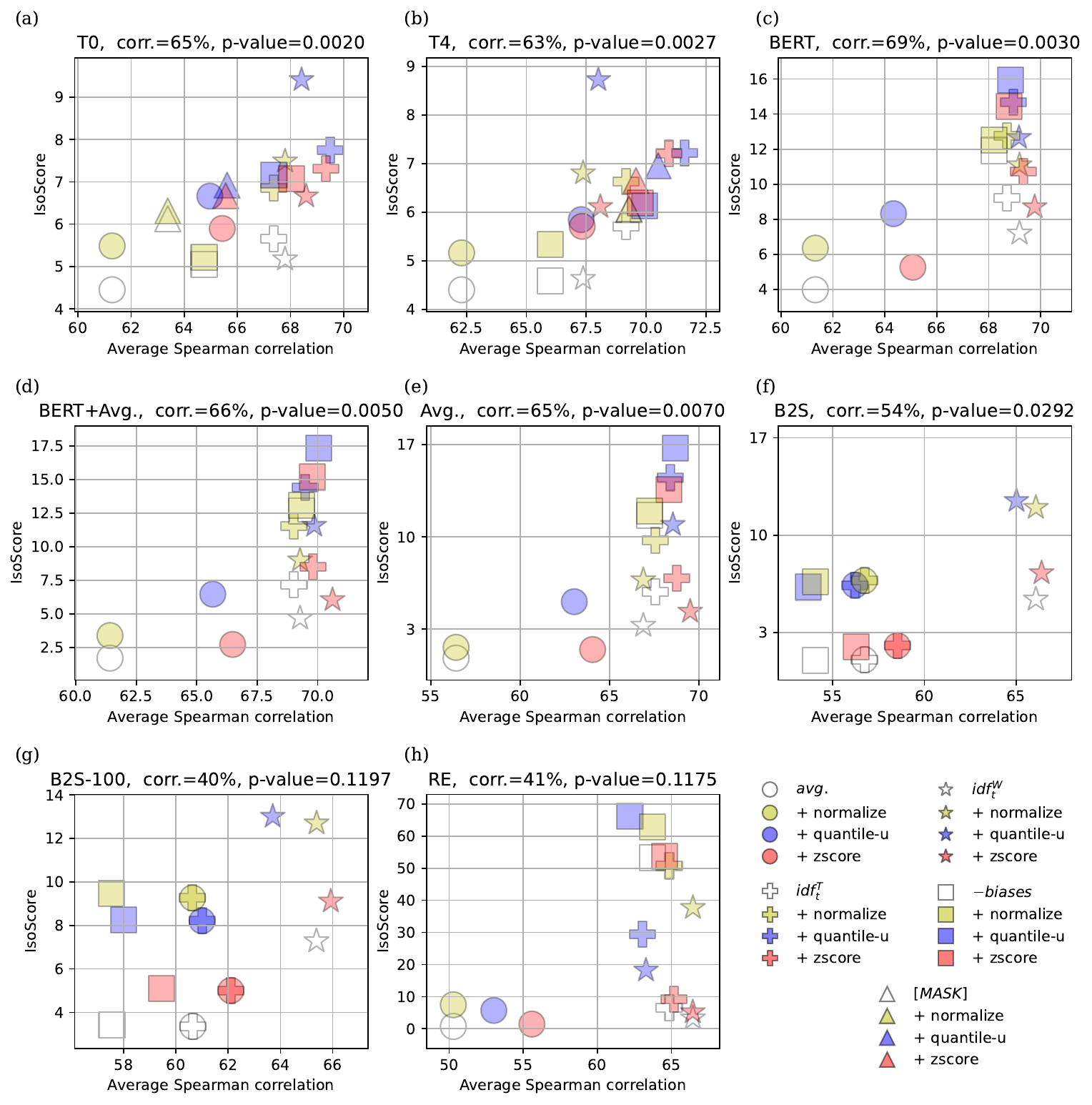}}
\caption{Relation between average Spearman correlation for STS tasks and IsoScore of Wikitext representations for each model. Pearson correlation coefficients are shown.}
\label{fig_iso1}
\centering
\end{figure}

\begin{figure}[h]
{\includegraphics[width=\textwidth]{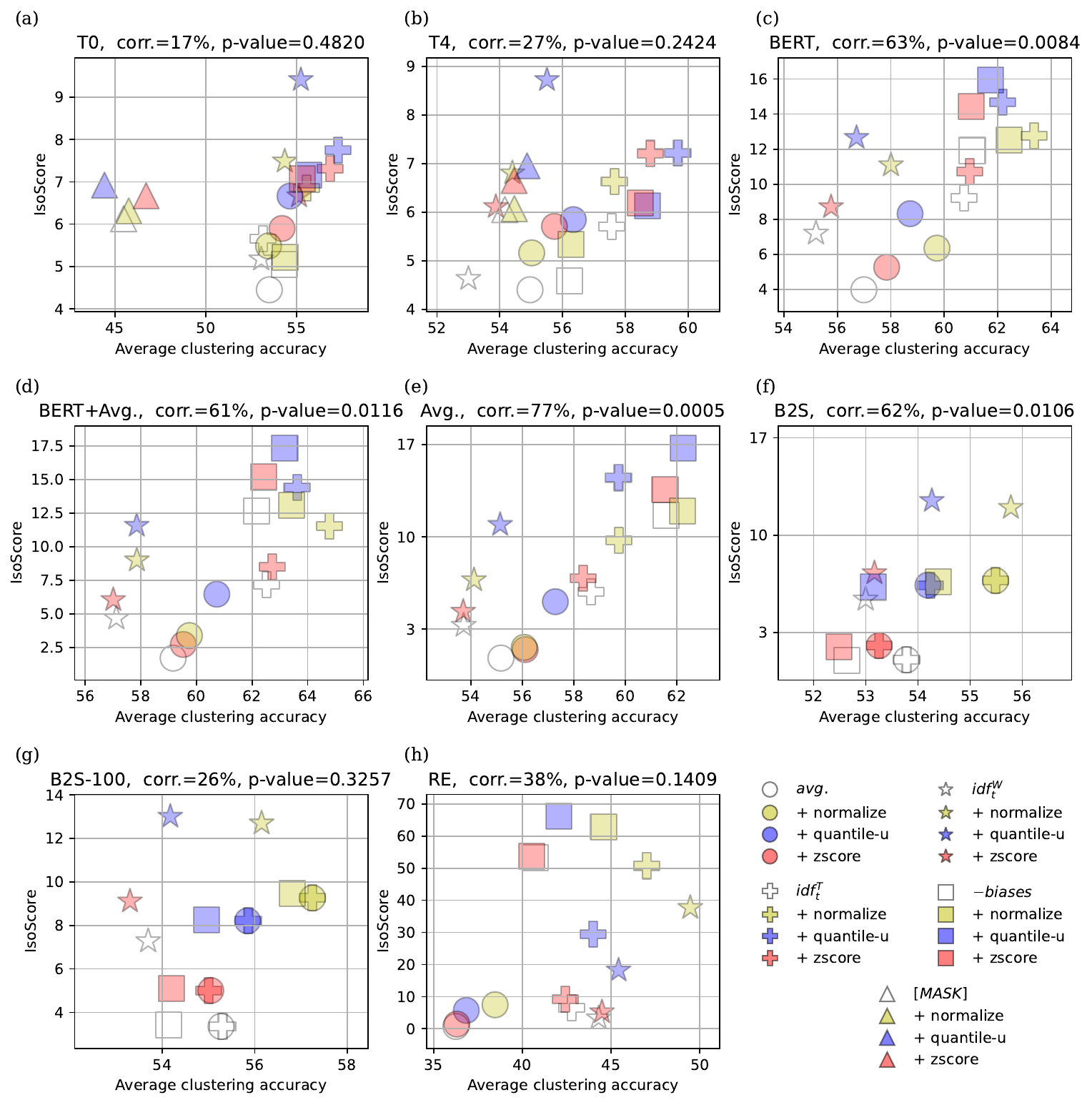}}
\caption{Relation between average clustering accuracy and IsoScore of Wikitext representations for each model. Pearson correlation coefficients are shown.}
\label{fig_iso2}
\centering
\end{figure}

We calculate the IsoScore isotropy metric on the Wikitext-2 dataset after applying various post-processing and token aggregation methods for multiple models. We exclude whitening post-processing, as it always produces representations with the IsoScore at the maximum of 100\%, just due to its working principle. Then, we compare the isotropy score to the STS, clustering, and classification task performance by calculating the Pearson correlation coefficient. For semantic textual similarity tasks, results are shown in Figure~\ref{fig_iso1}, and for clustering tasks, results are shown in Figure~\ref{fig_iso2}.

For each method, the baseline IsoScore is very low: always below 5\%. After applying weighted token aggregation techniques or post-processing, it is always increased. The smallest improvements of up to 7-8\% score are observed for templated models T0 and T4, while the highest, over 80\%, are for random embeddings. For other models, the IsoScore reaches around 17\%.

Why is the IsoScore of random embeddings improved so much compared to the other models? The answer may be related to the inner workings of RE. Once we generate the random embeddings, these token-level representations have a maximum possible isotropy. The RE model represents text sequences as averages of these vectors; thus, isotropy going from the token level to the document is reduced. However, token aggregation and post-processing techniques mitigate this isotropy loss, allowing for document embeddings to regain most of it back from the token-level ones. Other models start with already low isotropy for token-level representations and thus have less space to improve it.

\paragraph{IsoScore Correlation with Task Performance}

Six out of eight models have a moderate correlation (more than 54\%) between Isoscore and the average Spearman correlation of semantic textual similarity tasks (see Figure~\ref{fig_iso1}). For clustering (see Figure~\ref{fig_iso2}), it is less apparent, with only four models reaching moderate Pearson correlation. However, for both STS and clustering tasks, the best score for each model is always reached by some improvement of IsoScore, compared to the isotropy score of the plain averaging setup.

\vspace{-5pt}

\vspace{-12pt}

In contrast, the classification does not improve with token aggregation and post-processing techniques; therefore, we do not observe correlation (and do not show it here).

\FloatBarrier

\subsubsection{Alignment and Uniformity}

We observed that token pooling and post-processing techniques do not improve the alignment and uniformity properties of the representations. Let us remind the reader that alignment is calculated with only those STS-B pairs with a similarity score of 5, while uniformity with all pairs, as defined in Section~\ref{align_uniform}, and smaller values are better. 

\begin{figure}[h]
{\includegraphics[width=\textwidth]{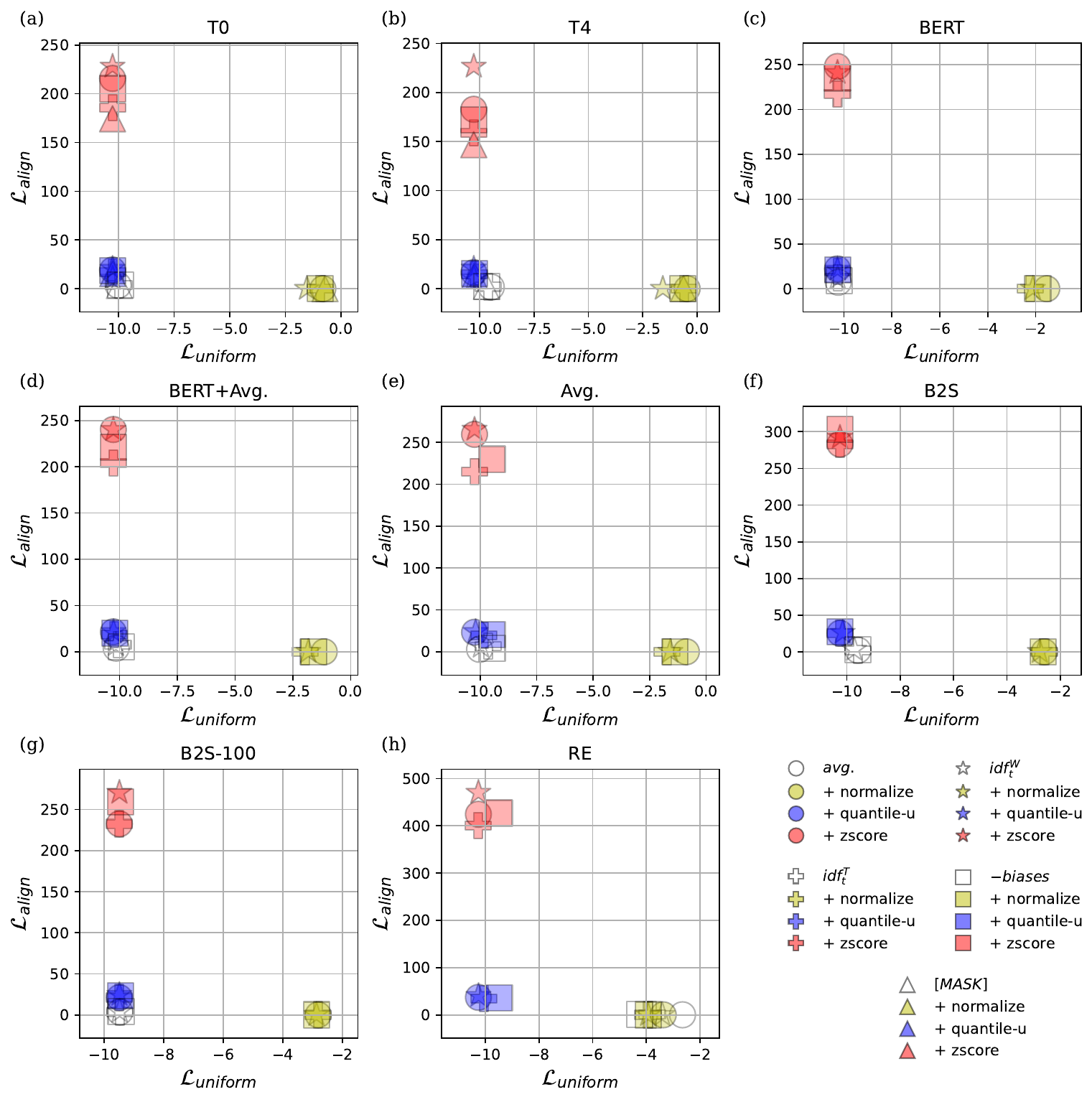}}
\caption{Alignment and uniformity of representations in relation to various token pooling and post-processing techniques. Lower values are better.}
\label{fig_iso3}
\centering
\end{figure}

We can observe in Figure~\ref{fig_iso3} that the z-score post-processing always makes alignment worse, while normalization of embeddings almost always does the same for the uniformity of the representations. Excluding the random embeddings model, the best alignment and uniformity properties are almost always with plain averaging and no post-processing applied. 

\vspace{-5pt}

Both alignment and uniformity are sensitive to the scaling of the embedding vectors. Depending on the resulting scaling from the post-processing method, either one or the other is increased. However, decreasing them both using these methods is found to be difficult. This finding strengthens the reputation of these two metrics that only training the transformer model, as shown in \cite{gao-etal-2021-simcse}, is capable of improving them.

As we mentioned, the outlier in its behavior here is the RE model, which improves its uniformity by different weighting of tokens or using quantile-uniform post-processing. Indeed, the latter post-processing is the least harmful for all models considered and disturbs alignment and uniformity properties less.

\FloatBarrier

\subsection{Using Prompts}\label{prompt_results}

The normal use of prompts, as presented in \cite{jiang-etal-2022-promptbert}, is to place the sentence \texttt{[X]} inside a template as ``\texttt{This sentence: "[X]" means [MASK]}'' and use only the vector of \texttt{[MASK]} token from the final layer as the full representation of \texttt{[X]}. Our experiments show that in general, using prompts is beneficial; however, we found some ways to improve performance even more.

First, extracting only the vector of \texttt{[MASK]} is not necessarily the best. We found that a simple average of all tokens, including the ones from the prompt template, and also the \texttt{[MASK]} token, is still a valid approach. Even more, it outperforms only the \texttt{[MASK]} token approach for clustering and classification tasks, with corresponding 55.0 and 79.9 average accuracies versus 54.2 and 76.8 for using only the \texttt{[MASK]} token (see T4 model results in Table~\ref{tokens_sts}).

\begin{figure}[h]
{\includegraphics[width=\textwidth]{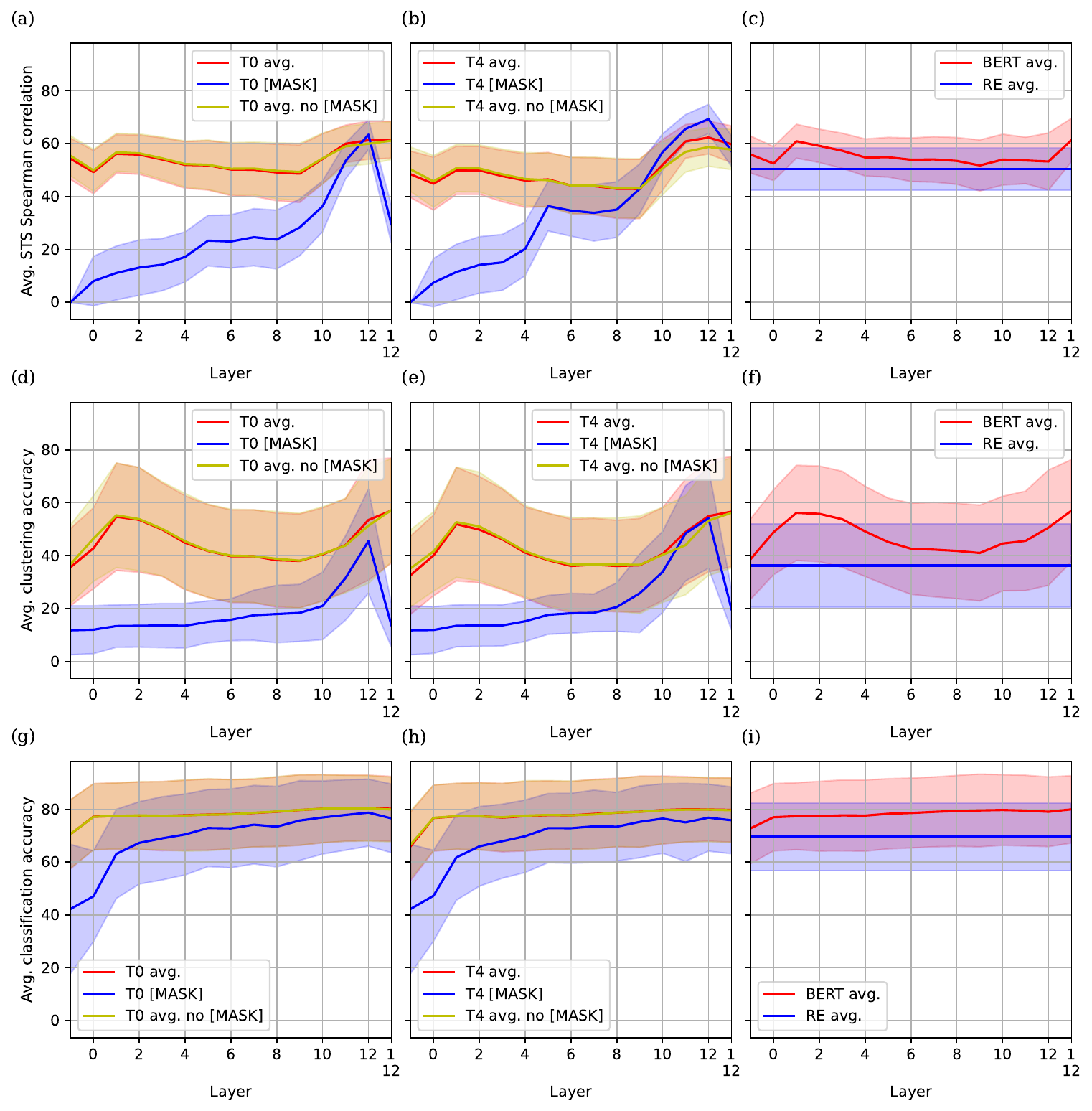}}
\caption{Layer-wise %
 performance of templated models T0 (subfigures a, d, g) and T4 (b, e, h), as well as BERT versus RE with no weighting or post-processing (c, f, i). The average performance of STS (a, b, c), clustering (d, e, f), and classification tasks (g, h, i) is shown by the lines, while shadow areas correspond to the standard deviation. We also show first + last aggregation over layers as the last tick $1 \atop 12$ on the horizontal axis.}
\label{figprompt}

\end{figure}

The use of the additional text template around the text enriches the target text representation. Let us compare the performance of 12th layer averaged token representations with and without a template (see Figure~\ref{figprompt}, T0 avg. and T4 avg. with a template versus BERT avg. without). For all 3 groups of tasks---semantic textual similarity, clustering, and classification---the T0 template achieves 61.3 average Spearman correlation, 53.5 clustering accuracy, and 80.5 classification accuracy; the T4 template reaches 62.3, 55.0, and 79.9, while a regular, non-templated text obtains just 53.2, 50.6, and 79.0, respectively. This shows that the use of the templates allows the model to enrich target text representation. However, this effect peaks at the last, 12th layer, and the achieved performance is similar to the first + last layer combination of the regular non-templated vectors.

Performance of using only the \texttt{[MASK]} token also peaks at the 12th layer, so we investigated what influence it has in the enrichment of templated target text representation. Could it be that this token is the main culprit for better performance of averaged templated text representations? To answer this question, we tried to omit the \texttt{[MASK]} token from averaging (see Figure~\ref{figprompt} T0 avg. no \texttt{[MASK]} and T4 avg. no \texttt{[MASK]}). For the T4 template and 12th layer representations, averaging all tokens yields 62.3 average Spearman correlation, 55.0 clustering accuracy, and 79.9 classification accuracy; dropping the \texttt{[MASK]} token from averaging yields 58.7, 53.3, and 79.8; non-templated performance is 53.2, 50.6, and 79.0, respectively. We see that omitting the \texttt{[MASK]} token in averaging indeed hurts the performance. On the other hand, the results show that it is responsible only for approximately half of the improvements, with the other half coming from the other tokens in the templated text.

One good reason to use averaging instead of only the \texttt{[MASK]} token is that then token weighting can be also applied. As we already showed in Table~\ref{tokens_sts}), the T4 template together with $idf_{t}^{T}$ token weighting and uniform quantile post-processing allowed us to reach the average Spearman correlation of 71.6, which was the best among the tried methods. Also, we observed that post-processing on text representations from \texttt{[MASK]} token was not as effective as from averaged ones. This also suggests that all tokens in templated texts have richer representations.

\vspace{-3pt}

\FloatBarrier

\subsection{BERT + Avg. Model}\label{bert_and_avg}

One of the ways we sought to improve BERT transformer model representations is to combine embeddings of the regular BERT with the ones averaged over multiple contexts. That is, for each token in each BERT layer, we collected many different contexts, and the averaged vector became the vector of the Avg. model. We then combined BERT and Avg. model, according to the parameter $w$, which shows the fraction of Avg. model representations in the resulting vector $v$:
\begin{equation}
v = v_{BERT}(1-w) + v_{Avg.}w.
\label{eq_bert_avg}
\end{equation} 
{We have} also varied the $w$ parameter to the negative values in~\eqref{eq_bert_avg} to see if subtracting (instead of adding) the context-average representations from the context-aware ones helps. 

\begin{figure}[h]
\hspace{-2pt}{\includegraphics[width=\textwidth]{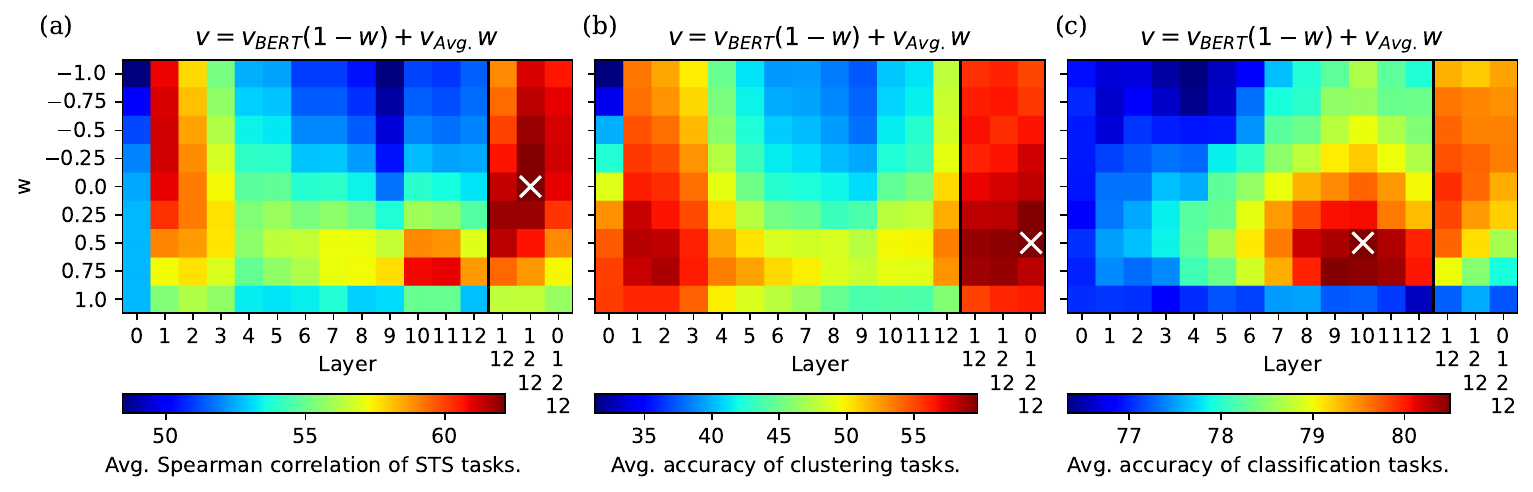}}
\caption{BERT + Avg. %
 model performance dependence on the weight $w$ of Avg. model and layer, from which (for both models) representations are used. To the right of the black line on the horizontal axis, average aggregation of multiple layers is also shown. Tokens are simply averaged and no post-processing is used. The horizontal line with $w=0.0$ corresponds to a regular Bert (B) model, $w=0.5$ is B + Avg., and $w=1.0$ is the Avg. model. The white $\times$ marks the maximum value.}
\label{fig_bert_and_avg}
\centering
\end{figure}

The impact of the $w$ parameter and the choice of the layer to source the representations (same layer for both BERT and Avg.) is presented in Figure~\ref{fig_bert_and_avg}.

One can see that for clustering and classification tasks, the combination of both models in equal portions of $w=0.5$ is better than these models alone ($w=0.0$ corresponds to BERT, $w=1.0$ is a single Avg. model). If we look at the best scores with the two techniques applied in Table~\ref{tokens_sts}, for the average clustering accuracy this translates to an improvement of 1.4\% from the 63.4\% of BERT to 64.8\% of BERT + Avg. 

Meanwhile, as we see in the same table, the average accuracy of classification tasks without additional techniques applied to the combination of the first and last layers is similar between only BERT (79.9\%) and BERT + Avg.~(79.7\%). However, as we see in Figure~\ref{fig_bert_and_avg}, differently from the first + last optimal layer setting for BERT, BERT + Avg. has a sweet spot in the 10th layer with an average classification accuracy of 80.5\%, surpassing that of BERT by 0.6\%. 

These results suggest a conclusion that for clustering and classification tasks, combining a regular BERT token with the same one but averaged over multiple contexts is beneficial.

As we already mentioned, STS tasks preferred the regular BERT ($w=0$) to the BERT + Avg. on average. However, for several individual semantic textual similarity datasets, the best weights turned out to be even negative. This is the case for STS15 ($w=-0.25$) and STS16 ($w=-0.75$). The same negative weight of $w=-0.5$ preference was also observed for the searchsnippets clustering dataset. Although the Spearman correlation for three STS tasks was only up to several percent higher, for searchsnippets, the clustering accuracy increased to a staggering 80.2\% from 72.2\% of $w=0.0$. This shows that the determined values of $w=0.0$ for STS and $w=0.5$ for clustering and classification are not universal, and for a small percentage of tasks, they can differ. For more BERT + Avg. task-wise details, see Appendix~\ref{appendix_bert_avg} figures for STS (Figure~\ref{weights_times_layers_sts}), clustering (Figure~\ref{weights_times_layers_clustering}), and classification (Figure~\ref{weights_times_layers_classification}) performance.

\subsubsection*{{Layers} } 
Figure~\ref{fig_bert_and_avg} also depicts the task performance versus BERT layers. Our first observation is that for STS and clustering tasks, there is a strong preference for the first layers. As a result, the best combination of layers also involves the first ones; for semantic textual similarity tasks, it is the average of representations from one, two, and twelve layers, and for clustering, zero, one, two, and twelve layers. Even individual SICK-E, SICK-R, and STS-B classification tasks, which originate from semantic textual similarity ones, have strong first layers (see Figure~\ref{weights_times_layers_classification} in Appendix~\ref{appendix_bert_avg}), although on average, the best layer for classification tasks is the tenth. This shows that a lot of performance for tasks that work on similarities between texts (STS and clustering) depends on the first layers, which, according to \cite{voita-etal-2019-bottom}, have more generalized token representations. However, the last layer, which receives recreated token identities \cite{voita-etal-2019-bottom} is also important, as the best combinations (such as 1 + 2 + 12) also involve it. This first + last cooperation of layers can be distinctly observed as a U shape for clustering performance in layers (see Figure~\ref{fig_bert_and_avg} and also Figure~\ref{figprompt}).

For most classification tasks (excluding those originating from STS), the U shape is inversed. This is very clearly seen for binary classification tasks (see Figure~\ref{weights_times_layers_classification} in Appendix~\ref{appendix_bert_avg}), where the highest classification accuracy is concentrated between the ninth and the eleventh layers. This aligns with the explanations of the previous work \cite{chung2021rethinking}, which argues that the last layer is over-specialized for the training objective.

\FloatBarrier
\section{Conclusions}\label{conclusion}

We empirically evaluated the effects of various aggregation and post-processing techniques of token representations in a trained transformer and other models to form good text-, paragraph-, or sentence-level embeddings. We carried out the empirical evaluation of the embeddings on three classes of downstream text-level tasks: Semantic Textual Similarity (STS), clustering, and classification. 

We found the techniques to benefit all models studied for the unsupervised STS (the best model average Spearman correlation increased from 62.3\% to 71.6\%) and clustering (the best model average clustering accuracy increased from 59.2\% to 64.8\%), while it had no positive effect on the supervised classification tasks (see Table~\ref{tokens_sts}).

We present a strong and very simple baseline model of Random Embeddings (RE), where every token is assigned a random vector as its embedding. Combined with token aggregation and post-processing techniques, it also almost matches the average STS performance of the BERT model with the techniques applied, with 66.4\% versus 69.8\% average Spearman correlation. It also shows very high performance for some tasks, like stackoverflow classification, where BERT token contextualization may not work well on code samples in the texts (see Section~\ref{re_results} for more details). We encourage future work to use RE as a baseline, due to its mid-level performance, simple implementation, and ability to separate the contribution to the performance of the learned contexts from the aggregation and post-processing techniques. 

We found that the aggregation and post-processing techniques tried typically increase the isotropy of the representations, and the isotropy for most models is positively correlated (up to 69\% Pearson correlation) with the Spearman correlation of STS tasks (Section~\ref{isotropy_results}). The highest isotropy improvement is observed for our Random Embeddings model, since its token representations have the maximal isotropy to start with. We did not find the token aggregation and post-processing techniques to improve the alignment and uniformity properties of representations.

We question the use of prompts (adding a sentence into a certain text template) for retrieving the representation of the sentence from only the \texttt{[MASK]} token. Our experiments show that averaging all the templated text tokens, with idf weighting and post-processing for the unsupervised tasks, is better. Meanwhile, the average increase in performance due to the added template is only obvious for the STS tasks, gives no improvement in clustering, and is very slight in classification (see Table~\ref{tokens_sts}, Section~\ref{prompt_results}).

We presented a static vector model Avg., which simply contains BERT tokens averaged over multiple different contexts. Our experiments show that it outperforms a more complex BERT2Static \cite{gupta-jaggi-2021-obtaining}, also a static word-level model, yet specially trained on BERT contexts. With the best post-processing and token aggregation techniques, the advantage for unsupervised STS tasks is 69.5 versus 66.4, and for clustering, it is 62.2 versus 57.2., with a negligible difference for classification tasks. Moreover, we show that combining Avg. with the parent BERT model can bring even further improvements. In particular, BERT + Avg. reached the highest average clustering accuracy of 64.8 out of all our considered models, as well as the classification accuracy of 80.5 (Section~\ref{bert_and_avg}). We encourage future work to also use Avg. as a baseline, both due to its upper-level performance and simple implementation.

In our work, we also analyzed prompt and BERT + Avg. models layer-wise. We found that for the STS tasks, taking the representations from the first layers performs better, and for the clustering task, the performance profile forms a ``U'' shape with tops at the first and last layers. Therefore, for these two task groups, we mostly use the average of first + last layers, harnessing both of the tops for the best performance. On the other hand, classification tasks have an inverted ``U'' shape with the top in the 10th layer (Figure~\ref{fig_bert_and_avg}). We did not find token aggregation and post-processing techniques to change such profile curvature.

In this research, we used a pre-trained BERT as a manageable representative of transformer models, but the findings should be transferable to other types of transformers, including large language models. We also specifically did not perform task-specific fine-tuning of the model to keep it universal. This enables the same model to be used in multiple ways. For example, given a prompt, its text-level embedding can be extracted from the model using the techniques investigated here, and this embedding can be used to search an external database for related information to add the prompt to the same model, i.e., we can use the same model for both query encoding and generation in retrieval-augmented generation \cite{RAG}. Alternatively, one of the simpler baseline models proposed here could be used as the query encoder.

\bibliographystyle{IEEEtran}
{\footnotesize
\bibliography{references}}

\newpage
\appendix{}

\section{Detailed Performance on Individual Tasks}\label{appendix1}

\subsection{Token Aggregation and Post-Processing}\label{appendix_agg_post}

\vspace{-6pt}
\begin{table}[h]\small
\caption{Spearman %
 correlation dependence on token aggregation and post-processing techniques for semantic textual similarity STS12 task. The best result for each model is bolded, while underlined result is the best across all the models.}
\label{STS12}
\setlength\tabcolsep{5.3pt}
\begin{tabular}{lrrrrrrrr}
\toprule
\textbf{Model} & \textbf{T0} & \textbf{T4} & \textbf{BERT} & \textbf{BERT + Avg.} & \textbf{Avg.} & \textbf{B2S} & \textbf{B2S-100} & \textbf{RE} \\
\cmidrule(r){2-3} \cmidrule(lr){4-6} \cmidrule(l){7-9}
\textbf{Layer} & \multicolumn{2}{c}{\textbf{Last}} & \multicolumn{3}{c}{\textbf{First + Last}} & \multicolumn{3}{c}{\textbf{--}} \\
\midrule
avg. & 48.7 & 53.9 & 45.1 & 48.7 & 48.5 & 47.2 & 54.2 & 34.7 \\
\quad+ normalize %
 & 48.7 & 53.9 & 45.1 & 48.7 & 48.5 & 47.2 & 54.2 & 34.7 \\
\quad+ quantile-u & 52.1 & 56.2 & 48.7 & 51.9 & 51.5 & 45.8 & 52.4 & 40.1 \\
\quad+ quantile-u$^W$ & 50.2 & 55.1 & 46.6 & 50.7 & 50.3 & 44.6 & 52.5 & 36.9 \\
\quad+ whiten & 41.9 & 42.9 & 43.2 & 44.5 & 46.0 & 45.0 & 44.6 & 44.1 \\
\quad+ whiten$^W$ & 46.2 & 50.4 & 47.3 & 52.8 & 57.7 & 54.7 & 55.0 & 54.2 \\
\quad+ zscore & 51.5 & 54.1 & 50.2 & 53.4 & 53.1 & 49.1 & 54.7 & 43.6\vspace{5pt}\\

$\mbox{idf}_t^W$ & 56.2 & 56.8 & 57.5 & 58.7 & 58.3 & 57.6 & \textbf{56.9} & 55.4 \\
\quad+ normalize & 56.2 & 56.8 & 57.5 & 58.7 & 58.3 & \textbf{57.7} & 56.9 & 55.4 \\
\quad+ quantile-u & 56.2 & 56.5 & 56.4 & 57.9 & 57.5 & 54.5 & 53.6 & 52.0 \\
\quad+ quantile-u$^W$ & 57.0 & 57.5 & 57.1 & 58.7 & 58.4 & 55.4 & 54.4 & 52.6 \\
\quad+ whiten & 43.7 & 43.8 & 46.1 & 46.7 & 47.2 & 45.3 & 44.7 & 44.6 \\
\quad+ whiten$^W$ & 52.6 & 53.3 & 57.5 & 58.8 & 58.7 & 57.5 & 56.8 & 55.5 \\
\quad+ zscore & 56.1 & 55.9 & 57.4 & 58.7 & 58.2 & 56.9 & 56.4 & 55.6\vspace{5pt}\\

$\mbox{idf}_t^T$ & \textbf{59.2} & 64.4 & 57.3 & 59.0 & 59.0 & 47.2 & 54.2 & 55.1 \\
\quad+ normalize & 59.2 & \underline{\textbf{64.4}} & 57.3 & 59.0 & 59.0 & 47.2 & 54.2 & 55.1 \\
\quad+ quantile-u & 58.5 & 62.6 & 56.5 & 58.2 & 58.1 & 45.8 & 52.4 & 52.3 \\
\quad+ quantile-u$^W$ & 59.1 & 63.9 & 57.0 & 58.8 & 58.9 & 44.6 & 52.5 & 52.9 \\
\quad+ whiten & 43.8 & 45.3 & 46.0 & 46.7 & 47.4 & 45.0 & 44.6 & 44.6 \\
\quad+ whiten$^W$ & 52.6 & 56.3 & 56.9 & 58.2 & 58.4 & 54.7 & 55.0 & 55.1 \\
\quad+ zscore & 56.9 & 59.7 & 57.7 & 59.3 & 59.3 & 49.1 & 54.7 & 55.6\vspace{5pt}\\

-biases & 55.1 & 58.2 & 58.6 & \textbf{62.2} & \textbf{62.9} & 45.1 & 51.9 & 57.5 \\
\quad+ normalize & 55.1 & 58.2 & \textbf{58.6} & 62.2 & 62.8 & 45.1 & 51.9 & 57.5 \\
\quad+ quantile-u & 56.0 & 59.6 & 57.9 & 60.9 & 61.2 & 44.0 & 50.3 & 54.6 \\
\quad+ quantile-u$^W$ & 54.7 & 59.1 & 58.2 & 61.5 & 61.6 & 42.7 & 50.3 & 55.1 \\
\quad+ whiten$^W$ & 51.8 & 54.0 & 57.2 & 60.2 & 61.6 & 53.2 & 53.3 & 58.1 \\
\quad+ whiten & 42.5 & 43.2 & 45.1 & 46.2 & 47.6 & 44.2 & 44.2 & 46.2 \\
\quad+ zscore & 55.1 & 57.5 & 57.8 & 60.4 & 60.6 & 47.4 & 52.7 & \textbf{58.2}\vspace{5pt}\\

\verb|[MASK]| & 55.2 & 60.6 &  &  &  &  &  &  \\
\quad+ normalize & 55.2 & 60.6 &  &  &  &  &  &  \\
\quad+ quantile-u & 56.0 & 60.8 &  &  &  &  &  &  \\
\quad+ quantile-u$^W$ & 56.6 & 61.7 &  &  &  &  &  &  \\
\quad+ whiten & 47.7 & 48.0 &  &  &  &  &  &  \\
\quad+ whiten$^W$ & 52.5 & 57.7 &  &  &  &  &  &  \\
\quad+ zscore & 55.4 & 59.8 &  &  &  &  &  &  \\
\bottomrule
\end{tabular}
\end{table}

\begin{table}[h]\small
\caption{Spearman correlation dependence on token aggregation and post-processing techniques for semantic textual similarity STS13 task. The best result for each model is bolded, while underlined result is the best across all the models.}
\label{STS13}
\setlength\tabcolsep{5.3pt}
\begin{tabular}{lrrrrrrrr}
\toprule
\textbf{Model} & \textbf{T0} & \textbf{T4} & \textbf{BERT} & \textbf{BERT + Avg.} & \textbf{Avg.} & \textbf{B2S} & \textbf{B2S-100} & \textbf{RE} \\
\cmidrule(r){2-3} \cmidrule(lr){4-6} \cmidrule(l){7-9}
\textbf{Layer} & \multicolumn{2}{c}{\textbf{Last}} & \multicolumn{3}{c}{\textbf{First + Last}} & \multicolumn{3}{c}{\textbf{--}} \\
\midrule
avg. & 62.6 & 64.7 & 64.3 & 63.4 & 56.6 & 61.4 & 66.2 & 48.8 \\
\quad+ normalize %
 & 62.6 & 64.7 & 64.3 & 63.4 & 56.6 & 61.4 & 66.2 & 48.8 \\
\quad+ quantile-u & 68.4 & 70.9 & 68.3 & 69.0 & 65.5 & 60.4 & 68.9 & 54.8 \\
\quad+ quantile-u$^W$ & 64.0 & 66.5 & 65.7 & 66.8 & 64.0 & 59.2 & 67.2 & 53.3 \\
\quad+ whiten & 76.0 & 76.6 & 77.6 & 78.3 & 78.0 & 76.4 & 76.1 & \textbf{75.1} \\
\quad+ whiten$^W$ & 62.2 & 66.5 & 64.8 & 65.8 & 66.2 & 60.6 & 60.8 & 63.7 \\
\quad+ zscore & 70.7 & 72.5 & 70.4 & 70.7 & 67.1 & 62.8 & 68.9 & 55.9\vspace{5pt}\\

$\mbox{idf}_t^W$ & 75.4 & 74.7 & 77.2 & 79.1 & 77.8 & \textbf{77.5} & 77.3 & 72.5 \\
\quad+ normalize & 75.4 & 74.7 & 77.2 & 79.1 & 77.8 & 77.5 & 77.3 & 72.5 \\
\quad+ quantile-u & 76.7 & 76.0 & 77.7 & 79.3 & 78.7 & 77.1 & 76.3 & 73.4 \\
\quad+ quantile-u$^W$ & 75.6 & 74.7 & 77.1 & 79.3 & 79.1 & 77.2 & 76.7 & 74.2 \\
\quad+ whiten & 76.8 & 76.0 & \textbf{78.3} & 78.6 & 77.6 & 77.1 & 76.7 & 74.6 \\
\quad+ whiten$^W$ & 68.1 & 68.1 & 74.1 & 75.2 & 75.4 & 73.1 & 73.9 & 72.7 \\
\quad+ zscore & 76.7 & 76.4 & 78.2 & \underline{\textbf{79.7}} & \textbf{79.3} & 77.5 & \textbf{78.0} & 73.0\vspace{5pt}\\

$\mbox{idf}_t^T$ & 70.8 & 73.7 & 74.8 & 76.5 & 76.0 & 61.4 & 66.2 & 68.3 \\
\quad+ normalize & 70.8 & 73.7 & 74.8 & 76.5 & 76.0 & 61.4 & 66.2 & 68.3 \\
\quad+ quantile-u & 74.7 & 77.0 & 76.0 & 77.2 & 76.7 & 60.4 & 68.9 & 71.9 \\
\quad+ quantile-u$^W$ & 72.0 & 74.6 & 74.8 & 76.4 & 76.0 & 59.2 & 67.2 & 71.5 \\
\quad+ whiten & \textbf{77.1} & \textbf{78.0} & 78.1 & 78.3 & 77.3 & 76.4 & 76.1 & 74.0 \\
\quad+ whiten$^W$ & 67.2 & 71.0 & 70.3 & 70.4 & 70.1 & 60.6 & 60.8 & 68.1 \\
\quad+ zscore & 75.9 & 77.7 & 76.2 & 76.9 & 75.8 & 62.8 & 68.9 & 69.8\vspace{5pt}\\

-biases & 66.5 & 68.7 & 68.4 & 69.9 & 68.1 & 56.3 & 60.5 & 61.3 \\
\quad+ normalize & 66.5 & 68.7 & 68.4 & 69.9 & 68.1 & 56.3 & 60.5 & 61.3 \\
\quad+ quantile-u & 70.9 & 73.5 & 70.8 & 71.9 & 70.3 & 55.0 & 63.2 & 66.2 \\
\quad+ quantile-u$^W$ & 67.4 & 70.3 & 68.6 & 69.9 & 68.2 & 53.9 & 61.3 & 64.4 \\
\quad+ whiten$^W$ & 63.1 & 67.1 & 65.2 & 65.7 & 65.9 & 56.6 & 56.7 & 63.0 \\
\quad+ whiten & 76.2 & 76.8 & 77.9 & 78.5 & 78.1 & 73.3 & 73.1 & 74.7 \\
\quad+ zscore & 73.0 & 74.9 & 72.5 & 72.9 & 70.6 & 58.5 & 64.0 & 64.6\vspace{5pt}\\

\verb|[MASK]| & 63.4 & 76.2 &  &  &  &  &  &  \\
\quad+ normalize & 63.4 & 76.2 &  &  &  &  &  &  \\
\quad+ quantile-u & 68.1 & 77.3 &  &  &  &  &  &  \\
\quad+ quantile-u$^W$ & 65.2 & 76.0 &  &  &  &  &  &  \\
\quad+ whiten & 73.1 & 77.6 &  &  &  &  &  &  \\
\quad+ whiten$^W$ & 61.7 & 70.9 &  &  &  &  &  &  \\
\quad+ zscore & 69.4 & 76.9 &  &  &  &  &  &  \\
\bottomrule
\end{tabular}
\end{table}

\vspace{-9pt}
\begin{table}[h]\small
\caption{Spearman %
 correlation dependence on token aggregation and post-processing techniques for semantic textual similarity STS14 task. The best result for each model is bolded, while underlined result is the best across all the models.}
\label{STS14}
\setlength\tabcolsep{5.3pt}
\begin{tabular}{lrrrrrrrr}
\toprule
\textbf{Model} & \textbf{T0} & \textbf{T4} & \textbf{BERT} & \textbf{BERT + Avg.} & \textbf{Avg.} & \textbf{B2S} & \textbf{B2S-100} & \textbf{RE} \\
\cmidrule(r){2-3} \cmidrule(lr){4-6} \cmidrule(l){7-9}
\textbf{Layer} & \multicolumn{2}{c}{\textbf{Last}} & \multicolumn{3}{c}{\textbf{First + Last}} & \multicolumn{3}{c}{\textbf{--}} \\
\midrule
avg. & 53.9 & 54.5 & 54.6 & 55.4 & 51.7 & 55.9 & 61.2 & 48.2 \\
\quad+ normalize %
 & 53.9 & 54.5 & 54.6 & 55.4 & 51.7 & 55.9 & 61.2 & 48.2 \\
\quad+ quantile-u & 57.8 & 60.1 & 58.1 & 60.0 & 58.2 & 53.5 & 62.9 & 52.3 \\
\quad+ quantile-u$^W$ & 55.6 & 57.1 & 57.0 & 59.4 & 58.1 & 53.5 & 62.2 & 52.1 \\
\quad+ whiten & 64.0 & 65.0 & 66.2 & 68.1 & 69.4 & 67.7 & 68.5 & 68.3 \\
\quad+ whiten$^W$ & 53.5 & 57.0 & 57.6 & 61.1 & 64.5 & 60.7 & 61.3 & 64.6 \\
\quad+ zscore & 58.6 & 61.3 & 58.7 & 61.2 & 59.6 & 55.7 & 61.3 & 53.5\vspace{5pt}\\

$\mbox{idf}_t^W$ & 65.5 & 65.9 & 65.2 & 67.7 & 67.8 & 67.2 & 67.8 & 67.6 \\
\quad+ normalize & 65.5 & 65.9 & 65.2 & 67.7 & 67.8 & 67.2 & 67.8 & 67.6 \\
\quad+ quantile-u & \textbf{66.7} & 66.7 & 66.5 & 69.0 & 69.3 & 67.2 & 68.1 & 66.4 \\
\quad+ quantile-u$^W$ & 66.5 & 66.6 & 66.2 & 68.9 & 69.3 & 67.1 & 68.0 & 67.0 \\
\quad+ whiten & 65.4 & 65.4 & 67.6 & 68.5 & 68.8 & \textbf{68.3} & \textbf{68.7} & 67.9 \\
\quad+ whiten$^W$ & 59.2 & 59.4 & 64.7 & 67.0 & 68.2 & 66.6 & 67.4 & 68.0 \\
\quad+ zscore & 65.7 & 65.3 & 66.2 & 68.8 & 69.2 & 66.8 & 68.0 & 67.8\vspace{5pt}\\

$\mbox{idf}_t^T$ & 61.6 & 63.9 & 64.3 & 66.7 & 67.1 & 55.9 & 61.2 & 65.5 \\
\quad+ normalize & 61.6 & 63.9 & 64.3 & 66.7 & 67.1 & 55.9 & 61.2 & 65.5 \\
\quad+ quantile-u & 64.2 & 66.5 & 65.8 & 67.8 & 68.0 & 53.5 & 62.9 & 65.3 \\
\quad+ quantile-u$^W$ & 62.9 & 65.4 & 65.3 & 67.5 & 67.8 & 53.5 & 62.2 & 65.7 \\
\quad+ whiten & 65.2 & \textbf{66.8} & 67.5 & 68.3 & 68.5 & 67.7 & 68.5 & 67.4 \\
\quad+ whiten$^W$ & 58.0 & 61.6 & 63.0 & 64.6 & 65.3 & 60.7 & 61.3 & 65.3 \\
\quad+ zscore & 64.5 & 66.8 & 65.1 & 67.1 & 67.2 & 55.7 & 61.3 & 65.7\vspace{5pt}\\

-biases & 60.7 & 62.0 & 63.8 & 66.9 & 66.9 & 53.1 & 58.1 & 64.7 \\
\quad+ normalize & 60.7 & 62.0 & 63.8 & 66.9 & 66.9 & 53.1 & 58.1 & 64.7 \\
\quad+ quantile-u & 63.4 & 65.9 & 65.2 & 67.9 & 67.8 & 50.8 & 59.8 & 65.2 \\
\quad+ quantile-u$^W$ & 61.9 & 64.3 & 64.8 & 67.7 & 67.6 & 50.9 & 59.2 & 65.3 \\
\quad+ whiten$^W$ & 57.3 & 60.4 & 62.8 & 65.5 & 67.3 & 58.5 & 58.9 & 66.2 \\
\quad+ whiten & 64.9 & 66.2 & \textbf{68.3} & \textbf{69.6} & \underline{\textbf{70.4}} & 65.8 & 66.4 & \textbf{68.8} \\
\quad+ zscore & 64.1 & 66.3 & 64.8 & 67.4 & 67.2 & 53.3 & 58.6 & 65.3\vspace{5pt}\\

\verb|[MASK]| & 53.9 & 63.9 &  &  &  &  &  &  \\
\quad+ normalize & 53.9 & 63.9 &  &  &  &  &  &  \\
\quad+ quantile-u & 56.1 & 65.1 &  &  &  &  &  &  \\
\quad+ quantile-u$^W$ & 54.8 & 64.7 &  &  &  &  &  &  \\
\quad+ whiten & 60.8 & 66.3 &  &  &  &  &  &  \\
\quad+ whiten$^W$ & 51.5 & 61.3 &  &  &  &  &  &  \\
\quad+ zscore & 56.7 & 65.1 &  &  &  &  &  &  \\
\bottomrule
\end{tabular}
\end{table}

\vspace{-10pt}
\begin{table}[h]\small
\caption{Spearman %
 correlation dependence on token aggregation and post-processing techniques for semantic textual similarity STS15 task. The best result for each model is bolded, while underlined result is the best across all the models.}
\label{STS15}
\setlength\tabcolsep{5.3pt}
\begin{tabular}{lrrrrrrrr}
\toprule
\textbf{Model} & \textbf{T0} & \textbf{T4} & \textbf{BERT} & \textbf{BERT + Avg.} & \textbf{Avg.} & \textbf{B2S} & \textbf{B2S-100} & \textbf{RE} \\
\cmidrule(r){2-3} \cmidrule(lr){4-6} \cmidrule(l){7-9}
\textbf{Layer} & \multicolumn{2}{c}{\textbf{Last}} & \multicolumn{3}{c}{\textbf{First + Last}} & \multicolumn{3}{c}{\textbf{--}} \\
\midrule
avg. & 70.2 & 70.6 & 70.5 & 70.0 & 64.2 & 69.8 & 74.2 & 62.1 \\
\quad+ normalize %
 & 70.2 & 70.6 & 70.5 & 70.0 & 64.2 & 69.8 & 74.2 & 62.1 \\
\quad+ quantile-u & 71.8 & 73.7 & 72.7 & 74.3 & 72.4 & 68.9 & 73.6 & 61.4 \\
\quad+ quantile-u$^W$ & 71.5 & 72.5 & 72.3 & 74.1 & 72.6 & 69.4 & 74.4 & 62.7 \\
\quad+ whiten & 62.6 & 62.9 & 64.0 & 65.9 & 69.3 & 69.3 & 68.8 & 67.9 \\
\quad+ whiten$^W$ & 70.2 & 71.8 & 71.9 & 74.9 & 77.0 & 74.7 & 75.2 & 74.7 \\
\quad+ zscore & 71.9 & 73.0 & 72.6 & 74.3 & 72.4 & 69.3 & 73.9 & 64.3\vspace{5pt}\\

$\mbox{idf}_t^W$ & 75.4 & 74.4 & 75.8 & 75.4 & 71.5 & 73.7 & 74.4 & 74.4 \\
\quad+ normalize & 75.4 & 74.4 & 75.8 & 75.4 & 71.5 & 73.7 & 74.4 & 74.4 \\
\quad+ quantile-u & 75.1 & 74.2 & 75.3 & 75.7 & 73.7 & 73.4 & 71.9 & 68.5 \\
\quad+ quantile-u$^W$ & \textbf{76.5} & 75.6 & 76.3 & 76.9 & 74.8 & 74.6 & 73.7 & 71.1 \\
\quad+ whiten & 62.7 & 61.4 & 64.3 & 65.2 & 66.3 & 66.8 & 66.7 & 65.0 \\
\quad+ whiten$^W$ & 72.2 & 71.6 & 75.0 & 76.5 & 76.6 & \textbf{75.4} & \textbf{75.5} & 74.4 \\
\quad+ zscore & 74.2 & 73.1 & 75.2 & 76.2 & 75.0 & 73.8 & 74.4 & 73.2\vspace{5pt}\\

$\mbox{idf}_t^T$ & 74.1 & 75.2 & 76.2 & 76.8 & 75.1 & 69.8 & 74.2 & 73.8 \\
\quad+ normalize & 74.1 & 75.2 & 76.2 & 76.8 & 75.1 & 69.8 & 74.2 & 73.8 \\
\quad+ quantile-u & 75.2 & \textbf{77.1} & 76.0 & 76.8 & 75.7 & 68.9 & 73.6 & 69.5 \\
\quad+ quantile-u$^W$ & 75.4 & 77.0 & \textbf{76.9} & 77.8 & 76.7 & 69.4 & 74.4 & 71.6 \\
\quad+ whiten & 63.1 & 64.2 & 64.5 & 66.0 & 68.5 & 69.3 & 68.8 & 67.2 \\
\quad+ whiten$^W$ & 72.9 & 74.6 & 75.4 & 76.2 & 75.8 & 74.7 & 75.2 & 73.1 \\
\quad+ zscore & 75.0 & 76.0 & 75.9 & 76.8 & 75.8 & 69.3 & 73.9 & 72.7\vspace{5pt}\\

-biases & 72.9 & 73.3 & 76.2 & 77.3 & 75.2 & 67.3 & 70.8 & 74.3 \\
\quad+ normalize & 72.9 & 73.3 & 76.2 & 77.3 & 75.2 & 67.3 & 70.8 & 74.3 \\
\quad+ quantile-u & 74.1 & 76.1 & 76.3 & 77.5 & 76.2 & 66.3 & 70.1 & 69.6 \\
\quad+ quantile-u$^W$ & 73.7 & 75.3 & 76.8 & \underline{\textbf{78.4}} & \textbf{77.3} & 66.9 & 70.9 & 71.7 \\
\quad+ whiten$^W$ & 72.6 & 73.7 & 75.4 & 77.0 & 77.3 & 72.6 & 72.6 & \textbf{74.8} \\
\quad+ whiten & 64.2 & 64.8 & 65.2 & 66.5 & 69.1 & 67.4 & 66.4 & 66.8 \\
\quad+ zscore & 74.4 & 75.1 & 75.7 & 76.9 & 75.7 & 67.0 & 70.7 & 73.6\vspace{5pt}\\

\verb|[MASK]| & 67.0 & 74.1 &  &  &  &  &  &  \\
\quad+ normalize & 67.0 & 74.1 &  &  &  &  &  &  \\
\quad+ quantile-u & 68.0 & 74.7 &  &  &  &  &  &  \\
\quad+ quantile-u$^W$ & 68.3 & 75.7 &  &  &  &  &  &  \\
\quad+ whiten & 61.0 & 65.1 &  &  &  &  &  &  \\
\quad+ whiten$^W$ & 67.7 & 74.5 &  &  &  &  &  &  \\
\quad+ zscore & 68.0 & 72.9 &  &  &  &  &  &  \\
\bottomrule
\end{tabular}
\end{table}

\vspace{-10pt}
\begin{table}[h]\small
\caption{Spearman %
 correlation dependence on token aggregation and post-processing techniques for semantic textual similarity STS16 task. The best result for each model is bolded, while underlined result is the best across all the models.}
\label{STS16}
\setlength\tabcolsep{5.3pt}
\begin{tabular}{lrrrrrrrr}
\toprule
\textbf{Model} & \textbf{T0} & \textbf{T4} & \textbf{BERT} & \textbf{BERT + Avg.} & \textbf{Avg.} & \textbf{B2S} & \textbf{B2S-100} & \textbf{RE} \\
\cmidrule(r){2-3} \cmidrule(lr){4-6} \cmidrule(l){7-9}
\textbf{Layer} & \multicolumn{2}{c}{\textbf{Last}} & \multicolumn{3}{c}{\textbf{First + Last}} & \multicolumn{3}{c}{\textbf{--}} \\
\midrule
avg. & 69.1 & 69.7 & 67.9 & 65.2 & 56.4 & 55.2 & 58.5 & 55.5 \\
\quad+ normalize %
 & 69.1 & 69.7 & 67.9 & 65.2 & 56.4 & 55.1 & 58.5 & 55.5 \\
\quad+ quantile-u & 70.1 & 72.4 & 70.3 & 71.0 & 67.0 & 55.9 & 59.5 & 54.8 \\
\quad+ quantile-u$^W$ & 69.7 & 71.5 & 69.8 & 70.0 & 65.3 & 54.6 & 57.9 & 55.1 \\
\quad+ whiten & 65.4 & 65.9 & 67.1 & 68.9 & 69.9 & 65.0 & 63.0 & 67.1 \\
\quad+ whiten$^W$ & 64.7 & 67.2 & 69.9 & 71.5 & 71.1 & 63.1 & 62.8 & 68.3 \\
\quad+ zscore & 69.3 & 71.0 & 72.0 & 73.2 & 69.8 & 60.2 & 62.3 & 60.4\vspace{5pt}\\

$\mbox{idf}_t^W$ & 70.1 & 69.1 & 73.9 & 73.0 & 69.5 & 68.8 & 68.0 & 71.9 \\
\quad+ normalize & 70.1 & 69.1 & 73.9 & 73.0 & 69.5 & 68.8 & 68.0 & 71.9 \\
\quad+ quantile-u & 70.5 & 69.9 & 73.7 & 74.2 & 72.9 & 68.1 & 65.8 & 68.9 \\
\quad+ quantile-u$^W$ & 70.5 & 69.8 & 74.0 & 74.4 & 72.7 & 67.9 & 66.2 & 69.6 \\
\quad+ whiten & 64.6 & 64.0 & 68.0 & 68.9 & 69.1 & 65.6 & 63.4 & 67.3 \\
\quad+ whiten$^W$ & 70.0 & 69.1 & 75.6 & 76.0 & \textbf{74.8} & 69.6 & 68.7 & \textbf{72.4} \\
\quad+ zscore & 72.2 & 71.7 & \textbf{75.6} & \underline{\textbf{76.4}} & 74.8 & \textbf{70.3} & \textbf{69.5} & 72.2\vspace{5pt}\\

$\mbox{idf}_t^T$ & 72.9 & 73.1 & 72.8 & 72.0 & 69.7 & 55.2 & 58.5 & 69.1 \\
\quad+ normalize & 72.9 & 73.1 & 72.8 & 72.0 & 69.7 & 55.2 & 58.5 & 69.1 \\
\quad+ quantile-u & \textbf{73.9} & \textbf{75.0} & 72.8 & 73.0 & 71.8 & 55.9 & 59.5 & 67.3 \\
\quad+ quantile-u$^W$ & 73.7 & 74.6 & 72.7 & 72.7 & 71.2 & 54.6 & 57.9 & 67.5 \\
\quad+ whiten & 67.0 & 68.6 & 68.0 & 68.7 & 69.5 & 65.0 & 63.0 & 67.8 \\
\quad+ whiten$^W$ & 69.0 & 71.0 & 73.2 & 72.8 & 71.3 & 63.1 & 62.8 & 69.3 \\
\quad+ zscore & 73.1 & 73.1 & 74.4 & 74.4 & 73.1 & 60.2 & 62.3 & 70.1\vspace{5pt}\\  

-biases & 70.1 & 71.9 & 72.2 & 72.1 & 68.4 & 52.8 & 56.3 & 65.4 \\
\quad+ normalize & 70.1 & 71.9 & 72.2 & 72.1 & 68.4 & 52.8 & 56.3 & 65.4 \\
\quad+ quantile-u & 71.1 & 73.7 & 72.3 & 73.3 & 71.6 & 53.5 & 57.5 & 64.7 \\
\quad+ quantile-u$^W$ & 70.6 & 73.2 & 72.1 & 72.7 & 70.4 & 52.3 & 55.8 & 64.4 \\
\quad+ whiten$^W$ & 65.2 & 68.0 & 71.0 & 72.1 & 71.5 & 61.1 & 60.9 & 68.5 \\
\quad+ whiten & 64.9 & 66.2 & 67.3 & 68.5 & 69.3 & 63.4 & 61.4 & 67.4 \\
\quad+ zscore & 71.1 & 72.5 & 73.3 & 74.1 & 72.0 & 58.2 & 60.4 & 66.7\vspace{5pt}\\

\verb|[MASK]| & 67.2 & 71.0 &  &  &  &  &  &  \\
\quad+ normalize & 67.2 & 71.0 &  &  &  &  &  &  \\
\quad+ quantile-u & 68.2 & 71.8 &  &  &  &  &  &  \\
\quad+ quantile-u$^W$ & 67.7 & 71.7 &  &  &  &  &  &  \\
\quad+ whiten & 65.5 & 70.0 &  &  &  &  &  &  \\
\quad+ whiten$^W$ & 64.1 & 69.4 &  &  &  &  &  &  \\
\quad+ zscore & 66.6 & 68.9 &  &  &  &  &  &  \\
\bottomrule
\end{tabular}
\end{table}

\vspace{-10pt}
\begin{table}[h]\small
\caption{Spearman %
 correlation dependence on token aggregation and post-processing techniques for semantic textual similarity STS-B task. The best result for each model is bolded, while underlined result is the best across all the models.}
\label{STS-B}
\setlength\tabcolsep{5.3pt}
\begin{tabular}{lrrrrrrrr}
\toprule
\textbf{Model} & \textbf{T0} & \textbf{T4} & \textbf{BERT} & \textbf{BERT + Avg.} & \textbf{Avg.} & \textbf{B2S} & \textbf{B2S-100} & \textbf{RE} \\
\cmidrule(r){2-3} \cmidrule(lr){4-6} \cmidrule(l){7-9}
\textbf{Layer} & \multicolumn{2}{c}{\textbf{Last}} & \multicolumn{3}{c}{\textbf{First + Last}} & \multicolumn{3}{c}{\textbf{--}} \\
\midrule
avg. & 60.4 & 61.0 & 59.0 & 59.9 & 54.2 & 56.1 & 60.7 & 46.5 \\
\quad+ normalize %
 & 60.4 & 61.0 & 59.0 & 59.9 & 54.2 & 56.1 & 60.7 & 46.5 \\
\quad+ quantile-u & 67.7 & 70.9 & 63.6 & 65.2 & 61.7 & 56.9 & 61.1 & 52.4 \\
\quad+ quantile-u$^W$ & 62.8 & 65.0 & 61.3 & 63.5 & 59.7 & 54.7 & 60.4 & 50.5 \\
\quad+ whiten & 68.2 & 70.0 & 68.7 & 70.0 & 69.6 & 66.7 & 64.6 & 68.1 \\
\quad+ whiten$^W$ & 62.1 & 65.8 & 62.6 & 67.3 & 69.4 & 65.1 & 64.1 & 67.5 \\
\quad+ zscore & 68.7 & 71.1 & 65.0 & 66.7 & 63.0 & 58.8 & 62.5 & 54.6\vspace{5pt}\\

$\mbox{idf}_t^W$ & 69.0 & 68.2 & 70.3 & 69.6 & 66.4 & 68.0 & 66.9 & 69.8 \\
\quad+ normalize & 69.0 & 68.2 & 70.3 & 69.6 & 66.4 & 68.0 & 66.9 & 69.8 \\
\quad+ quantile-u & 69.4 & 68.6 & 69.3 & 69.6 & 67.8 & 66.7 & 64.7 & 64.4 \\
\quad+ quantile-u$^W$ & 69.4 & 68.8 & 70.2 & 70.3 & 68.1 & 67.0 & 65.3 & 65.7 \\
\quad+ whiten & 65.6 & 65.0 & 67.9 & 68.0 & 67.2 & 65.8 & 64.1 & 66.5 \\
\quad+ whiten$^W$ & 67.0 & 66.6 & \textbf{71.2} & \textbf{72.1} & \textbf{71.5} & \textbf{69.8} & \textbf{68.2} & \textbf{70.4} \\
\quad+ zscore & 70.7 & 69.9 & 71.0 & 71.7 & 70.1 & 69.0 & 67.8 & 70.0\vspace{5pt}\\

$\mbox{idf}_t^T$ & 68.8 & 70.7 & 69.6 & 69.3 & 67.3 & 56.1 & 60.7 & 67.0 \\
\quad+ normalize & 68.8 & 70.7 & 69.6 & 69.3 & 67.3 & 56.1 & 60.7 & 67.0 \\
\quad+ quantile-u & \textbf{72.4} & \underline{\textbf{75.1}} & 69.2 & 69.3 & 67.7 & 56.9 & 61.1 & 64.2 \\
\quad+ quantile-u$^W$ & 70.4 & 73.2 & 69.7 & 69.6 & 67.7 & 54.7 & 60.4 & 65.0 \\
\quad+ whiten & 69.6 & 72.2 & 68.4 & 68.4 & 67.7 & 66.7 & 64.6 & 67.0 \\
\quad+ whiten$^W$ & 67.8 & 71.0 & 69.6 & 69.9 & 68.7 & 65.1 & 64.1 & 67.1 \\
\quad+ zscore & 72.4 & 74.7 & 70.1 & 70.3 & 68.6 & 58.8 & 62.5 & 67.4\vspace{5pt}\\

-biases & 64.7 & 65.4 & 70.2 & 71.0 & 67.8 & 52.9 & 57.3 & 66.6 \\
\quad+ normalize & 64.7 & 65.4 & 70.2 & 71.0 & 67.8 & 52.9 & 57.3 & 66.6 \\
\quad+ quantile-u & 70.1 & 73.4 & 69.9 & 71.0 & 68.9 & 54.2 & 58.1 & 62.8 \\
\quad+ quantile-u$^W$ & 66.1 & 69.2 & 70.3 & 71.5 & 69.0 & 51.7 & 57.1 & 63.9 \\
\quad+ whiten$^W$ & 66.1 & 68.5 & 69.6 & 71.3 & 70.9 & 61.8 & 60.8 & 68.8 \\
\quad+ whiten & 68.1 & 69.9 & 69.5 & 69.9 & 69.3 & 64.6 & 62.4 & 66.5 \\
\quad+ zscore & 71.1 & 73.2 & 70.2 & 71.1 & 68.9 & 56.1 & 59.4 & 67.4\vspace{5pt}\\ 

\verb|[MASK]| & 68.5 & 73.3 &  &  &  &  &  &  \\
\quad+ normalize & 68.5 & 73.3 &  &  &  &  &  &  \\
\quad+ quantile-u & 70.8 & 74.8 &  &  &  &  &  &  \\
\quad+ quantile-u$^W$ & 69.3 & 74.4 &  &  &  &  &  &  \\
\quad+ whiten & 70.5 & 74.3 &  &  &  &  &  &  \\
\quad+ whiten$^W$ & 65.3 & 70.9 &  &  &  &  &  &  \\
\quad+ zscore & 70.5 & 73.7 &  &  &  &  &  &  \\
\bottomrule
\end{tabular}
\end{table}

\vspace{-10pt}
\begin{table}[h]\small
\caption{Spearman %
 correlation dependence on token aggregation and post-processing techniques for semantic textual similarity SICK-R task. The best result for each model is bolded, while underlined result is the best across all the models.}
\label{SICK-R}
\setlength\tabcolsep{5.3pt}
\begin{tabular}{lrrrrrrrr}
\toprule
\textbf{Model} & \textbf{T0} & \textbf{T4} & \textbf{BERT} & \textbf{BERT + Avg.} & \textbf{Avg.} & \textbf{B2S} & \textbf{B2S-100} & \textbf{RE} \\
\cmidrule(r){2-3} \cmidrule(lr){4-6} \cmidrule(l){7-9}
\textbf{Layer} & \multicolumn{2}{c}{\textbf{Last}} & \multicolumn{3}{c}{\textbf{First + Last}} & \multicolumn{3}{c}{\textbf{--}} \\
\midrule
avg. & 64.4 & 64.2 & 63.8 & 63.1 & 60.3 & 60.2 & 61.2 & 53.1 \\
\quad+ normalize %
 & 64.4 & 64.2 & 63.8 & 63.1 & 60.3 & 60.2 & 61.2 & 53.1 \\
\quad+ quantile-u & 66.1 & 66.3 & 64.9 & 64.7 & 62.5 & 61.0 & 61.8 & 54.8 \\
\quad+ quantile-u$^W$ & 64.9 & 64.7 & 64.3 & 64.1 & 61.6 & 59.7 & 61.0 & 53.1 \\
\quad+ whiten & 61.0 & 60.8 & 60.4 & 59.1 & 55.1 & 55.3 & 55.4 & 53.3 \\
\quad+ whiten$^W$ & 63.1 & 63.2 & 63.7 & 63.9 & 61.5 & 61.0 & 59.9 & 58.4 \\
\quad+ zscore & 66.3 & 66.7 & 65.0 & 64.9 & 62.8 & 62.1 & 63.1 & 56.3\vspace{5pt}\\

$\mbox{idf}_t^W$ & 62.5 & 62.0 & 61.7 & 59.7 & 57.5 & 61.9 & 60.1 & 57.4 \\
\quad+ normalize & 62.5 & 62.0 & 61.7 & 59.7 & 57.5 & 61.9 & 60.2 & 57.4 \\
\quad+ quantile-u & 63.0 & 62.6 & 61.8 & 60.4 & 58.5 & 60.8 & 59.4 & 52.9 \\
\quad+ quantile-u$^W$ & 62.9 & 62.5 & 61.8 & 60.5 & 59.0 & 61.1 & 59.5 & 54.7 \\
\quad+ whiten & 57.7 & 57.5 & 57.1 & 55.9 & 53.8 & 54.5 & 54.7 & 52.0 \\
\quad+ whiten$^W$ & 60.9 & 61.1 & 61.9 & 60.8 & 59.0 & 61.2 & 60.1 & 56.7 \\
\quad+ zscore & 63.7 & 63.4 & 63.2 & 62.3 & 60.9 & \textbf{62.9} & 61.3 & 57.3\vspace{5pt}\\

$\mbox{idf}_t^T$ & 64.4 & 65.0 & 62.8 & 61.1 & 59.2 & 60.2 & 61.3 & 56.8 \\
\quad+ normalize & 64.4 & 65.0 & 62.8 & 61.1 & 59.2 & 60.2 & 61.2 & 56.8 \\
\quad+ quantile-u & 66.0 & 66.3 & 63.2 & 61.6 & 59.7 & 61.0 & 61.8 & 54.3 \\
\quad+ quantile-u$^W$ & 65.0 & 65.6 & 63.0 & 61.3 & 59.5 & 59.7 & 61.0 & 55.2 \\
\quad+ whiten & 59.9 & 60.3 & 58.1 & 56.8 & 54.4 & 55.3 & 55.4 & 52.5 \\
\quad+ whiten$^W$ & 63.5 & 64.2 & 62.4 & 61.0 & 58.9 & 61.0 & 59.9 & 56.3 \\
\quad+ zscore & 66.2 & 66.5 & 63.7 & 62.4 & 60.8 & 62.1 & \textbf{63.1} & 57.0\vspace{5pt}\\

-biases & 65.1 & 65.9 & 65.6 & 65.1 & 63.8 & 59.0 & 59.7 & 59.7 \\
\quad+ normalize & 65.1 & 65.9 & 65.6 & 65.1 & 63.8 & 59.0 & 59.7 & 59.7 \\
\quad+ quantile-u & 66.9 & 67.7 & 66.0 & 65.7 & 64.7 & 59.9 & 60.2 & 56.5 \\
\quad+ quantile-u$^W$ & 65.6 & 66.5 & 65.8 & 65.5 & 64.4 & 58.6 & 59.3 & 57.3 \\
\quad+ whiten$^W$ & 64.8 & 65.4 & 64.7 & 63.9 & 61.3 & 59.8 & 58.6 & 58.0 \\
\quad+ whiten & 61.3 & 61.1 & 60.3 & 58.9 & 55.4 & 54.2 & 54.1 & 53.1 \\
\quad+ zscore & 67.1 & \underline{\textbf{67.8}} & \textbf{66.3} & \textbf{66.3} & \textbf{65.6} & 61.1 & 62.0 & \textbf{60.4}\vspace{5pt}\\

\verb|[MASK]| & 64.8 & 65.0 &  &  &  &  &  &  \\
\quad+ normalize & 64.8 & 65.0 &  &  &  &  &  &  \\
\quad+ quantile-u & 67.1 & 66.0 &  &  &  &  &  &  \\
\quad+ quantile-u$^W$ & 65.6 & 65.6 &  &  &  &  &  &  \\
\quad+ whiten & 61.7 & 61.8 &  &  &  &  &  &  \\
\quad+ whiten$^W$ & 63.3 & 64.2 &  &  &  &  &  &  \\
\quad+ zscore & \textbf{67.1} & 66.2 &  &  &  &  &  &  \\
\bottomrule
\end{tabular}
\end{table}
\vspace{-10pt}

\begin{table}[h]\small
\caption{Spearman %
 correlation dependence on token aggregation and post-processing techniques for semantic textual similarity STR task. The best result for each model is bolded, while underlined result is the best across all the models.}
\label{STR}
\setlength\tabcolsep{5.3pt}
\begin{tabular}{lrrrrrrrr}
\toprule
\textbf{Model} & \textbf{T0} & \textbf{T4} & \textbf{BERT} & \textbf{BERT + Avg.} & \textbf{Avg.} & \textbf{B2S} & \textbf{B2S-100} & \textbf{RE} \\
\cmidrule(r){2-3} \cmidrule(lr){4-6} \cmidrule(l){7-9}
\textbf{Layer} & \multicolumn{2}{c}{\textbf{Last}} & \multicolumn{3}{c}{\textbf{First + Last}} & \multicolumn{3}{c}{\textbf{--}} \\
\midrule
avg. & 60.9 & 59.8 & 65.4 & 65.6 & 59.4 & 48.2 & 48.9 & 53.3 \\
\quad+ normalize %
 & 60.9 & 59.8 & 65.4 & 65.6 & 59.4 & 48.2 & 48.9 & 53.3 \\
\quad+ quantile-u & 65.7 & 67.8 & 68.1 & 69.2 & 65.4 & 47.3 & 47.9 & 53.6 \\
\quad+ quantile-u$^W$ & 63.6 & 64.1 & 67.3 & 68.8 & 65.3 & 48.0 & 48.6 & 53.7 \\
\quad+ whiten & 68.9 & 69.6 & 70.1 & 69.7 & 67.0 & 51.4 & 48.8 & 62.1 \\
\quad+ whiten$^W$ & 65.2 & 67.7 & 66.7 & 68.7 & 68.6 & 51.3 & 49.2 & 63.2 \\
\quad+ zscore & 66.6 & 68.9 & 66.7 & 67.5 & 64.5 & 50.1 & 50.1 & 56.0\vspace{5pt}\\

$\mbox{idf}_t^W$ & 68.2 & 67.8 & 71.9 & 70.9 & 66.3 & 53.8 & 51.5 & 62.6 \\
\quad+ normalize & 68.2 & 67.8 & 71.9 & 70.9 & 66.3 & 53.8 & 51.5 & 62.6 \\
\quad+ quantile-u & 69.7 & 69.5 & \textbf{72.5} & 72.7 & 70.0 & 52.4 & 49.7 & 59.7 \\
\quad+ quantile-u$^W$ & 69.2 & 69.0 & 72.5 & \textbf{72.8} & \textbf{70.4} & 53.3 & 50.8 & 61.2 \\
\quad+ whiten & 69.5 & 68.8 & 70.1 & 68.9 & 66.2 & 51.9 & 49.5 & 61.1 \\
\quad+ whiten$^W$ & 68.2 & 67.8 & 70.8 & 70.6 & 67.7 & 52.4 & 49.9 & 61.7 \\
\quad+ zscore & 69.5 & 68.9 & 71.4 & 71.2 & 68.7 & \textbf{54.0} & \textbf{52.2} & 62.4\vspace{5pt}\\

$\mbox{idf}_t^T$ & 67.1 & 67.5 & 71.8 & 70.8 & 66.9 & 48.2 & 48.9 & \textbf{63.3} \\
\quad+ normalize & 67.1 & 67.5 & 71.8 & 70.8 & 66.9 & 48.2 & 48.9 & 63.3 \\
\quad+ quantile-u & \textbf{71.0} & 73.2 & 72.1 & 71.9 & 69.3 & 47.3 & 47.9 & 59.3 \\
\quad+ quantile-u$^W$ & 69.8 & 71.6 & 72.0 & 72.0 & 69.7 & 48.0 & 48.6 & 60.6 \\
\quad+ whiten & 70.4 & 71.1 & 70.4 & 69.0 & 66.3 & 51.4 & 48.8 & 61.4 \\
\quad+ whiten$^W$ & 69.0 & 71.3 & 70.4 & 70.0 & 67.6 & 51.3 & 49.2 & 62.1 \\
\quad+ zscore & 70.8 & 73.2 & 71.4 & 71.3 & 69.3 & 50.1 & 50.1 & 63.1\vspace{5pt}\\

-biases & 62.8 & 62.5 & 70.7 & 70.1 & 64.9 & 46.0 & 45.8 & 60.3 \\
\quad+ normalize & 62.8 & 62.5 & 70.7 & 70.1 & 64.9 & 46.0 & 45.8 & 60.3 \\
\quad+ quantile-u & 66.7 & 69.6 & 72.2 & 72.1 & 68.6 & 45.7 & 45.0 & 58.0 \\
\quad+ quantile-u$^W$ & 65.2 & 67.1 & 71.9 & 71.9 & 68.4 & 46.2 & 45.5 & 58.9 \\
\quad+ whiten$^W$ & 67.8 & 69.3 & 69.2 & 69.2 & 66.4 & 49.4 & 47.0 & 60.1 \\
\quad+ whiten & 69.1 & 69.5 & 69.6 & 68.5 & 65.0 & 49.6 & 46.9 & 59.6 \\
\quad+ zscore & 68.3 & 70.8 & 69.7 & 69.1 & 65.9 & 48.5 & 47.6 & 60.0\vspace{5pt}\\

\verb|[MASK]| & 67.0 & 70.1 &  &  &  &  &  &  \\
\quad+ normalize & 67.0 & 70.1 &  &  &  &  &  &  \\
\quad+ quantile-u & 70.7 & \underline{\textbf{73.5}} &  &  &  &  &  &  \\
\quad+ quantile-u$^W$ & 69.1 & 73.2 &  &  &  &  &  &  \\
\quad+ whiten & 70.5 & 72.3 &  &  &  &  &  &  \\
\quad+ whiten$^W$ & 68.1 & 72.6 &  &  &  &  &  &  \\
\quad+ zscore & 70.7 & 73.1 &  &  &  &  &  &  \\
\bottomrule
\end{tabular}
\end{table}
\vspace{-10pt}

\begin{table}[h]\small
\caption{Clustering %
 accuracy dependence on token aggregation and post-processing techniques for the agnews dataset. The best result for each model is bolded, while underlined result is the best across all the models.}
\label{agnews}
\setlength\tabcolsep{5.3pt}
\begin{tabular}{lrrrrrrrr}
\toprule
\textbf{Model} & \textbf{T0} & \textbf{T4} & \textbf{BERT} & \textbf{BERT + Avg.} & \textbf{Avg.} & \textbf{B2S} & \textbf{B2S-100} & \textbf{RE} \\
\cmidrule(r){2-3} \cmidrule(lr){4-6} \cmidrule(l){7-9}
\textbf{Layer} & \multicolumn{2}{c}{\textbf{Last}} & \multicolumn{3}{c}{\textbf{First + Last}} & \multicolumn{3}{c}{\textbf{--}} \\
\midrule
avg. & 85.4 & 80.7 & 81.2 & 84.4 & 79.0 & 85.0 & 85.0 & 28.0 \\
\quad+ normalize %
 & 85.5 & 80.6 & 85.6 & 86.1 & 79.6 & \textbf{85.7} & 85.9 & 27.2 \\
\quad+ quantile-u & 85.2 & 86.9 & 84.0 & 85.1 & 79.7 & 85.2 & 85.2 & 27.5 \\
\quad+ quantile-u$^W$ & 85.4 & 86.7 & 85.3 & 85.7 & 80.6 & 85.4 & 85.4 & 27.2 \\
\quad+ whiten & 34.3 & 31.3 & 31.3 & 31.5 & 30.5 & 29.9 & 29.9 & 28.8 \\
\quad+ whiten$^W$ & 75.2 & 75.0 & 72.7 & 72.8 & 58.6 & 70.6 & 69.0 & 28.4 \\
\quad+ zscore & 86.0 & 81.2 & 81.3 & 83.8 & 79.8 & 84.8 & 85.2 & 27.7\vspace{5pt}\\

$\mbox{idf}_t^W$ & 79.5 & 81.4 & 80.6 & 84.0 & 78.0 & 83.6 & 83.7 & 39.9 \\
\quad+ normalize & 84.9 & 85.6 & 84.7 & 85.1 & 78.3 & 85.0 & 84.9 & 41.4 \\
\quad+ quantile-u & 85.8 & 85.7 & 81.7 & 84.4 & 78.6 & 84.2 & 84.2 & 41.2 \\
\quad+ quantile-u$^W$ & 85.6 & 85.8 & 81.8 & 84.7 & 79.1 & 84.6 & 84.7 & \textbf{43.4} \\
\quad+ whiten & 30.9 & 32.0 & 31.3 & 31.8 & 30.7 & 30.7 & 30.1 & 27.6 \\
\quad+ whiten$^W$ & 74.5 & 78.1 & 73.8 & 74.4 & 70.9 & 72.5 & 72.1 & 38.4 \\
\quad+ zscore & 85.9 & 81.8 & 80.4 & 84.0 & 78.9 & 83.7 & 83.7 & 41.0\vspace{5pt}\\

$\mbox{idf}_t^T$ & 80.1 & 82.1 & 80.9 & 82.5 & 79.3 & 85.0 & 85.0 & 28.7 \\
\quad+ normalize & 85.2 & 82.2 & 85.1 & 84.8 & 81.7 & 85.7 & \textbf{85.9} & 28.5 \\
\quad+ quantile-u & 85.9 & 87.0 & 82.4 & 81.9 & 80.9 & 85.2 & 85.2 & 28.6 \\
\quad+ quantile-u$^W$ & 85.9 & 87.0 & 82.0 & 84.4 & 82.1 & 85.3 & 85.5 & 28.4 \\
\quad+ whiten & 30.2 & 32.6 & 30.9 & 31.3 & 29.9 & 29.9 & 29.8 & 28.0 \\
\quad+ whiten$^W$ & 72.1 & 68.4 & 73.0 & 71.9 & 45.5 & 70.6 & 69.0 & 28.9 \\
\quad+ zscore & 85.9 & 82.6 & 81.1 & 82.1 & 79.7 & 84.8 & 85.2 & 28.8\vspace{5pt}\\

-biases & 85.7 & 86.4 & 81.4 & 85.6 & 83.4 & 84.0 & 83.8 & 28.5 \\
\quad+ normalize & 85.7 & 86.6 & \textbf{86.8} & \textbf{86.8} & \textbf{83.8} & 85.3 & 85.5 & 28.4 \\
\quad+ quantile-u & 85.4 & 86.7 & 85.7 & 85.4 & 83.4 & 84.7 & 84.7 & 28.5 \\
\quad+ quantile-u$^W$ & 85.6 & 87.1 & 86.1 & 86.0 & 83.6 & 84.6 & 84.9 & 29.4 \\
\quad+ whiten$^W$ & 73.5 & 73.4 & 72.4 & 72.1 & 55.5 & 68.8 & 67.6 & 28.2 \\
\quad+ whiten & 31.1 & 31.5 & 33.0 & 30.4 & 30.3 & 29.5 & 29.8 & 28.3 \\
\quad+ zscore & \textbf{86.0} & \underline{\textbf{87.1}} & 81.7 & 84.8 & 83.5 & 84.4 & 84.5 & 28.1\vspace{5pt}\\

\verb|[MASK]| & 74.8 & 81.9 &  &  &  &  &  &  \\
\quad+ normalize & 75.1 & 82.2 &  &  &  &  &  &  \\
\quad+ quantile-u & 63.1 & 82.1 &  &  &  &  &  &  \\
\quad+ quantile-u$^W$ & 76.1 & 84.2 &  &  &  &  &  &  \\
\quad+ whiten & 30.9 & 31.8 &  &  &  &  &  &  \\
\quad+ whiten$^W$ & 52.9 & 48.3 &  &  &  &  &  &  \\
\quad+ zscore & 76.1 & 81.7 &  &  &  &  &  &  \\
\bottomrule
\end{tabular}
\end{table}

\vspace{-10pt}
\begin{table}[h]\small
\caption{Clustering %
 accuracy dependence on token aggregation and post-processing techniques for the biomedical dataset. The best result for each model is bolded, while underlined result is the best across all the models.}
\label{biomedical}
\setlength\tabcolsep{5.3pt}
\begin{tabular}{lrrrrrrrr}
\toprule
\textbf{Model} & \textbf{T0} & \textbf{T4} & \textbf{BERT} & \textbf{BERT + Avg.} & \textbf{Avg.} & \textbf{B2S} & \textbf{B2S-100} & \textbf{RE} \\
\cmidrule(r){2-3} \cmidrule(lr){4-6} \cmidrule(l){7-9}
\textbf{Layer} & \multicolumn{2}{c}{\textbf{Last}} & \multicolumn{3}{c}{\textbf{First + Last}} & \multicolumn{3}{c}{\textbf{--}} \\
\midrule
avg. & 32.7 & 32.4 & 34.8 & 36.1 & 30.7 & 33.1 & 37.0 & 19.4 \\
\quad+ normalize %
 & 32.8 & 32.0 & 35.1 & 36.3 & 30.3 & 33.8 & \textbf{37.6} & 20.7 \\
\quad+ quantile-u & 33.6 & 32.5 & 37.0 & 39.1 & 33.1 & 32.9 & 36.7 & 21.0 \\
\quad+ quantile-u$^W$ & 33.4 & 32.5 & 35.4 & 37.9 & 34.5 & 33.0 & 37.0 & 21.1 \\
\quad+ whiten & 29.4 & 30.2 & 27.2 & 23.2 & 18.2 & 12.7 & 13.2 & 15.9 \\
\quad+ whiten$^W$ & 34.8 & 34.7 & 40.2 & \underline{\textbf{42.7}} & 36.5 & 35.0 & 34.4 & 29.4 \\
\quad+ zscore & 33.5 & 33.0 & 36.1 & 37.4 & 32.1 & 32.4 & 36.1 & 20.0\vspace{5pt}\\

$\mbox{idf}_t^W$ & 31.0 & 30.0 & 32.3 & 32.3 & 28.1 & 34.5 & 34.5 & 32.1 \\
\quad+ normalize & 31.2 & 30.2 & 32.5 & 32.6 & 28.7 & 34.7 & 34.8 & \textbf{35.2} \\
\quad+ quantile-u & 32.5 & 31.3 & 33.6 & 33.6 & 29.8 & 33.9 & 35.0 & 32.4 \\
\quad+ quantile-u$^W$ & 31.7 & 30.7 & 31.9 & 33.6 & 30.7 & 33.7 & 34.4 & 33.4 \\
\quad+ whiten & 27.3 & 27.6 & 26.4 & 20.6 & 17.1 & 11.8 & 12.4 & 16.7 \\
\quad+ whiten$^W$ & 32.8 & 32.1 & 37.0 & 38.7 & 35.4 & \textbf{37.2} & 37.2 & 30.8 \\
\quad+ zscore & 32.0 & 30.8 & 32.6 & 32.5 & 28.2 & 34.3 & 34.3 & 31.8\vspace{5pt}\\

$\mbox{idf}_t^T$ & 33.4 & 33.6 & 36.0 & 36.2 & 32.6 & 33.1 & 37.0 & 30.0 \\
\quad+ normalize & 33.6 & 34.1 & 36.0 & 37.1 & 33.9 & 33.8 & 37.6 & 32.2 \\
\quad+ quantile-u & 35.4 & 35.3 & 36.9 & 37.2 & 34.1 & 32.9 & 36.7 & 31.3 \\
\quad+ quantile-u$^W$ & 35.0 & 35.7 & 36.6 & 37.5 & 34.7 & 33.0 & 37.0 & 32.2 \\
\quad+ whiten & 28.3 & 30.8 & 20.9 & 19.4 & 16.3 & 12.7 & 13.2 & 15.2 \\
\quad+ whiten$^W$ & 36.8 & 37.4 & 40.4 & 39.9 & 33.1 & 35.0 & 34.4 & 29.2 \\
\quad+ zscore & 34.7 & 35.4 & 36.9 & 36.8 & 32.4 & 32.4 & 36.1 & 29.5\vspace{5pt}\\

-biases & 35.1 & 32.2 & 39.0 & 39.2 & 37.2 & 31.6 & 35.9 & 29.2 \\
\quad+ normalize & 35.2 & 32.5 & 39.4 & 39.4 & \textbf{37.7} & 32.9 & 36.9 & 30.8 \\
\quad+ quantile-u & 36.0 & 33.0 & 39.8 & 39.6 & 37.2 & 32.3 & 36.2 & 29.9 \\
\quad+ quantile-u$^W$ & 36.6 & 34.6 & 39.6 & 39.9 & 37.3 & 32.4 & 36.3 & 30.9 \\
\quad+ whiten$^W$ & \textbf{38.6} & \textbf{38.7} & \textbf{41.7} & 40.3 & 34.4 & 34.4 & 33.4 & 28.7 \\
\quad+ whiten & 29.8 & 31.6 & 22.0 & 17.1 & 13.5 & 12.5 & 12.0 & 13.2 \\
\quad+ zscore & 35.6 & 32.7 & 39.4 & 39.5 & 37.0 & 30.9 & 35.6 & 28.7\vspace{5pt}\\

\verb|[MASK]| & 24.0 & 34.8 &  &  &  &  &  &  \\
\quad+ normalize & 25.1 & 35.3 &  &  &  &  &  &  \\
\quad+ quantile-u & 27.4 & 35.9 &  &  &  &  &  &  \\
\quad+ quantile-u$^W$ & 25.1 & 34.7 &  &  &  &  &  &  \\
\quad+ whiten & 21.9 & 28.4 &  &  &  &  &  &  \\
\quad+ whiten$^W$ & 23.7 & 32.8 &  &  &  &  &  &  \\
\quad+ zscore & 26.4 & 35.0 &  &  &  &  &  &  \\
\bottomrule
\end{tabular}
\end{table}

\vspace{-10pt}
\begin{table}[h]\small
\caption{Clustering %
 accuracy dependence on token aggregation and post-processing techniques for the googleTS dataset. The best result for each model is bolded, while underlined result is the best across all the models.}
\label{googleTS}
\setlength\tabcolsep{5.3pt}
\begin{tabular}{lrrrrrrrr}
\toprule
\textbf{Model} & \textbf{T0} & \textbf{T4} & \textbf{BERT} & \textbf{BERT + Avg.} & \textbf{Avg.} & \textbf{B2S} & \textbf{B2S-100} & \textbf{RE} \\
\cmidrule(r){2-3} \cmidrule(lr){4-6} \cmidrule(l){7-9}
\textbf{Layer} & \multicolumn{2}{c}{\textbf{Last}} & \multicolumn{3}{c}{\textbf{First + Last}} & \multicolumn{3}{c}{\textbf{--}} \\
\midrule
avg. & 63.5 & 64.3 & 66.1 & 66.1 & 65.1 & 66.2 & 65.8 & 61.2 \\
\quad+ normalize %
 & 62.7 & 64.6 & 66.2 & 67.1 & 66.2 & \textbf{68.3} & \textbf{67.7} & 67.1 \\
\quad+ quantile-u & 64.8 & 65.1 & 67.1 & 67.4 & 67.3 & 66.3 & 66.9 & 63.5 \\
\quad+ quantile-u$^W$ & 63.9 & 64.7 & 66.4 & 68.0 & 66.7 & 66.9 & 67.2 & 63.6 \\
\quad+ whiten & 59.2 & 60.3 & 57.5 & 57.3 & 54.5 & 58.1 & 58.4 & 53.1 \\
\quad+ whiten$^W$ & 62.4 & 62.0 & 65.6 & 66.1 & 65.2 & 65.8 & 65.8 & 61.7 \\
\quad+ zscore & 63.6 & 65.0 & 66.7 & 67.1 & 65.9 & 66.0 & 66.6 & 61.8\vspace{5pt}\\

$\mbox{idf}_t^W$ & 63.5 & 62.6 & 65.2 & 66.2 & 65.6 & 64.5 & 65.8 & 63.8 \\
\quad+ normalize & 64.3 & 63.1 & 66.0 & 66.5 & 66.0 & 66.5 & 66.7 & \underline{\textbf{69.5}} \\
\quad+ quantile-u & 64.0 & 65.4 & 66.6 & 66.5 & 67.0 & 66.4 & 66.1 & 64.1 \\
\quad+ quantile-u$^W$ & 64.4 & 64.9 & 67.5 & 67.8 & 66.6 & 67.4 & 65.9 & 63.4 \\
\quad+ whiten & 57.9 & 59.9 & 58.3 & 56.5 & 55.2 & 58.9 & 58.6 & 51.4 \\
\quad+ whiten$^W$ & 61.9 & 61.4 & 66.2 & 67.7 & 66.7 & 65.5 & 66.7 & 63.3 \\
\quad+ zscore & 63.5 & 63.3 & 66.1 & 66.4 & 65.7 & 65.9 & 65.6 & 63.9\vspace{5pt}\\

$\mbox{idf}_t^T$ & 63.2 & 64.9 & 66.4 & 67.4 & 65.9 & 66.2 & 66.0 & 61.3 \\
\quad+ normalize & 64.7 & 64.7 & 67.7 & \textbf{68.6} & 65.9 & 68.3 & 67.7 & 67.4 \\
\quad+ quantile-u & \textbf{65.2} & 65.5 & \textbf{68.5} & 67.9 & 67.5 & 66.6 & 67.0 & 63.7 \\
\quad+ quantile-u$^W$ & 64.8 & \textbf{66.2} & 67.7 & 67.3 & 67.3 & 67.1 & 67.2 & 62.9 \\
\quad+ whiten & 58.8 & 59.2 & 58.7 & 59.0 & 55.6 & 57.7 & 58.5 & 55.1 \\
\quad+ whiten$^W$ & 60.1 & 60.4 & 65.1 & 66.7 & 64.8 & 65.8 & 65.8 & 61.2 \\
\quad+ zscore & 63.8 & 65.0 & 66.4 & 67.6 & 66.3 & 66.0 & 66.3 & 62.1\vspace{5pt}\\  

-biases & 63.2 & 64.1 & 66.0 & 66.3 & 65.7 & 65.7 & 65.3 & 62.0 \\
\quad+ normalize & 62.6 & 63.4 & 66.3 & 66.8 & 65.8 & 67.0 & 66.8 & 67.1 \\
\quad+ quantile-u & 64.8 & 65.1 & 66.7 & 68.0 & \textbf{67.7} & 65.7 & 65.4 & 64.0 \\
\quad+ quantile-u$^W$ & 64.4 & 65.5 & 67.1 & 67.4 & 66.7 & 66.5 & 66.7 & 64.1 \\
\quad+ whiten$^W$ & 60.9 & 61.6 & 66.0 & 66.4 & 65.2 & 63.9 & 65.5 & 63.2 \\
\quad+ whiten & 61.3 & 60.3 & 59.2 & 59.2 & 57.6 & 57.3 & 57.1 & 56.9 \\
\quad+ zscore & 63.8 & 64.9 & 65.7 & 66.7 & 66.1 & 65.3 & 65.0 & 62.7\vspace{5pt}\\

\verb|[MASK]| & 45.5 & 56.0 &  &  &  &  &  &  \\
\quad+ normalize & 45.9 & 56.5 &  &  &  &  &  &  \\
\quad+ quantile-u & 47.8 & 56.2 &  &  &  &  &  &  \\
\quad+ quantile-u$^W$ & 46.9 & 56.4 &  &  &  &  &  &  \\
\quad+ whiten & 53.4 & 53.1 &  &  &  &  &  &  \\
\quad+ whiten$^W$ & 43.0 & 53.3 &  &  &  &  &  &  \\
\quad+ zscore & 47.0 & 56.0 &  &  &  &  &  &  \\
\bottomrule
\end{tabular}
\end{table}

\vspace{-10pt}
\begin{table}[h]\small
\caption{Clustering %
 accuracy dependence on token aggregation and post-processing techniques for the searchsnippets dataset. The best result for each model is bolded, while underlined result is the best across all the models.}
\label{searchsnippets}
\setlength\tabcolsep{5.3pt}
\begin{tabular}{lrrrrrrrr}
\toprule
\textbf{Model} & \textbf{T0} & \textbf{T4} & \textbf{BERT} & \textbf{BERT + Avg.} & \textbf{Avg.} & \textbf{B2S} & \textbf{B2S-100} & \textbf{RE} \\
\cmidrule(r){2-3} \cmidrule(lr){4-6} \cmidrule(l){7-9}
\textbf{Layer} & \multicolumn{2}{c}{\textbf{Last}} & \multicolumn{3}{c}{\textbf{First + Last}} & \multicolumn{3}{c}{\textbf{--}} \\
\midrule
avg. & 67.3 & 73.3 & 72.2 & 70.8 & 63.6 & 72.0 & 72.0 & 23.3 \\
\quad+ normalize %
 & 68.2 & 73.8 & 81.7 & 73.1 & 65.7 & 76.4 & 76.3 & 20.4 \\
\quad+ quantile-u & 69.8 & 72.4 & 74.1 & 73.0 & 66.8 & 75.2 & 72.9 & 22.3 \\
\quad+ quantile-u$^W$ & 69.7 & 73.5 & 81.6 & 74.7 & 68.6 & 75.7 & 75.1 & 21.7 \\
\quad+ whiten & 33.8 & 36.1 & 32.3 & 31.7 & 26.3 & 22.9 & 24.8 & 21.2 \\
\quad+ whiten$^W$ & 47.7 & 49.2 & 58.3 & 61.8 & 56.3 & 58.8 & 58.3 & 24.0 \\
\quad+ zscore & 67.9 & 73.2 & 72.7 & 71.3 & 63.7 & 72.3 & 72.1 & 22.8\vspace{5pt}\\

$\mbox{idf}_t^W$ & 67.1 & 67.0 & 71.9 & 70.4 & 63.0 & 69.9 & 69.9 & 29.2 \\
\quad+ normalize & 68.4 & 69.4 & 81.8 & 73.0 & 65.2 & \textbf{81.4} & \textbf{81.4} & \textbf{36.6} \\
\quad+ quantile-u & 70.3 & 71.6 & 73.8 & 72.5 & 65.9 & 74.4 & 72.2 & 30.7 \\
\quad+ quantile-u$^W$ & 70.8 & 72.5 & 80.6 & 74.0 & 67.1 & 74.8 & 80.7 & 32.1 \\
\quad+ whiten & 32.7 & 34.8 & 33.0 & 31.7 & 23.7 & 27.3 & 24.1 & 20.5 \\
\quad+ whiten$^W$ & 47.8 & 51.9 & 64.5 & 66.1 & 61.9 & 67.6 & 68.5 & 30.0 \\
\quad+ zscore & 69.8 & 70.0 & 72.2 & 70.4 & 62.7 & 70.1 & 70.2 & 29.7\vspace{5pt}\\

$\mbox{idf}_t^T$ & 55.1 & 75.3 & 72.0 & 70.5 & 59.3 & 72.0 & 72.0 & 26.1 \\
\quad+ normalize & 63.2 & 75.6 & 81.2 & 76.9 & 60.9 & 76.4 & 76.3 & 24.8 \\
\quad+ quantile-u & 68.8 & 78.5 & 74.6 & 76.1 & 59.1 & 75.2 & 72.9 & 26.1 \\
\quad+ quantile-u$^W$ & \textbf{71.2} & 78.7 & 81.3 & \textbf{77.0} & 61.0 & 75.7 & 75.1 & 25.8 \\
\quad+ whiten & 30.6 & 35.7 & 30.8 & 29.8 & 25.0 & 22.9 & 24.8 & 21.3 \\
\quad+ whiten$^W$ & 46.8 & 46.1 & 62.7 & 63.0 & 52.9 & 58.8 & 58.3 & 26.7 \\
\quad+ zscore & 68.6 & 79.1 & 73.3 & 72.4 & 59.0 & 72.3 & 72.1 & 25.8\vspace{5pt}\\

-biases & 66.5 & 70.4 & 82.6 & 73.0 & 71.7 & 70.5 & 71.9 & 22.8 \\
\quad+ normalize & 67.0 & 70.5 & \underline{\textbf{82.9}} & 74.8 & 73.8 & 75.2 & 80.9 & 22.6 \\
\quad+ quantile-u & 68.8 & 80.3 & 80.7 & 73.8 & 72.3 & 73.5 & 73.8 & 23.8 \\
\quad+ quantile-u$^W$ & 69.8 & \textbf{80.6} & 82.6 & 76.5 & \textbf{74.2} & 75.6 & 72.7 & 22.9 \\
\quad+ whiten$^W$ & 48.4 & 46.0 & 64.5 & 64.4 & 54.9 & 57.1 & 53.6 & 23.9 \\
\quad+ whiten & 31.6 & 36.3 & 33.2 & 29.3 & 22.7 & 23.3 & 23.9 & 22.0 \\
\quad+ zscore & 67.6 & 79.8 & 82.1 & 72.6 & 71.4 & 72.6 & 72.6 & 23.3\vspace{5pt}\\  

\verb|[MASK]| & 61.1 & 69.9 &  &  &  &  &  &  \\
\quad+ normalize & 61.0 & 69.6 &  &  &  &  &  &  \\
\quad+ quantile-u & 60.3 & 69.6 &  &  &  &  &  &  \\
\quad+ quantile-u$^W$ & 61.4 & 69.8 &  &  &  &  &  &  \\
\quad+ whiten & 31.1 & 34.0 &  &  &  &  &  &  \\
\quad+ whiten$^W$ & 39.4 & 45.3 &  &  &  &  &  &  \\
\quad+ zscore & 61.8 & 69.8 &  &  &  &  &  &  \\
\bottomrule
\end{tabular}
\end{table}

\vspace{-10pt}
\begin{table}[h]\small
\caption{Clustering %
 accuracy dependence on token aggregation and post-processing techniques for the stackoverflow dataset. The best result for each model is bolded, while underlined result is the best across all the models.}
\label{stackoverflow}
\setlength\tabcolsep{5.3pt}
\begin{tabular}{lrrrrrrrr}
\toprule
\textbf{Model} & \textbf{T0} & \textbf{T4} & \textbf{BERT} & \textbf{BERT + Avg.} & \textbf{Avg.} & \textbf{B2S} & \textbf{B2S-100} & \textbf{RE} \\
\cmidrule(r){2-3} \cmidrule(lr){4-6} \cmidrule(l){7-9}
\textbf{Layer} & \multicolumn{2}{c}{\textbf{Last}} & \multicolumn{3}{c}{\textbf{First + Last}} & \multicolumn{3}{c}{\textbf{--}} \\
\midrule
avg. & 25.0 & 30.9 & 35.9 & 43.5 & 41.9 & 14.8 & 20.2 & 39.2 \\
\quad+ normalize %
 & 25.4 & 30.7 & 36.9 & 42.8 & 42.3 & 15.0 & 20.7 & 39.0 \\
\quad+ quantile-u & 27.8 & 32.7 & 38.3 & 45.9 & 45.1 & 14.7 & 20.5 & 38.6 \\
\quad+ quantile-u$^W$ & 25.7 & 32.4 & 38.4 & 46.9 & 47.9 & 15.0 & 20.4 & 40.0 \\
\quad+ whiten & 39.8 & 39.7 & 34.2 & 17.8 & 13.0 & 9.8 & 12.6 & 12.2 \\
\quad+ whiten$^W$ & 28.9 & 33.9 & 49.4 & 57.3 & 62.0 & \textbf{21.7} & \textbf{23.5} & 56.6 \\
\quad+ zscore & 26.7 & 33.1 & 37.3 & 44.2 & 43.6 & 14.0 & 19.8 & 39.2\vspace{5pt}\\

$\mbox{idf}_t^W$ & 28.6 & 29.8 & 30.6 & 37.7 & 39.1 & 16.1 & 17.5 & 55.4 \\
\quad+ normalize & 29.0 & 30.3 & 31.1 & 37.2 & 38.0 & 16.8 & 18.6 & 62.7 \\
\quad+ quantile-u & 29.8 & 31.1 & 32.0 & 38.1 & 40.7 & 16.8 & 17.5 & 57.7 \\
\quad+ quantile-u$^W$ & 30.0 & 30.9 & 32.1 & 39.9 & 40.8 & 16.3 & 18.0 & 58.8 \\
\quad+ whiten & 30.6 & 30.1 & 26.1 & 15.4 & 12.2 & 9.7 & 9.7 & 12.5 \\
\quad+ whiten$^W$ & 29.8 & 31.4 & 40.0 & 50.3 & 55.8 & 16.2 & 17.1 & 56.8 \\
\quad+ zscore & 30.2 & 30.6 & 31.1 & 37.4 & 38.7 & 16.1 & 16.8 & 54.1\vspace{5pt}\\

$\mbox{idf}_t^T$ & 36.2 & 39.0 & 54.2 & 63.4 & 60.5 & 14.8 & 20.2 & 61.6 \\
\quad+ normalize & 36.9 & 38.6 & 55.1 & \textbf{66.6} & 62.4 & 15.0 & 20.7 & \underline{\textbf{70.6}} \\
\quad+ quantile-u & 38.7 & 40.7 & 57.1 & 64.3 & \textbf{63.1} & 14.7 & 20.5 & 63.7 \\
\quad+ quantile-u$^W$ & 38.6 & 40.0 & 57.7 & 65.6 & 63.0 & 15.0 & 20.4 & 65.2 \\
\quad+ whiten & \textbf{43.5} & \textbf{44.6} & 32.1 & 19.7 & 16.6 & 9.8 & 12.6 & 16.7 \\
\quad+ whiten$^W$ & 35.7 & 36.7 & \textbf{59.9} & 63.1 & 58.7 & 21.7 & 23.5 & 61.6 \\
\quad+ zscore & 38.3 & 40.6 & 54.5 & 62.0 & 58.9 & 14.0 & 19.8 & 60.7\vspace{5pt}\\

-biases & 28.5 & 34.5 & 44.8 & 56.7 & 59.7 & 14.3 & 18.2 & 53.9 \\
\quad+ normalize & 28.4 & 35.3 & 45.2 & 57.0 & 58.7 & 14.6 & 19.4 & 63.6 \\
\quad+ quantile-u & 31.0 & 36.8 & 44.5 & 56.7 & 59.3 & 14.0 & 19.0 & 58.0 \\
\quad+ quantile-u$^W$ & 30.8 & 35.5 & 46.4 & 59.9 & 60.6 & 14.8 & 19.3 & 57.4 \\
\quad+ whiten$^W$ & 35.1 & 36.1 & 55.4 & 62.8 & 61.0 & 21.0 & 20.9 & 52.2 \\
\quad+ whiten & 42.6 & 38.7 & 21.4 & 15.7 & 11.3 & 8.0 & 11.6 & 11.3 \\
\quad+ zscore & 31.5 & 36.9 & 45.2 & 57.5 & 57.9 & 13.4 & 18.5 & 53.0\vspace{5pt}\\

\verb|[MASK]| & 26.4 & 35.5 &  &  &  &  &  &  \\
\quad+ normalize & 26.5 & 36.1 &  &  &  &  &  &  \\
\quad+ quantile-u & 26.0 & 37.7 &  &  &  &  &  &  \\
\quad+ quantile-u$^W$ & 24.9 & 37.0 &  &  &  &  &  &  \\
\quad+ whiten & 22.7 & 41.0 &  &  &  &  &  &  \\
\quad+ whiten$^W$ & 18.0 & 32.2 &  &  &  &  &  &  \\
\quad+ zscore & 27.3 & 37.4 &  &  &  &  &  &  \\
\bottomrule
\end{tabular}
\end{table}

\begin{table}[h]\small
\caption{Clustering %
 accuracy dependence on token aggregation and post-processing techniques for the tweet dataset. The best result for each model is bolded, while underlined result is the best across all the models.}
\label{tweet}
\setlength\tabcolsep{5.3pt}
\begin{tabular}{lrrrrrrrr}
\toprule
\textbf{Model} & \textbf{T0} & \textbf{T4} & \textbf{BERT} & \textbf{BERT + Avg.} & \textbf{Avg.} & \textbf{B2S} & \textbf{B2S-100} & \textbf{RE} \\
\cmidrule(r){2-3} \cmidrule(lr){4-6} \cmidrule(l){7-9}
\textbf{Layer} & \multicolumn{2}{c}{\textbf{Last}} & \multicolumn{3}{c}{\textbf{First + Last}} & \multicolumn{3}{c}{\textbf{--}} \\
\midrule
avg. & 47.0 & 48.2 & 51.8 & 54.1 & 50.7 & 51.7 & 51.7 & 46.5 \\
\quad+ normalize %
 & 46.2 & 48.3 & 53.0 & 53.2 & 52.4 & \textbf{53.7} & \textbf{55.2} & 56.4 \\
\quad+ quantile-u & 46.6 & 48.5 & 51.9 & 53.9 & 51.7 & 50.8 & 52.9 & 48.2 \\
\quad+ quantile-u$^W$ & 46.9 & 48.0 & 53.6 & 53.4 & 53.0 & 51.4 & 51.1 & 47.8 \\
\quad+ whiten & 15.9 & 16.3 & 15.9 & 17.0 & 18.5 & 18.1 & 17.2 & 17.6 \\
\quad+ whiten$^W$ & 44.9 & 44.1 & 51.1 & 51.2 & 51.7 & 50.5 & 49.9 & 49.0 \\
\quad+ zscore & 47.5 & 49.0 & 53.0 & 53.3 & 51.4 & 50.0 & 50.5 & 46.4\vspace{5pt}\\

$\mbox{idf}_t^W$ & 48.6 & 47.2 & 50.5 & 52.1 & 48.5 & 49.4 & 50.8 & 45.4 \\
\quad+ normalize & 48.3 & 47.8 & 51.9 & 52.7 & 48.5 & 50.3 & 50.4 & 51.5 \\
\quad+ quantile-u & 49.0 & 47.9 & 52.7 & 51.9 & 48.8 & 49.8 & 49.9 & 46.4 \\
\quad+ quantile-u$^W$ & 48.3 & 47.9 & 52.7 & 52.8 & 48.3 & 49.4 & 49.6 & 46.0 \\
\quad+ whiten & 15.3 & 15.1 & 15.7 & 16.2 & 18.3 & 17.4 & 17.0 & 15.9 \\
\quad+ whiten$^W$ & 43.0 & 42.3 & 50.0 & 50.8 & 50.1 & 49.8 & 49.3 & 45.5 \\
\quad+ zscore & 49.3 & 46.7 & 52.2 & 51.4 & 48.0 & 48.9 & 49.1 & 46.3\vspace{5pt}\\

$\mbox{idf}_t^T$ & \textbf{50.8} & 50.5 & 55.0 & 55.2 & \textbf{54.5} & 51.7 & 51.7 & 49.0 \\
\quad+ normalize & 49.6 & 50.6 & \textbf{55.1} & 54.7 & 53.6 & 53.7 & 55.2 & \underline{\textbf{58.5}} \\
\quad+ quantile-u & 49.7 & 51.0 & 53.5 & 54.3 & 53.7 & 50.8 & 52.9 & 50.5 \\
\quad+ quantile-u$^W$ & 50.0 & \textbf{52.3} & 53.4 & 54.8 & 53.3 & 51.4 & 51.1 & 51.7 \\
\quad+ whiten & 17.6 & 18.0 & 15.5 & 16.5 & 18.2 & 18.1 & 17.2 & 17.6 \\
\quad+ whiten$^W$ & 44.4 & 46.6 & 51.9 & 53.2 & 54.0 & 50.5 & 49.9 & 48.8 \\
\quad+ zscore & 49.7 & 50.0 & 53.5 & 55.4 & 53.8 & 50.0 & 50.5 & 47.8\vspace{5pt}\\

-biases & 47.1 & 49.6 & 52.4 & 52.3 & 51.7 & 49.6 & 49.6 & 48.0 \\
\quad+ normalize & 47.3 & 49.3 & 54.0 & \textbf{55.7} & 53.4 & 51.3 & 51.3 & 55.1 \\
\quad+ quantile-u & 47.9 & 50.4 & 53.0 & 55.6 & 53.4 & 48.6 & 50.7 & 48.2 \\
\quad+ quantile-u$^W$ & 48.8 & 50.0 & 53.4 & 54.4 & 53.5 & 50.0 & 49.0 & 47.4 \\
\quad+ whiten$^W$ & 45.9 & 46.0 & 52.2 & 53.0 & 53.3 & 47.3 & 47.5 & 47.4 \\
\quad+ whiten & 16.3 & 16.9 & 16.4 & 17.4 & 18.7 & 18.7 & 17.7 & 18.5 \\
\quad+ zscore & 47.3 & 49.4 & 52.0 & 53.5 & 53.2 & 48.3 & 49.0 & 47.4\vspace{5pt}\\

\verb|[MASK]| & 41.1 & 47.0 &  &  &  &  &  &  \\
\quad+ normalize & 41.0 & 47.1 &  &  &  &  &  &  \\
\quad+ quantile-u & 42.0 & 47.7 &  &  &  &  &  &  \\
\quad+ quantile-u$^W$ & 41.9 & 47.4 &  &  &  &  &  &  \\
\quad+ whiten & 14.9 & 16.6 &  &  &  &  &  &  \\
\quad+ whiten$^W$ & 38.4 & 41.7 &  &  &  &  &  &  \\
\quad+ zscore & 41.6 & 46.7 &  &  &  &  &  &  \\
\bottomrule
\end{tabular}
\end{table}

\FloatBarrier

\subsection{BERT + Avg. Model in Different Layers}\label{appendix_bert_avg}

\vspace{-6pt}
\begin{figure}[h]
\centering{\includegraphics[width=\textwidth]{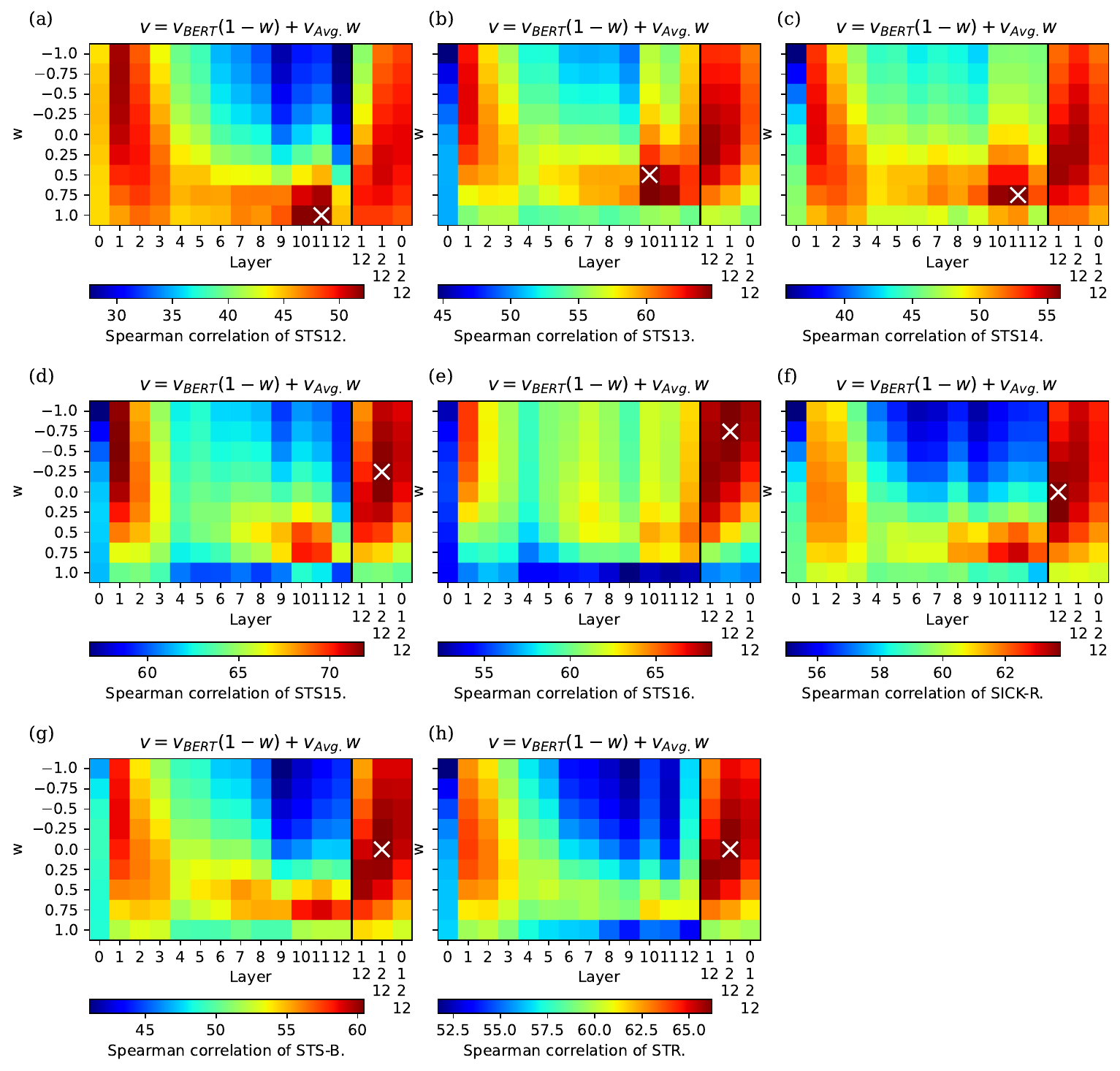}}
\caption{BERT + Avg. model individual task STS performance dependence on the weight $w$ of Avg. model and layer, from which (for both models) representations are used. To the right of the black line on the horizontal axis, average aggregation of multiple layers is also shown. Tokens are simply averaged and no post-processing is used. The horizontal line with $w=0.0$ corresponds to a regular Bert (B) model, $w=0.5$ is B + Avg., and $w=1.0$ is the Avg. model. The white $\times$ marks the maximum value.}
\label{weights_times_layers_sts}
\centering
\end{figure}

\begin{figure}[h]
\centering %
{\includegraphics[width=\textwidth]{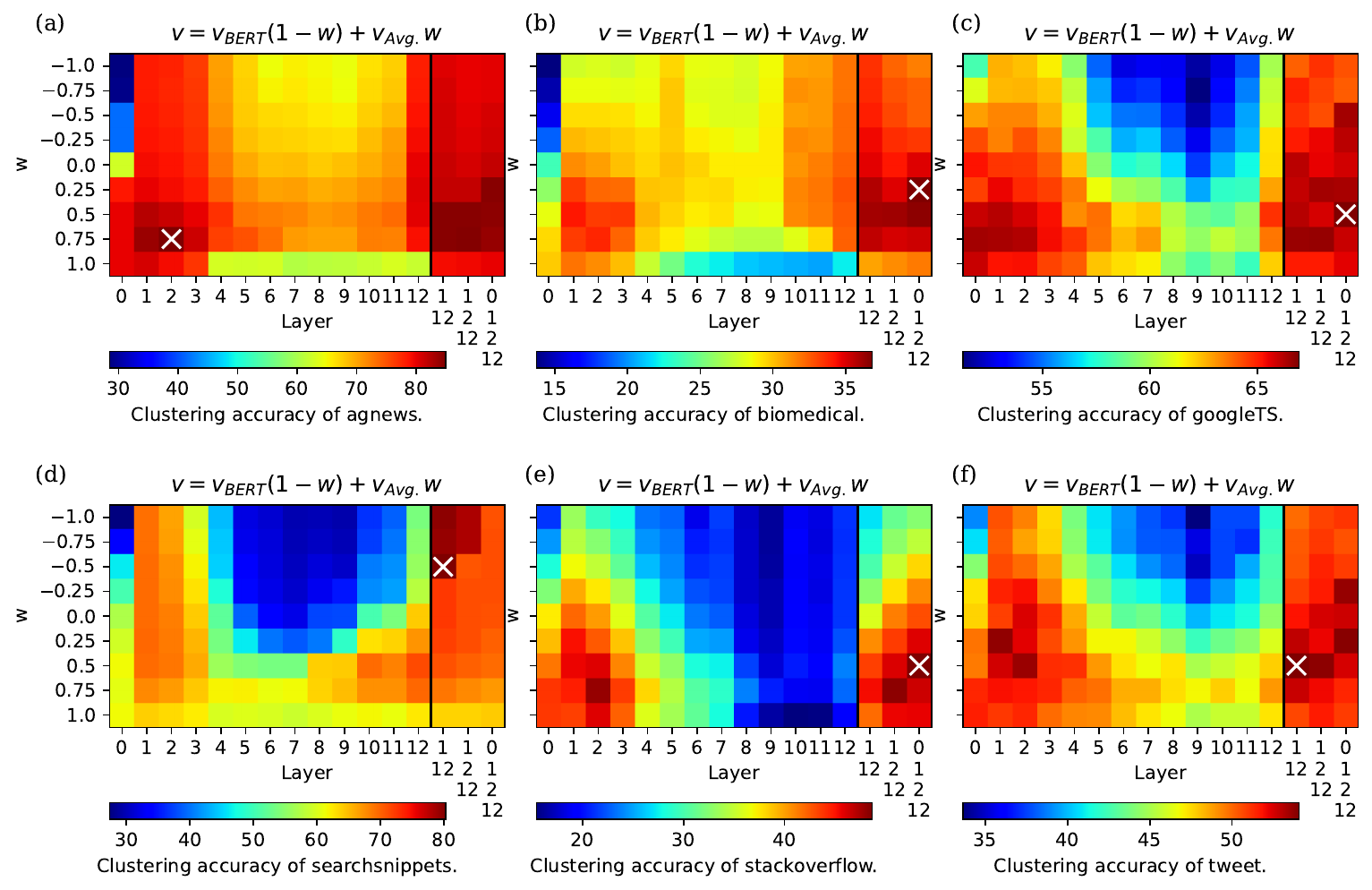}}
\caption{BERT + Avg. %
 model individual task clustering performance dependence on the weight $w$ of Avg. model and layer, from which (for both models) representations are used. To the right of the black line on the horizontal axis, average aggregation of multiple layers is also shown. Tokens are simply averaged and no post-processing is used. The horizontal line with $w=0.0$ corresponds to a regular Bert (B) model, $w=0.5$ is B + Avg., and $w=1.0$ is the Avg. model. The white $\times$ marks the maximum value.}
\label{weights_times_layers_clustering}
\centering
\end{figure}

\begin{figure}[h]
\hspace{-1pt}{\includegraphics[width=\textwidth]{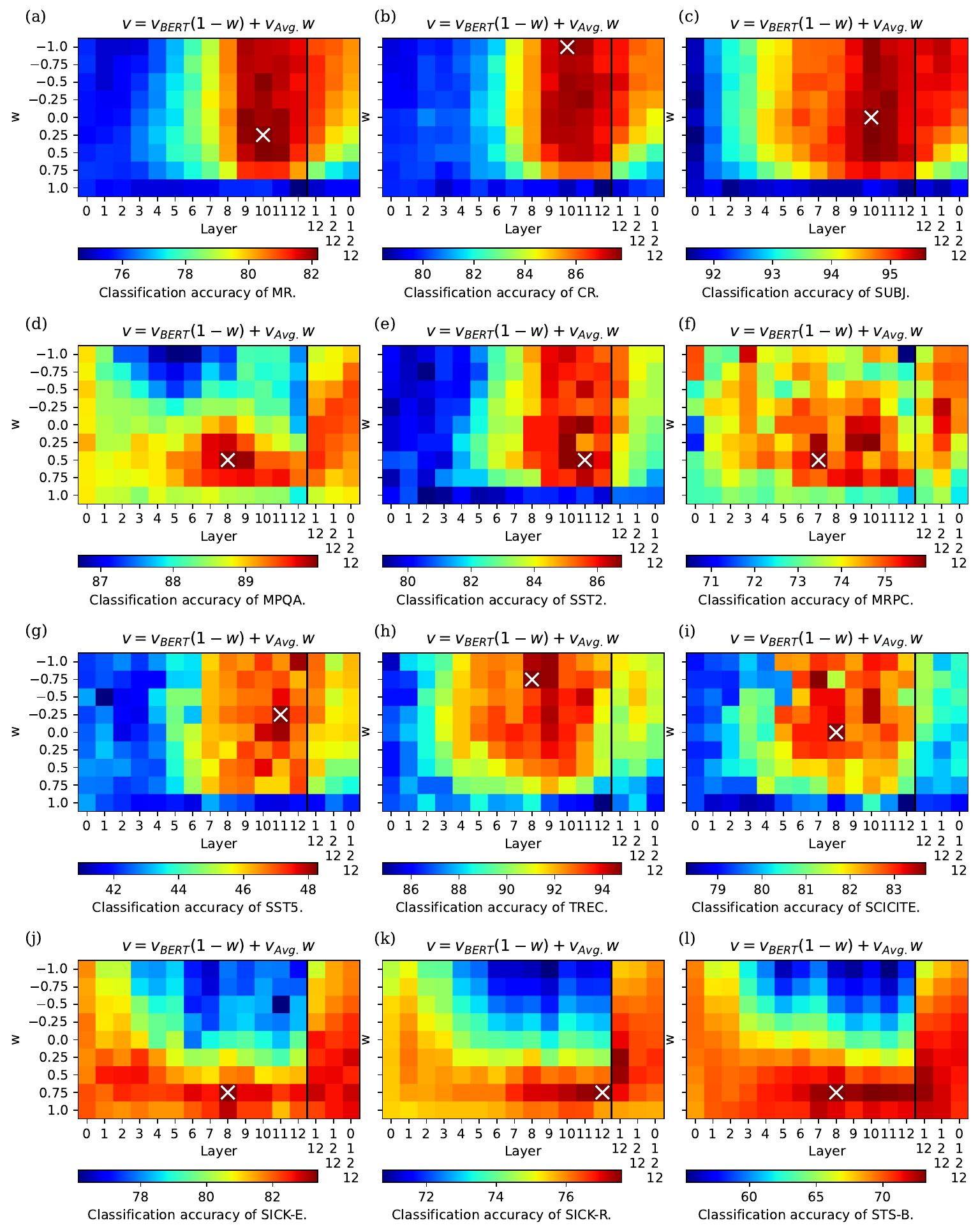}}
\caption{BERT + Avg. %
model individual task supervised classification performance dependence on the weight $w$ of Avg. model and layer, from which (for both models) representations are used. To the right of the black line on the horizontal axis, average aggregation of multiple layers is also shown. Tokens are simply averaged and no post-processing is used. The horizontal line with $w=0.0$ corresponds to a regular Bert (B) model, $w=0.5$ is B + Avg., and $w=1.0$ is the Avg. model. The white $\times$ marks the maximum value.}
\label{weights_times_layers_classification}
\centering
\end{figure}

\end{document}